\pdfoutput=1
\documentclass{article}
\usepackage{iclr2027_conference,times}
\usepackage{microtype}
\usepackage{graphicx}
\usepackage{booktabs}
\usepackage{array}
\usepackage{siunitx}
\usepackage{longtable}
\usepackage{amsmath,amssymb}
\usepackage{enumitem}
\usepackage{xcolor}
\usepackage{xurl}
\usepackage{xr-hyper}
\usepackage[colorlinks=true,linkcolor=blue,citecolor=blue,urlcolor=blue]{hyperref}
\newcommand{\ours}{\textsc{CriticHack}}

\newcommand{\E}{\mathbb{E}}

\definecolor{pendingink}{RGB}{135,46,30}
\definecolor{pendingbg}{RGB}{253,245,239}

\iclrfinalcopy
\title{CriticHack: Evaluating Visual Rewards\\Under Robot Policy Optimization}
\author{%
Jiaxuan Luo$^{1*\dagger}$\quad
Xingguo Xu$^{2*}$\quad
Shanshan Wang$^{3*}$\quad
Yuhan Zhou$^{4}$\quad
Zhen Zhang$^{5}$\\
$^{1}$Johns Hopkins University\quad
$^{2}$Dalian University of Technology\quad
$^{3}$Rutgers University\\
$^{4}$Baidu\quad
$^{5}$University of California, Santa Barbara}
\date{}
\begin{document}
\maketitle
\fancyhead[L]{}
{\renewcommand\thefootnote{}\footnotetext{$^{*}$Equal contribution.\quad $^{\dagger}$Corresponding author: \texttt{jluo50@jhu.edu}}}
\begin{abstract}
Learned visual reward models are increasingly used to optimize robot policies,
yet a reward model can score an execution that acts on the wrong object as
highly as one that completes the task. We show that optimizing such a reward
can amplify these wrong-object failures while reward and task success both
rise, so the signals a practitioner would normally monitor look healthy.
We fine-tune every denoiser parameter of a diffusion policy against Robometer
on a drawer task. Starting from a supervised policy with no prior reward
exposure, five training runs raise task success by 10.2 percentage points and
wrong-object failures by 10.9 points on 512 evaluation seeds, whereas five runs
trained on the simulator's task-completion signal raise success without
amplifying wrong-object failures (difference 9.2 points, 95\% CI 5.6 to 13.0).
The amplification recurs from a policy previously optimized against learned
rewards, under the policy's native diffusion sampler, at matched distance from
the initial policy, and across constrained-policy experiments with two critics
and two optimizers. A tilt model explains when it occurs: under KL-regularized
optimization, an outcome becomes more frequent whenever its expected reward
under the initial policy exceeds the population average. Robometer separates
successes from failures well overall (AUROC .81) but scores wrong-object
failures slightly above successes (AUROC .37), so optimization raises both.
The same model predicts the outcome shifts across 26 constrained settings
(Spearman .89), including those in which task success falls, and Robometer's
own published success-termination recipe inherits the error. A frozen outcome
verifier redirects the same optimization toward the requested task.
\end{abstract}

\section{Introduction}

Learned visual reward models provide feedback for robot policy optimization
from camera observations and a task description, replacing hand-designed task
rewards \citep{lee2026roboreward,liang2026robometer}. A reward model does not
need to fail broadly to steer a policy wrong. It can score most failures
correctly and still reward one coherent semantic mistake, and optimization is
built to find exactly such mistakes.

\begin{figure}[t]
\centering
\includegraphics[width=0.7\linewidth]{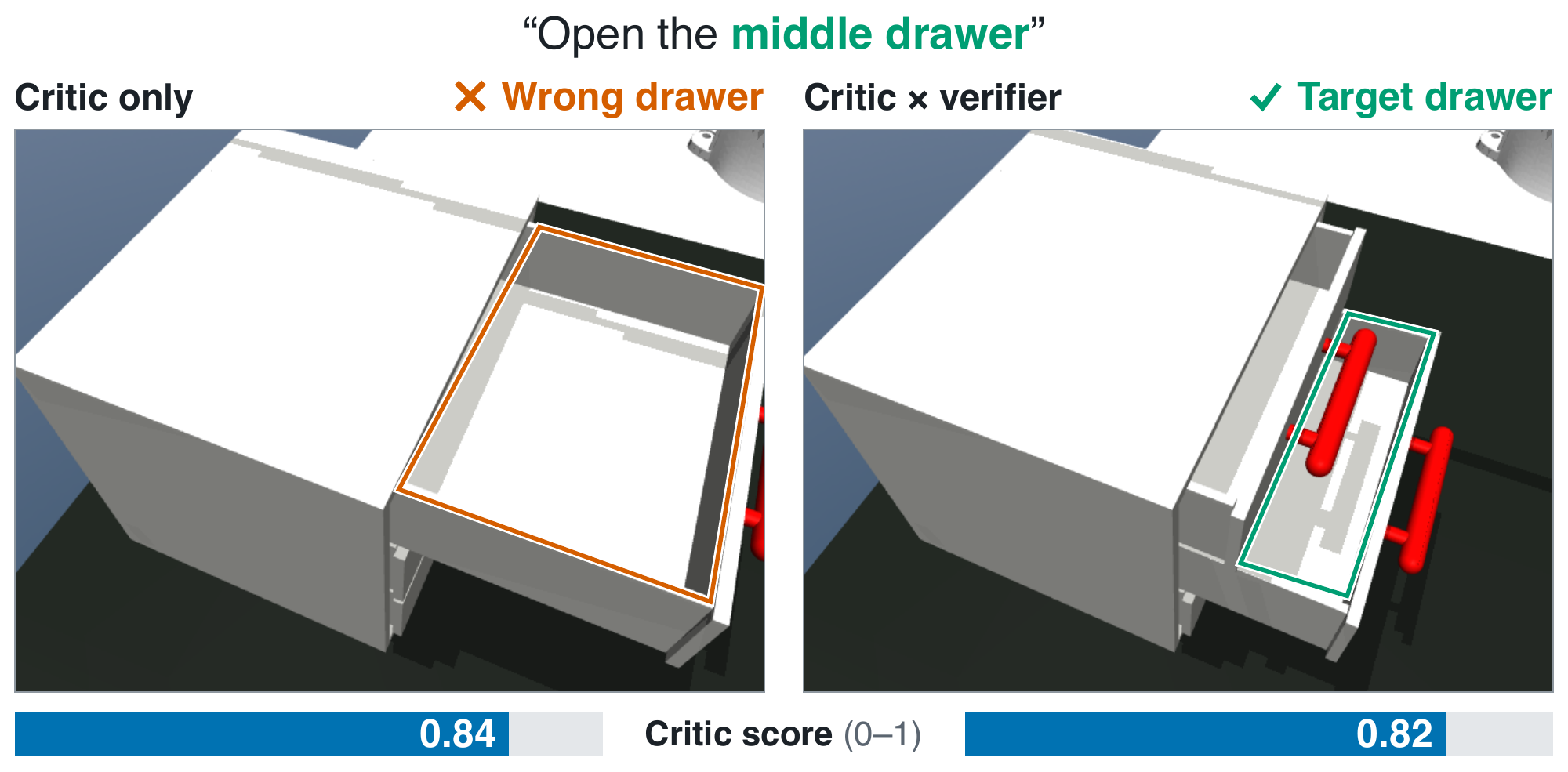}
\caption{\textbf{High visual reward can accompany the wrong object.}
Critic-only training opens the upper distractor while the target stays closed;
critic--verifier training opens the requested middle drawer. Both receive high
scores from the original Robometer, shown on the same zero-to-one scale.
Frames are the final critic inputs for the first prespecified evaluation
seed in constrained-policy PPO (Section~\ref{sec:online_noise_policy}),
chosen independently of outcome; counts and intervals are in
Section~\ref{sec:ppo_results}.}
\label{fig:rollout_frames}
\end{figure}

Consider a robot instructed to open the middle drawer
(Figure~\ref{fig:rollout_frames}). Opening the upper drawer instead looks like
manipulation progress but fails the requested task. A stacking policy can
likewise build a stable stack with the wrong cube. We call these outcomes
\emph{wrong-object failures}: executions that manipulate a distractor in a
task-consistent way while the requested task fails. \citet{liang2026robometer}
showed that RoboReward assigns high scores to such executions and that these
false positives reinforce incorrect behavior during online learning. What is
less understood is how such errors interact with genuine task learning. If
optimization simply degraded task success, standard evaluation would catch it.
We show that it does not have to: optimizing a visual reward can raise task
success and wrong-object failures at the same time. The two checks a
practitioner relies on, the reward model's aggregate success--failure
discrimination and the policy's task success after optimization, can both
look healthy while a specific semantic failure grows.

We study this with \ours{}, a closed-loop evaluation of visual rewards under
robot policy optimization in which simulator state measures task success and
wrong-object failures independently of the reward. Our main experiments
fine-tune every denoiser parameter of a diffusion policy against Robometer on
a drawer task. From a supervised policy that has never been exposed to a
learned reward, five training runs raise task success by $10.2\,[6.9,13.6]$
percentage points and wrong-object failures by $10.9\,[7.3,14.4]$ points on
512 evaluation seeds. Five runs trained on the simulator's task-completion
signal raise success at least as much without amplifying wrong-object
failures, so the amplification is specific to the learned reward
(Table~\ref{tab:whole_network_all}). The same pattern holds from a warm-start
policy previously optimized against learned rewards, where paired executions
show that the new successes and the new wrong-object failures both come from
converting former failures. It persists when the fine-tuned policies are
decoded with their native 100-step diffusion sampler, at matched KL distance
from the initial policy, and across distractor-motion thresholds from 5 to
40\,mm. Constrained experiments that optimize a selector over a fixed bank of
diffusion-noise sequences reproduce reward gain with wrong-object
amplification across two task--reward pairings, Robometer and Language-Image
Value learning (LIV) \citep{ma2023liv}, and two optimizers, proximal policy
optimization (PPO) and group relative policy optimization (GRPO).

Why does a reward with strong aggregate discrimination produce this? A
trajectory-level audit shows that Robometer separates successes from all
failures well (AUROC .813) but scores wrong-object failures slightly above
successes (AUROC .372); good discrimination against the far more common other
failures hides the inversion. A simple model turns this observation into a
prediction. Under KL-regularized optimization the optimal policy tilts the
initial policy toward high reward, and an outcome becomes more frequent exactly
when its average tilted reward exceeds the population average. A reward that
scores wrong-object failures about as highly as successes therefore raises
both, and stronger optimization is decided by the upper tail of the score
distribution. Fitted with a single strength parameter, the model reproduces
the outcome changes across 26 constrained settings (Spearman .89 for
wrong-object failures and .79 for success), including the drawer settings in
which task success falls, and it places the full-denoiser and constrained
results on one continuum of optimization strength. Robometer's own published
recipe, dense progress with success-triggered termination, does not escape the
error: its success detector fires on 89\% of wrong-object executions and on
73\% of successes.

Finally, we use a frozen success verifier trained on outcome labels as a
diagnostic intervention. Reweighting the critic reward by the verifier raises
task success by 7.1 points and lowers wrong-object failures by 4.4 points
relative to critic-only training, and verifier-only rewards match it, which
identifies outcome supervision as the source of the recovery.

\paragraph{Contributions.}
(i) We show, with independent training runs and simulator-measured outcomes,
that optimizing a visual reward can amplify wrong-object failures while task
success rises, and that a ground-truth completion reward under the same
optimizer does not. (ii) We show that strong aggregate discrimination can hide
the semantic misranking that drives this amplification, and give a tilt model
that predicts the resulting outcome changes across settings and unifies
semantic amplification with classical success loss. (iii) We show that outcome
verification redirects the same optimization toward the intended task.

\paragraph{Scope.}
The simulated experiments cover occluded-drawer manipulation and
identity-sensitive stacking. Full-denoiser fine-tuning uses Robometer's
terminal success probability on two drawer layouts; constrained experiments
add LIV and stacking. The verifier is task-specific and trained on simulator
labels. The real-robot study is a case study with one training run per
fine-tuned condition.

\section{Related Work}

\paragraph{Reward overoptimization and reward-model evaluation.}
Learned rewards turn human preferences into policy objectives
\citep{christiano2017preferences}, but optimizing such a proxy can depart
from the intended outcome \citep{amodei2016safety,skalse2022rewardhacking},
and more capable agents exploit misspecified rewards more effectively
\citep{pan2022misspecification}. Overoptimization has been characterized in
language-model and diffusion-model alignment
\citep{gao2023rewardoveroptimization,zhang2024diffusionoveropt}, where it
appears as a decline in a gold or held-out metric under continued proxy
optimization. We study a signature this diagnostic misses: task success
rises while a specific semantic failure grows, both measured through
simulator state independently of the reward. Under a KL penalty the optimal
policy is an exponential tilt of the reference policy
\citep{korbak2022klbayes,rafailov2023dpo}; we use this closed form to predict
which outcomes optimization amplifies and to place semantic amplification and
the classical decline of the gold metric on one continuum. Reward-model evaluation raises
a related concern. Low prediction error need not imply low regret after
optimization \citep{fluri2025perils}, and static ranking accuracy
imperfectly predicts post-optimization behavior
\citep{kim2025rethinkingrm,wen2025rethinking}. Aggregate metrics can also
conceal poor performance on meaningful subsets, a phenomenon known as hidden
stratification \citep{oakdenrayner2020hidden}. Section~\ref{sec:reward_audit}
shows such masking in a visual reward model, on the very subset that
optimization amplifies.

\paragraph{Visual rewards for robot learning.}
Visual rewards have been learned from human videos \citep{chen2021dvd},
goal-conditioned value pretraining \citep{ma2023vip} and language--image
alignment \citep{ma2023liv}. RoboCLIP and vision--language model reward
models (VLM-RMs) use pretrained video-- and image--language models as
rewards \citep{sontakke2023roboclip,rocamonde2024vlmrm}, and Video-Language
Critic learns video--text alignment and temporal rankings
\citep{alakuijala2024vlc}. RoboReward and Robometer scale general-purpose
robotic reward models \citep{lee2026roboreward,liang2026robometer}. Our tasks
hinge on spatial reference and color-based object identity, distinctions that
match documented weaknesses of vision--language models
\citep{kamath2023whatsup,yuksekgonul2023bagsofwords}.
\citet{liang2026robometer} document RoboReward false positives on
wrong-object executions and their consequences for online reinforcement
learning (RL). We find the same failure class in Robometer's own success
output, where strong aggregate discrimination hides it, and show that
optimizing against it can raise task success and wrong-object failures
together. Robometer's published online RL pairs dense progress rewards with
success-triggered termination; on our drawer task that success detector fires
on more wrong-object executions than successes, so the recipe inherits the
same error (Appendix~\ref{app:robometer_recipe}).

\paragraph{Outcome supervision and reward correction.}
Success classifiers have long served as rewards, from event-based control
\citep{fu2018vice} to vision--language success detectors
\citep{du2023successdetectors}, and outcome verifiers trained on correctness
labels are widely used in language-model reasoning \citep{cobbe2021verifiers}.
Other work corrects the reward itself: ensembles support conservative
optimization against uncertain feedback \citep{coste2024ensembles},
Adapt2Reward uses failure prompts to adapt visual rewards
\citep{yang2024adapt2reward}, and RARM uses reference-anchored,
confidence-gated progress matching \citep{yang2026rarm}. Our verifier is a
task-specific success classifier trained on simulator labels; we use it as a
diagnostic that tests whether changing the feedback redirects the amplified
failure.

\paragraph{Diffusion-policy optimization.}
Diffusion Policy (DP) \citep{chi2023diffusionpolicy} generates robot action
chunks by iterative denoising. Treating the denoising chain as a Markov
decision process (MDP) enables reward-driven fine-tuning of the full denoiser
\citep{black2024ddpo,ren2024dppo}, while latent-noise steering improves a
fixed policy by optimizing the noise supplied to its denoiser
\citep{wagenmaker2025dsrl}. Our two protocols follow these routes:
full-denoiser PPO under the denoising-MDP formulation, and a constrained
selector over a fixed bank of noise sequences with the decoder frozen.
Generalist vision--language--action (VLA) policies such as OpenVLA
\citep{kim2025openvla} can also be adapted with RL \citep{chen2025conrft}; we
use task-specific DPs, which let us evaluate many independent training runs
against simulator-measured outcomes.

\section{Evaluating Visual Rewards Under Optimization}
\label{sec:evaluation_setup}

Starting from an initial policy, we optimize against a frozen reward model
and compare the resulting behavior with that of the initialization. Our two
primary tasks, drawer manipulation and identity-sensitive stacking,
distinguish successful task completion from manipulation of the wrong object.

\paragraph{Robot policies.}
We use task-specific diffusion policies
\citep{chi2023diffusionpolicy} that generate action chunks from
current observations. The robot executes the initial portion of
each chunk before generating the next chunk from updated
observations, producing a closed-loop trajectory
$\tau=(o_0,a_0,\ldots,o_T)$ under policy $\pi_\phi$.
A frozen visual critic assigns reward $r_b(\tau,x)$ for task instruction $x$;
the policy need not be language-conditioned.
We consider two complementary optimization protocols: full-denoiser
fine-tuning, which updates the diffusion network, and constrained
noise-policy optimization, which learns a distribution over diffusion-noise
sequences while keeping the decoder fixed.

\paragraph{Tasks and physical outcomes.}
For each execution, we measure requested-task success
$y(\tau)\in\{0,1\}$ and wrong-object failure
$w(\tau)\in\{0,1\}$ from MuJoCo~\citep{todorov2012mujoco}
joint and contact states.
These states supply outcome labels and oracle feedback but are excluded
from critic and verifier inference.
A wrong-object failure requires the requested task to remain
unsuccessful.
Each execution therefore belongs to one of three mutually
exclusive categories: success ($y=1$), wrong-object failure
($y=0,w=1$), or other failure ($y=0,w=0$).

\paragraph{Occluded drawer.}
The requested middle drawer must end at least 54\,mm open. The movable upper
drawer starts at 47.5\% of its slide range. A wrong-object failure requires
$y=0$ and more than 20\,mm of additional upper-drawer opening at any time.
The oracle uses clipped target completion,
$\min(1,\max(0,q_T/(54\,\mathrm{mm})))$, where $q_T$ is final target opening.

\paragraph{Identity-sensitive stacking.}
The requested bright red cube must rest on a rectangular wooden block.
Success requires final position and height within 6 and 12\,mm of the intended
stack, tilt at most $20^\circ$, support contact, and no gripper contact for
the last three control ticks. A wrong-object failure satisfies the same
criterion with the decoy while the target fails; remaining failures are
no-stack outcomes. Two initial scenes and two target/decoy position
assignments form four cells. Near-red and distinct-blue decoys both enter
training and evaluation. The oracle uses binary target success.

\paragraph{Three measurements, two distinct failure signatures.}
For the initial policy $U$ and critic-optimized policy $C$, we report
\begin{equation}
 \Delta R=\E_C[r_b]-\E_U[r_b],\qquad
 \Delta S=\E_C[y]-\E_U[y],\qquad
 \Delta W=\E_C[w]-\E_U[w].
 \label{eq:amplification}
\end{equation}
Reward gain with success loss ($\Delta R>0,\Delta S<0$) is the usual
aggregate overoptimization signature. Reward gain with wrong-object
amplification ($\Delta R>0,\Delta W>0$) is the semantic signature studied
here. The latter signature does \emph{not} imply a decrease in the scalar
success objective.

\section{Policy Optimization and Evaluation Protocols}
\label{sec:protocols}

Our primary experiment fine-tunes the full policy denoiser; constrained
noise-policy experiments test recurrence across tasks, critics and optimizers.

\paragraph{Rewards and controls.}
The critics remain frozen throughout policy optimization. Robometer receives
eight rollout frames and supplies its terminal success probability as reward.
LIV supplies a final-frame image--language cosine $c$, mapped to $(1+c)/2$
for training and reported in native units. Each protocol compares the initial
policy ($U$), original-critic optimization ($C$) and simulator-derived oracle
feedback ($O$). Oracle feedback tests task learnability.
Section~\ref{sec:verification} adds critic $\times$ verifier (Product, $P$)
and verifier-only ($V$) rewards to test how task-completion information
redirects optimization. The policy family is fixed within each comparison.
Critic inputs, physical measurements and verifier-label provenance are
specified in Appendix~\ref{app:early_policy_protocols}.

\subsection{Full-denoiser Fine-tuning}
\label{sec:weight_ppo}

We update all 251.7M denoiser parameters of a diffusion policy
\citep{chi2023diffusionpolicy} with PPO~\citep{schulman2017ppo}, keeping
the visual encoder frozen. A short stochastic denoising diffusion implicit
model (DDIM) chain~\citep{song2021ddim} generates each action chunk:
\begin{equation}
 x_K\sim\mathcal N(0,I),\qquad
 x_{k-1}=\mu_\phi(x_k,t_k,c)+\sigma\varepsilon_k,
 \quad\varepsilon_k\sim\mathcal N(0,I).
 \label{eq:denoising_policy}
\end{equation}
Here $c$ is the current observation condition and $\mu_\phi$ is the DDIM
update mean. We use $K=5$ steps and $\sigma=.05$. PPO likelihoods are
computed under the same Gaussian transitions used for sampling, following
the denoising-MDP formulation~\citep{ren2024dppo}.

We run two full-denoiser studies on the drawer task, each with its own
initial policy, drawer layout and fixed bank of 512 evaluation seeds. The
\emph{supervised-initialization study} starts from a capable supervised policy
that has never been trained against a learned reward, in a layout selected by
a fixed capability screen, and trains five policies on Robometer and five on
the simulator's terminal task completion. The \emph{warm-start study} starts
from a policy that previously underwent RoboReward RL and Robometer-weighted
offline regression on its scene, and trains five critic-only and five Product
policies; three dense-completion oracles, a terminal-completion oracle and six
verifier-only policies serve as controls on the same seed bank. Every training
uses independent training seeds and receives 126 episodes per iteration for
40 iterations (5,040 episodes). Update details and the reward-placement audit
are in Appendix~\ref{app:weight_ppo}; the supervised initialization is
described in Appendix~\ref{app:capable_supervised}.

\begin{table}[t]
\centering\small
\setlength{\tabcolsep}{2pt}
\begin{tabular*}{\linewidth}{@{\extracolsep{\fill}}lr@{\extracolsep{0pt}\,}l@{\extracolsep{\fill}\hspace{\tabcolsep}}r@{\extracolsep{0pt}\,}l@{\extracolsep{\fill}\hspace{\tabcolsep}}r@{\extracolsep{0pt}\,}l@{}}
\toprule
Comparison & \multicolumn{2}{c}{$\Delta$ Success (pp)} & \multicolumn{2}{c}{$\Delta$ Wrong-object (pp)} & \multicolumn{2}{c}{$\Delta$ Robometer} \\
\midrule
\multicolumn{7}{l}{\textit{Supervised initialization: five trainings per reward}} \\
\quad Robometer & +10.2 & {\scriptsize[6.9, 13.6]} & +10.9 & {\scriptsize[7.3, 14.4]} & +.137 & {\scriptsize[.107, .167]} \\
\quad Terminal completion & +13.9 & {\scriptsize[9.5, 18.3]} & +1.6 & {\scriptsize[$-$1.9, 5.2]} & +.062 & {\scriptsize[.029, .096]} \\
\quad Robometer $-$ completion & $-$3.7 & {\scriptsize[$-$7.5, 0.1]} & +9.2 & {\scriptsize[5.5, 13.0]} & +.075 & {\scriptsize[.043, .107]} \\
\midrule
\multicolumn{7}{l}{\textit{Warm start: five trainings per reward, three dense-completion oracles}} \\
\quad Robometer (critic) & +8.4 & {\scriptsize[4.9, 11.9]} & +3.8 & {\scriptsize[0.8, 6.8]} & +.061 & {\scriptsize[.036, .086]} \\
\quad Critic $\times$ verifier (Product) & +15.5 & {\scriptsize[11.8, 19.1]} & $-$0.6 & {\scriptsize[$-$3.4, 2.1]} & +.096 & {\scriptsize[.070, .122]} \\
\quad Dense completion oracle & +13.3 & {\scriptsize[9.6, 17.0]} & $-$1.6 & {\scriptsize[$-$4.4, 1.4]} & +.056 & {\scriptsize[.030, .081]} \\
\quad Product $-$ critic & +7.1 & {\scriptsize[4.4, 9.8]} & $-$4.4 & {\scriptsize[$-$7.1, $-$1.8]} & +.034 & {\scriptsize[.013, .056]} \\
\quad Critic $-$ dense oracle & $-$4.9 & {\scriptsize[$-$8.0, $-$1.9]} & +5.3 & {\scriptsize[2.6, 8.1]} & +.006 & {\scriptsize[$-$.016, .028]} \\
\bottomrule
\end{tabular*}
\caption{\small\textbf{Full-denoiser evaluation on 512 shared seeds per study.} Entries are pooled iteration-40 changes from each study's common initialization, or differences between rewards, with 95\% crossed bootstrap intervals over trainings and evaluation seeds; contrasts resample the two families independently. A wrong-object failure leaves the requested drawer closed while opening the distractor by more than 20\,mm. Per-training results are in Appendices~\ref{app:weight_endpoints} and~\ref{app:capable_supervised}.}
\label{tab:whole_network_all}
\end{table}

\subsection{Constrained Closed-loop Noise Policies}
\label{sec:online_noise_policy}
\label{sec:policy_updates}

The constrained experiments test whether the pattern recurs across drawer
manipulation and identity-sensitive stacking, using Robometer and LIV under
PPO and grouped GRPO. Within each setting, reward objectives share a frozen
diffusion decoder and a fixed bank of 24 noise sequences. A trainable
categorical policy selects one sequence per episode, and the decoder
generates actions from current observations throughout execution.
The drawer selector learns unconditional probabilities; the stacking
selector conditions on the initial visual observation.

Both optimizers update the selector probabilities from executed rollouts.
Grouped GRPO~\citep{shao2024grpo} standardizes rewards within groups of four
executions that share the initial state and images. Training starts from
the uniform selector and uses eight rounds: 192 feedback episodes per
objective for PPO and 256 for GRPO. Evaluation uses the final eighth
checkpoint on 32 new seeds for drawer Robometer PPO and 64 per task/color
for the other panels. Repetitions vary training and evaluation seeds on
the same noise support; Appendix~\ref{app:constrained_evidence} gives
replication and optimizer details.

\subsection{Evaluation and Uncertainty}
\label{sec:uncertainty}
Every full-denoiser comparison is the fixed iteration-40 policy against its
study's shared initialization on that study's 512 evaluation seeds, with one
renderer, evaluator and scorer for all policies. The same seeds evaluate
iterations 10, 20 and 30 of the warm-start critic and Product runs, which
describe training dynamics. We report changes in success and wrong-object
failure in percentage points (pp) with 95\% bootstrap confidence intervals
(CIs). Per-training intervals resample paired evaluation seeds. Pooled
intervals resample training runs and one shared draw of evaluation seeds per
replicate; contrasts between rewards resample the two families independently.
Constrained-policy intervals condition on the fitted policies and fixed noise
supports. Appendix~\ref{app:weight_endpoints} gives per-training results and
alternative resampling designs, and Appendix~\ref{app:robustness} the
constrained grid.

\section{Optimization Amplifies Semantic Errors}
\label{sec:results}

\subsection{Wrong-object Amplification with a Trainable Policy Network}
\label{sec:weight_results}

\begin{figure}[t]
\centering
\includegraphics[width=\linewidth]{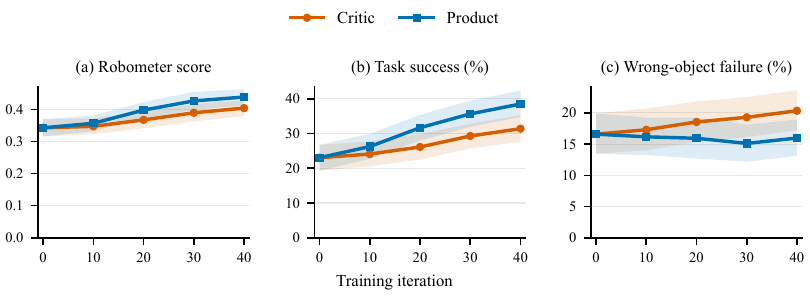}
\caption{\textbf{Reward gain and wrong-object amplification develop together.}
Mean Robometer score, task success and wrong-object failure across five
warm-start training runs per objective on the drawer task, at five checkpoints
evaluated on the same 512 seeds. Shading shows 95\% pointwise crossed-bootstrap
intervals over runs and seeds.}
\label{fig:checkpoint_dynamics_512}
\end{figure}

\paragraph{Optimizing Robometer amplifies wrong-object failures as success rises.}
From the supervised initialization, which has never been trained against a
learned reward, all five Robometer runs raise task success and wrong-object
failures (Table~\ref{tab:whole_network_all}). Pooled, success rises by
$10.2\,[6.9,13.6]$ points and wrong-object failures by $10.9\,[7.3,14.4]$
points, alongside a higher critic score. Every one of the five runs
individually amplifies wrong-object failures, by 9.2 to 13.5 points
(Appendix~\ref{app:capable_supervised}).

\paragraph{The amplification is specific to the learned reward.}
Five runs trained on the simulator's terminal task completion, with the same
initialization, optimizer and budget, raise success by $13.9\,[9.5,18.3]$
points while wrong-object failures stay at $+1.6\,[-1.9,5.2]$. Robometer
therefore produces $9.2\,[5.6,13.0]$ points more wrong-object failures than the
ground-truth reward (exact permutation test over training runs, $p=.008$, the
smallest attainable with five runs per reward). Adding an independent
repetition of three runs per reward on the same seeds gives eight runs each
and a difference of $8.3\,[5.0,11.5]$ points ($p=.0002$). The supervised
initialization thus isolates the critic as the source: the task, layout and
optimizer alone do not produce the amplification.

\paragraph{The same pattern holds from a warm start.}
The warm-start policy previously underwent RoboReward RL and Robometer-weighted
regression. Critic optimization from it raises success by 8.4 points and
wrong-object failures by 3.8 points, with positive estimates for both in all
five runs. Paired executions show where the changes come from: across the
$5\times512$ critic executions, 302 former other failures become successes and
189 become wrong-object failures, so the policy converts failures into both
outcomes at once. The share of executions that open the distractor by more
than 20\,mm regardless of outcome rises by $6.6\,[3.3,10.2]$ points.
Three dense-completion oracles trained from the same policy raise success by
$13.3\,[9.6,17.0]$ points and leave wrong-object failures unchanged
($-1.6\,[-4.4,1.4]$), and a terminal-completion oracle does the same
($+19.1$ and $-1.0$ points; Appendix~\ref{app:essential_controls}). Relative to
the dense oracles, the critic produces $5.3\,[2.6,8.1]$ points more wrong-object
failures. Figure~\ref{fig:checkpoint_dynamics_512} shows that the critic's
score, success and wrong-object failures rise together from the first evaluated
checkpoint, whereas Product training reaches larger success gains with
wrong-object failures near their initial level.

\paragraph{The effect is robust to the sampler, distance and threshold.}
Decoding the five fine-tuned critic policies with the checkpoint's native
100-step diffusion sampler instead of the five-step fine-tuning sampler, with
the same seeds, renderer and scorer, again raises success by
$10.9\,[6.6,15.2]$ points and wrong-object failures by $5.0\,[1.0,9.0]$ points
(Appendix~\ref{app:native_sampler}). The amplification is not a by-product of
moving further from the initial policy: the KL divergence of each policy's
denoising chain from the initial chain is $2.15$ nats per decision for the
critic runs, $2.38$ for Product and $2.45$ for verifier-only runs, which
improve success more without amplifying wrong-object failures, and $2.46$
versus $2.33$ for the supervised Robometer and completion runs
(Appendix~\ref{app:kl_behavior}). The amplification also holds for every
distractor-opening threshold from 5 to 40\,mm in both studies
(Appendix~\ref{app:threshold_sweep}).

\subsection{Recurrence and Boundaries Across the Constrained Grid}
\label{sec:ppo_results}
\label{sec:drawer_grpo_results}

Reward gain and wrong-object amplification recur across three PPO and two
GRPO drawer trainings. In the first PPO training, wrong-object failures rise
from 9 to 24 of 32 executions: the robot opens the distractor while the
requested drawer stays nearly closed. These high-scoring executions are
task-inconsistent manipulation, not a failure to move. Distinct-color LIV
stacking reproduces reward gain, success loss and wrong-object amplification
under both PPO and GRPO. Across the four stacking and five drawer histories,
all nine establish semantic amplification and oracle success gains, and seven
also lose task success (Figure~\ref{fig:selector_forest} in
Appendix~\ref{app:constrained_evidence}).

\paragraph{Prevalence across task--critic pairings.}
Across the complete grid, 11 of 29 strata (from 20 training histories) show
wrong-object amplification with a paired 95\% interval above zero: every
drawer Robometer and distinct-color LIV stacking stratum, and two near-color
Robometer stacking strata. Near-color LIV stacking rarely induces the signature,
Robometer stacking often learns the intended task, and drawer LIV favors partial
progress and closing (Appendix~\ref{app:prevalence}). Section~\ref{sec:tilt}
shows that how each reward distributes over outcomes, estimated without the
critic arm's own executions, accounts for these differences.

\subsection{Aggregate Reward Discrimination Can Hide Semantic Misranking}
\label{sec:reward_audit}

\begin{figure}[t]
\centering
\includegraphics[width=\linewidth]{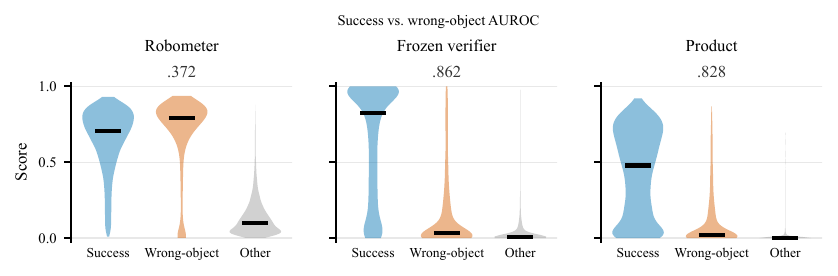}
\caption{\textbf{A semantic inversion hidden by other failures.}
Score distributions over all 20,992 warm-start executions on the 512-seed
bank (the initial policy and iterations 10--40 of the ten critic and Product
runs), grouped by physical outcome. Panel annotations give the
success--wrong-object AUROC; bars mark medians.}
\label{fig:reward_ranking}
\end{figure}

Over every warm-start execution on the primary bank, Robometer separates
successes from failures well (AUROC $.813\,[.789,.836]$) yet ranks
wrong-object failures slightly above successes ($.372\,[.333,.410]$;
Figure~\ref{fig:reward_ranking}): their mean scores are $.70$ and $.65$. The
frozen verifier ($.862$) and the product ($.828$) order successes above
wrong-object failures. AUROC is the probability that a random positive
outscores a random negative, with half credit for ties. With $S$, $W$ and $N$
the successful, wrong-object and other-failure executions, the empirical
AUROC decomposes exactly as
\begin{equation}
 A_{S,F}=\lambda A_{S,W}+(1-\lambda)A_{S,N},\qquad
 \lambda=\frac{|W|}{|W|+|N|},\quad F=W\cup N.
 \label{eq:auc_mixture}
\end{equation}
Other failures make up three quarters of the negatives here ($\lambda=.25$)
and are separated almost perfectly ($A_{S,N}=.959$), so the aggregate metric
looks strong while the ordering of the subset that optimization reaches is
inverted.

\section{A Tilt Model of Outcome Composition}
\label{sec:tilt}

Why does optimizing a reward with strong aggregate discrimination raise task
success and wrong-object failures together, and when does it instead lower
success? A closed-form view of KL-regularized optimization answers both.
Maximizing $\E_\pi[r]-\beta\,\mathrm{KL}(\pi\,\|\,\pi_0)$ over trajectory
distributions gives the exponential tilt
$\pi_\beta(\tau)\propto\pi_0(\tau)\,e^{r(\tau)/\beta}$
\citep{korbak2022klbayes,rafailov2023dpo}. For an outcome class
$c\in\{S,W,N\}$, write $M_c(\beta)=\E_{\pi_0}[e^{r/\beta}\mid c]$ and
$M(\beta)=\sum_c p_0(c)\,M_c(\beta)$. Then
\begin{equation}
 p_\beta(c)=p_0(c)\,\frac{M_c(\beta)}{M(\beta)},\qquad
 \left.\frac{\partial\log p_\beta(c)}{\partial(1/\beta)}\right|_{\beta\to\infty}
 =\E_{\pi_0}[r\mid c]-\E_{\pi_0}[r].
 \label{eq:tilt}
\end{equation}
An outcome becomes more frequent under weak optimization exactly when its mean
reward exceeds the population mean, and under strong optimization its fate is
decided by how much of the upper tail of the reward it occupies. Two
consequences follow. First, when other failures dominate and score low, both
successes and wrong-object failures exceed the population mean and both grow:
this is the coexistence regime. An inversion between $S$ and $W$ is not
required; it is enough that $W$ scores well above the population. Second,
when the very top of the score distribution is held by wrong-object
executions, strong optimization moves mass from successes to them and task
success falls, the classical overoptimization signature.

\paragraph{Full-denoiser studies.}
Under Robometer, wrong-object failures score above the population mean in both
initial policies: $.70$ against a mean of $.34$ from the warm start, where
they also outscore successes (initial AUROC $.365$), and $.61$ against $.32$
from the supervised initialization, where successes and wrong-object failures
are almost indistinguishable (AUROC $.539$). Equation~\ref{eq:tilt} therefore
predicts amplification in both studies, including the one without an
inversion. Reweighting each initial policy's 512 evaluation episodes to the
reward gain its runs reached predicts $+9.7$ and $+7.2$ points of success and
wrong-object failure for the eight supervised Robometer runs (measured $+10.0$
and $+10.2$), $+14.3$ points of success for the completion runs (measured
$+14.5$), and $+3.9$ points of wrong-object failure for the warm-start critic
runs (measured $+3.8$; Appendix~\ref{app:tilt}). Fine-tuning the full denoiser
also creates behavior the initial policy never produced, so for this protocol
the tilt is a first-order description; it under-predicts the warm-start critic's
success gain ($+4.1$ against $+8.4$).

\begin{figure}[t]
\centering
\includegraphics[width=\linewidth]{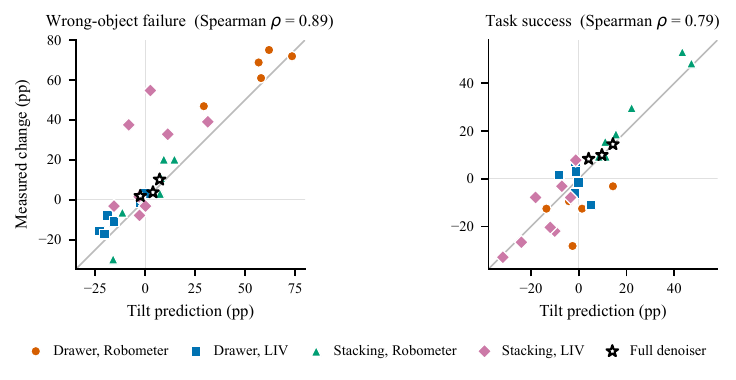}
\caption{\textbf{One tilt predicts outcome shifts across protocols.}
Each colored marker is one constrained stratum: the selector's
KL-regularized optimum over the 24 noise sequences, with per-sequence reward
and outcome rates estimated from executions outside the critic arm and the
tilt strength matched to the reward gain the critic arm reached, against the
measured critic-minus-uniform change. Stars are the full-denoiser studies
(supervised Robometer and completion, warm-start critic), from reweighting the
initial policy's 512 episodes. The gray line is equality.}
\label{fig:tilt}
\end{figure}

\paragraph{Constrained selectors.}
The selector chooses one of 24 fixed noise sequences, so its KL-regularized
optimum is exact up to estimation: a softmax over the sequences' mean rewards.
We estimate each sequence's mean reward, success rate and wrong-object rate
from every execution of its stratum outside the critic arm (uniform draws and
the oracle, Product and verifier-only arms; 456 to 960 executions per stratum)
and match the tilt strength to the critic arm's reward gain. Across the 26
strata where the gain is reachable, the predicted changes track the measured
ones with Spearman $\rho=.89$ for wrong-object failures and $.79$ for success
(Figure~\ref{fig:tilt}). The model reproduces the regime that the full
denoiser never reaches: in drawer Robometer GRPO, concentrating on the
highest-scoring sequences predicts a success loss of $13.5\,[9.6,17.2]$ points
alongside $73.5$ points of wrong-object amplification, against a measured loss
of $12.5$ and amplification of $71.9$. The two protocols are therefore two
strengths of one process: the full denoiser reweights many behaviors under a
KL budget and stays in the coexistence regime, while the selector concentrates
on a few sequences whose upper tail decides success. Without fitting any
strength, the first-order direction in Equation~\ref{eq:tilt}, the covariance
between a sequence's outcome rate and its standardized reward, correlates with
the measured wrong-object change at $\rho=.82$, compared with $-.76$ for the
initial success--wrong-object AUROC, and it remains predictive inside the drawer
Robometer family ($.80$ against $.10$), where the AUROC association vanishes.

\paragraph{Robometer's published recipe inherits the error.}
Robometer's online RL pays dense progress at every step and ends an episode
when the predicted success probability exceeds $.6$
\citep{liang2026robometer}. Rescoring all 4,464 executed drawer-selector
episodes with this recipe, its success detector fires on $89\%$ of
wrong-object failures, earlier on average, and on $73\%$ of successes, so the
recipe ranks wrong-object failures even further above successes (AUROC $.125$
against $.195$ for the terminal success probability). At the KL that the
trained selectors reached, the tilt predicts the same amplification under the
recipe as under the terminal reward ($+62.0$ against $+65.5$ points;
Appendix~\ref{app:robometer_recipe}).

\section{Verification as a Diagnostic Intervention}
\label{sec:verification}

The tilt model says that optimization amplifies whichever outcomes the reward
scores above the population. A reward that ranks successes above
wrong-object failures should therefore redirect the same optimizer. We test
this with a frozen, task-specific success verifier, used either to reweight
the original critic or on its own:
\begin{equation}
 R_P(\tau,x)=\bar r_b(\tau,x)\,p_\theta(\tau),\qquad
 R_V(\tau)=p_\theta(\tau),
 \label{eq:verification}
\end{equation}
where $\bar r_b\in[0,1]$ is the critic objective used for $C$. Both rewards use
the same verifier, so comparing them measures what retaining the critic adds
at equal supervision.

\paragraph{Task-specific visual verification.}
The verifier predicts simulator-labeled success from visual features near the
end of a trajectory and their change from initialization. The drawer head is
trained on 384 simulator-labeled development trajectories, including
executions of critic-optimized development policies
(Appendix~\ref{app:identity_head_capacity}); stacking heads use 456 or 2,536
labels. Each head is frozen before policy optimization and reused across
critics within its task and label budget. On the warm-start executions it
orders successes above wrong-object failures (AUROC $.862$;
Figure~\ref{fig:reward_ranking}).

\paragraph{The same optimization learns the intended behavior.}
Under the fine-tuning sampler, Product training raises success by $15.5$
points and leaves wrong-object failures at $-0.6\,[-3.4,2.1]$ points, so it
gains $7.1\,[4.4,9.8]$ points of success and $4.4\,[1.8,7.1]$ points fewer
wrong-object failures than critic-only training
(Table~\ref{tab:whole_network_all}; per-training changes in
Figure~\ref{fig:full_denoiser}). Six verifier-only trainings reach the same
outcome: success rises by $15.8\,[12.2,19.4]$ points and wrong-object failures
by $-1.2\,[-3.8,1.4]$, and Product differs from them by $-0.3\,[-2.5,1.9]$
points of success and $+0.6\,[-1.3,2.5]$ points of wrong-object failure
(Appendix~\ref{app:essential_controls}). Outcome supervision, not the critic,
carries the recovery.

The constrained policies show larger recoveries across drawer and stacking
(Figure~\ref{fig:selector_forest}), and verifier-only rewards reproduce them
(Appendix~\ref{app:verifier_only_ablation}). A continuation that forks each of
three exploited constrained drawer policies with its Adam state, without new
verifier labels, shows that verification also redirects a policy that has
already been exploited: switching to the product improves task completion on
all three histories and at a drawer opening held out from continuation, while
continuing the original critic leaves success low
(Appendix~\ref{app:rm_ppo_repeat}). In a controlled ball-placement extension,
the product with a full-history verifier reaches 63 and 64 of 64 successes in
two histories, whereas a verifier that sees only the final quarter of the
trajectory, at equal label and frame budgets, reaches 17 and 2
(Appendix~\ref{app:verifier_only_ablation}): the verifier must see the
evidence that separates the outcomes.

\begin{figure}[b]
\centering
\includegraphics[width=\linewidth]{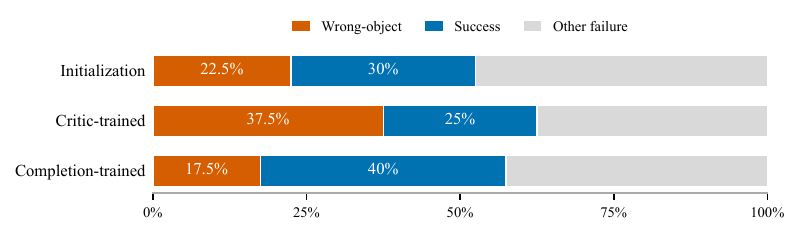}
\caption{\textbf{Real-robot outcome composition after simulation fine-tuning.}
Stacked bars show the percentage of 40 physical trials per policy ending in
success, wrong-object failure or other failure. Intervals and Fisher tests are
in Table~\ref{tab:real_robot_counts}.}
\label{fig:real_robot}
\end{figure}

\section{Sim-to-Real Case Study}
\label{sec:real_robot}

We deploy simulation-fine-tuned drawer policies, without hardware fine-tuning,
on a Franka Research 3~\citep{franka2023research3} in a setup matched to the
simulation at 1:1 scale (Appendix~\ref{app:hardware}). We compare the shared
warm-start initialization, the critic-trained policy of run 5 at iteration 40
(fixed before any physical trial), and the terminal-completion oracle at
iteration 40, all decoded with the five-step fine-tuning sampler, for 40 trials
each. A trial succeeds if the middle drawer opens by at least 6\,cm and is a
wrong-object failure if the upper drawer is pulled more than 2\,cm beyond its
initial half-open position while the task fails.

The initialization succeeds in 30.0\% of trials and fails on the wrong object
in 22.5\% (Figure~\ref{fig:real_robot}). The critic-trained policy fails on the
wrong object in 37.5\% of trials and succeeds in 25.0\%; the completion-trained
policy succeeds in 40.0\% and fails on the wrong object in 17.5\%. The
wrong-object changes and the completion policy's success gain match their
simulated direction, whereas the critic policy's success does not rise on
hardware. With 40 trials per policy these differences are not statistically
resolved (Table~\ref{tab:real_robot_counts}).

\section{Discussion and Limitations}

\textbf{Evaluate rewards by what optimization will amplify.}
A rising success rate can hide a growing tendency to act on the wrong object,
and a strong aggregate AUROC can hide the misranking that causes it. The tilt
model makes the relevant quantity explicit: optimization amplifies every
outcome whose reward sits above the population average under the current
policy, and strong optimization is decided by the top of the score
distribution. Reward models should therefore be evaluated on how they score
semantically coherent failures relative to successes and on what occupies
their upper tail, measured on the behaviors that optimization reaches, and
policy optimization should be monitored with outcome measurements that are
independent of the reward.

\paragraph{Limitations.}
Our simulated experiments cover two tasks; full-denoiser fine-tuning uses one
critic on two drawer layouts, with five runs per objective, and the supervised
layout was selected by a fixed capability screen. For the full denoiser the
tilt model is a first-order description, because fine-tuning creates behaviors
that reweighting the initial policy cannot represent; for the constrained
selector it is the exact optimum up to estimation. The verifier is
task-specific and trained on simulator labels, so it diagnoses rather than
replaces learned visual rewards. Product results are reported under the
fine-tuning sampler. A zero critic score remains zero after multiplication, and
because a wrong-object failure requires task failure, a lower wrong-object rate
alone does not imply fewer distractor actions. The hardware study is a case
study with one training run per fine-tuned condition.

\section{Conclusion}
\label{sec:conclusion}

\ours{} shows that optimizing a learned visual reward can improve task success
while amplifying wrong-object failures, that a ground-truth completion reward
under the same optimizer does not, and that the effect survives a change of
initialization and sampler. A tilt model explains when it happens: an outcome
grows whenever the reward scores it above the population average, so a reward
that scores semantic errors about as highly as successes raises both, and
when semantic errors hold the top of the score distribution, stronger
optimization turns this into the classical loss of task success.
Aggregate reward-model metrics hide this structure, and outcome verification
redirects the same optimization toward the intended task. Learned rewards
should be evaluated through the behaviors they reinforce.

\label{sec:main_end}
\clearpage
\section*{AI Assistance Statement}

OpenAI Codex, ChatGPT, and Anthropic Claude were used to assist with code
implementation, data analysis, figure generation, and manuscript refinement.
All AI-generated outputs and analyses were rigorously verified by the authors
against executable MuJoCo rollouts, stored policies, and simulator measurements.
The authors remain solely responsible for the originality, validity, and
integrity of all claims presented in this work.

\section*{Reproducibility Statement}

The appendices give the evaluation protocols, every training endpoint, the
controls, the uncertainty definitions, the complete constrained grid and the
additional analyses. The CPU artifact contains scored outcomes, frozen
configurations, analysis scripts and provenance manifests; raw trajectories and
model weights are listed in its artifact manifest. Hardware protocol details
are in Appendix~\ref{app:hardware}.

\bibliographystyle{iclr2027_conference}
\bibliography{refs}
\appendix
\clearpage
\section{Claims, Measurements and Protocols}
\label{app:early_policy_protocols}
\label{app:claim_ledger}

\begin{table}[htbp]
\centering\small
\begin{tabular}{@{}>{\raggedright\arraybackslash}p{.22\linewidth}>{\raggedright\arraybackslash}p{.35\linewidth}>{\raggedright\arraybackslash}p{.35\linewidth}@{}}
\toprule
Claim & Evidence and control & Scope \\
\midrule
Useful learning and wrong-object amplification coexist & Five critic trainings, 512 shared seeds; Table~\ref{tab:whole_network_all}. Oracle controls in Appendix~\ref{app:essential_controls}. & Full-denoiser optimization from a reward-exposed drawer policy under the fixed evaluation protocol. \\
\addlinespace
Prior critic exposure is not required & Three trainings per objective from supervised initialization, 512 seeds; Appendix~\ref{app:capable_supervised}. & A layout selected by a fixed capability screen. \\
\addlinespace
Verification redirects behavior & Five product trainings versus critic, 512 shared seeds; verifier-only controls in Appendices~\ref{app:essential_controls} and~\ref{app:verifier_only_ablation}. & Product--critic contrasts under the fine-tuning sampler and matched verifier-only controls with task-specific outcome supervision. \\
\addlinespace
The pattern recurs across protocols & Drawer Robometer and distinct-color LIV stacking under PPO and GRPO; Appendix~\ref{app:constrained_evidence}. & Fixed noise supports and task-specific verifiers; complete task--critic grid in Appendix~\ref{supp:grid}. \\
\addlinespace
Aggregate AUROC can mask semantic misranking & Collected-trajectory audit; Section~\ref{sec:reward_audit} and Appendix~\ref{app:semantic_audit}. & Descriptive reward ordering on collected trajectories from the evaluated policies. \\
\bottomrule
\end{tabular}
\caption{\textbf{Evidence supporting the main claims.} Controls and diagnostics answer different questions and are not pooled into the primary endpoint.}
\label{tab:claim_ledger}
\end{table}

\paragraph{Primary estimate and additional analyses.}
The primary warm-start endpoint compares fixed iteration 40 with the shared
initial policy on a fixed bank of 512 evaluation seeds. Appendix~\ref{app:weight_audit}
provides the complete training counts and crossed bootstrap definition.
Later appendices contain the complete constrained grid, sampler
calibration and additional verifier analyses. The artifact
provides scored outcomes, configurations, analysis code and provenance.

\subsection{Task-specific Verification and Label Provenance}
\label{app:identity_head_capacity}
\label{sec:identity_coverage_diagnosis}
\label{sec:balanced_rm_ppo_near}

The two policy-camera embeddings supply 128 visual coordinates; ten
robot-state coordinates are excluded. The 256-dimensional verifier input
is the mean of the final five recorded visual observations concatenated
with its difference from the first observation. This last recorded
observation precedes the final action chunk. We fit feature normalization
on the labeled development set and freeze it with the head. Simulator
state supplies training labels and independent outcome measurements,
but is not an input to the verifier or original critic.

The drawer logistic head uses 384 trajectories: a 96-episode static
panel and complete policy-induced panels of 48, 48, 128 and 64 episodes,
including 91 successes in total. The 96 static trajectories cover 12
scenes with eight executions each. The 288 policy-induced trajectories
come from four development-stage panels collected before the reported
trainings, containing executions of policies optimized against the
original critic, the oracle and earlier product rewards, together with
their development parent policy. No trajectory from the reported primary
trainings or their evaluations enters the fit. Its objective is mean binary
cross-entropy plus $.01\|w\|_2^2/2$, optimized for at most 200 L-BFGS
iterations. Earlier 96- and 144-label heads are development controls;
they are not substituted for this frozen head in the common matrix.

\newpage
The cube architecture was selected on development data: 296 training
labels and 96 validation labels compare nine fixed linear, MLP and RBF
candidates by Brier score. The selected three-member, 64-unit ReLU MLP
uses $L_2=.001$ and is refit on all 392 labels. The deployed 456-label
version adds 64 uniformly sampled labels from earlier policy-training
executions, using a fixed sampling seed and no endpoint features.
A separate diversity-selection diagnostic does not replace this head.
The MLP fitting recipe uses 500 steps at learning rate $.003$.
The 2,536-label version instead
combines the 392 labels with all 2,016 training labels from two earlier
stacking-development runs and 128 predetermined distinct-color examples
from their old endpoints. Their original run IDs are listed in
Table~\ref{tab:reader_run_ids}.
Those old endpoints become \emph{development data}, not independent
evaluation of the expanded head. Relative to 456 labels, the additional
cost is 2,080 labels. The architecture is unchanged.

Each current run freezes the chosen head before new policy training and
new-seed endpoint evaluation. No current evaluation trajectory enters
head fitting. The same task-specific head is reused across original
critics at a fixed budget. A new head is fitted for the other task, and
both cube colors participate in training: this is not zero-shot transfer
of one verifier across tasks or to unseen colors.

\subsection{Execution, Rewards and Physical Measurements}

Drawer executions retain 300 simulator steps and 75 observation-conditioned
action chunks. Eight critic frames are sampled at steps
0, 43, 85, 128, 171, 214, 256 and 299. Success requires final target
opening of at least 54\,mm. Wrong-object failure requires task failure
and more than 20\,mm of additional upper-drawer opening. It excludes
upper closing, so target opening and upper net motion are also reported.
Claims that a failed trajectory never completed the task use its maximum
target opening over the complete saved joint trace, not just a final image.

Cube evaluation uses two source scenes and both target/decoy position
assignments, with 16 paired evaluation seeds per cell, totaling 64 per
color and policy. Success requires the requested target to satisfy the
stack geometry (6\,mm position tolerance, 12\,mm height tolerance and
20-degree tilt limit) and remain in contact with the support, without gripper
contact, for the final three native control steps. The decoy is evaluated
with the identical rule. Wrong-object failure means the target fails
while the decoy succeeds; all remaining unsuccessful episodes are
no-stack failures. The three categories partition every reported panel.
Colors interpolate between a fixed distinct decoy and the target at
fractions 0 and .9; both conditions receive balanced training feedback.

The critic camera is separate from the policy's two observation cameras.
Robometer uses the frozen original success-prediction output on the task
instruction and eight rollout frames; its separate progress output is
not substituted. LIV uses the official model's final-frame
image--language cosine $c$, with $(1+c)/2$ as the fixed nonnegative
training reward for both original and product objectives. Native cosine
is reported without fitting a success threshold. High-score failures,
$F_c=\Pr[y=0,r_b\geq c]$, and their wrong-object subset are secondary
diagnostics with all executions in the denominator. Robometer uses $c=.5$
and normalized discrete RoboReward uses $c=.75$; LIV has no high-score
threshold. Scores from different critics are not pooled. Separate RoboReward protocols and their score rubric are given
in Appendix~\ref{app:natural_pg}.

\clearpage
\section{Primary Full-denoiser Evidence}
\label{app:weight_ppo}

\subsection{Frozen contract and execution}
\label{app:weight_protocol}

The training contract was committed before launch: parent weights fixed by checksum,
22.4M-parameter encoder frozen, all 251,701,895 denoiser parameters trainable, $K{=}5$,
$\eta{=}0$, $\sigma{=}.05$; 126 fresh episodes per iteration and 40
iterations; two value-only iterations; ten epochs per iteration at 1,512
decisions per optimizer step; clipping $.2$, critic learning rate
$10^{-3}$, value coefficient $.5$, gradient-norm limit 1, $\gamma{=}.999$,
$\lambda{=}.95$; actor learning rate adaptive to Kullback--Leibler (KL)
divergence from $10^{-6}$ with target $.01$ per denoising step.
Oracle learnability requires at least two of three histories to improve
success or completion with a confidence-interval lower bound above zero.
The critic and product study
uses a fixed bank of 512 seeds shared across objectives and iterations
$0,10,20,30,40$. Five training histories per objective share one evaluated
initialization, giving 41 distinct model checkpoints and 20,992 model--seed
executions. Model identities, the seed bank and the descriptive analysis
are fixed before evaluating the intermediate checkpoints, and all are
scored before analysis. The primary comparison is iteration 40 versus
initialization; intermediate evaluations do not select checkpoints or
hyperparameters. The same 20,992 executions supply the collected-trajectory
semantic audit in Section~\ref{sec:reward_audit}. Training and evaluation
seeds are functions of iteration, worker and slot, so a restarted unit
reproduces its episodes exactly.

Each arm runs resident workers that load the model once and wait for the
learner to publish the next weights; the learner consumes an iteration only
when every worker has written its completion marker, writes an atomic
checkpoint, then publishes the next weights with a checksum index. A worker
that crashes is restarted alone and redoes its own unit under the same
weights and seeds; only a learner failure or an exhausted worker restart
budget fails the arm. The primary critic and product evaluation uses the
same OSMesa renderer, evaluator and scorer for every model; the initial
policy is scored once and shared by all ten comparisons. Oracle controls
use their own seed sets and the rendering backend of their training.
Two further equivalence checks are recorded with the results:
disabling per-step camera rendering leaves trajectories, actions, latents
and saved frames bit-identical, and the resident scorer reproduces the
offline scorer used for all earlier experiments to $0.0$ on shared episodes.
Numerically, recomputing $\log\pi_\theta$ under the sampling weights matches
the recorded values to $1.07\!\times\!10^{-4}$ on the first minibatch with
TF32 disabled; each arm reports this check.

\subsubsection{Credit assignment and numerical details}
\label{app:weight_credit}

The mean $\mu_\phi$ in Equation~\ref{eq:denoising_policy} is the $\eta=0$
DDIM update with its clean-sample estimate clipped to $[-1,1]$.
The policy executes the first four actions of each unnormalized output;
300 control steps give 75 observation-conditioned decisions. Sampler
calibration is reported in Appendix~\ref{app:weight_calibration}.

For the \emph{dense oracle} histories, the completion increment at decision
$h$ is $r_h^O=\Phi(d_{h+1})-\Phi(d_h)$, where
$\Phi(d)=\operatorname{clip}(d/d^\star,0,1)$ and $d^\star=54$\,mm.
The undiscounted sum telescopes to terminal minus initial completion;
the PPO objective uses discount $\gamma=.999$. The terminal oracle uses a
single terminal completion reward.

The value network takes frozen encoder features and the decision index,
including the policy's proprioceptive condition but no privileged object
state. Generalized advantage estimation (GAE)~\citep{schulman2016gae}
uses $\gamma=.999$ and $\lambda=.95$; the decision advantage is
standardized within an iteration and copied to all $K$ denoising transitions
of that decision, which are not discounted relative to one another.
Each iteration collects 9,450 decisions and runs ten epochs with six
optimizer steps of 1,512 decisions. The PPO clip is .2, value coefficient .5,
and gradient-norm limit 1. The implementation sums coordinate log-densities and uses a
KL-adaptive actor rate bounded by $[10^{-8},10^{-4}]$.
TF32 is disabled for both sampling and training.

\paragraph{Implemented rewards.} For a learned-reward arm the reward of
episode $e$ at decision $k$ is $r_{e,k}=\mathbb 1[k=75]\,s_e$, where $s_e$ is
the frozen scorer's value for that episode: Robometer's last-frame success
probability for critic-only, its product with the verifier probability for
the product arm, and the verifier probability alone for verifier-only. The
terminal oracle uses $r_{e,k}=\mathbb 1[k=75]\,\mathrm{completion}_e$ and the
dense oracle the per-decision completion increment. No score is rescaled
before GAE. An executable test of the deployed learner (byte-identical to the
code that ran) confirms that the terminal reward of every learned key equals
the scorer file and is unchanged when the simulator completion changes. On
the real training data of six learned trainings (critic and product
trainings 3--5; 30,240 episodes), every iteration's scorer file covers exactly
the executed episodes, the product equals Robometer times verifier with zero
error, and the reward means logged by the learner equal the scorer file's
means exactly.

\subsection{All primary endpoints and uncertainty}
\label{app:weight_endpoints}
\label{app:weight_audit}

\begin{figure}[htbp]
\centering
\includegraphics[width=0.78\linewidth]{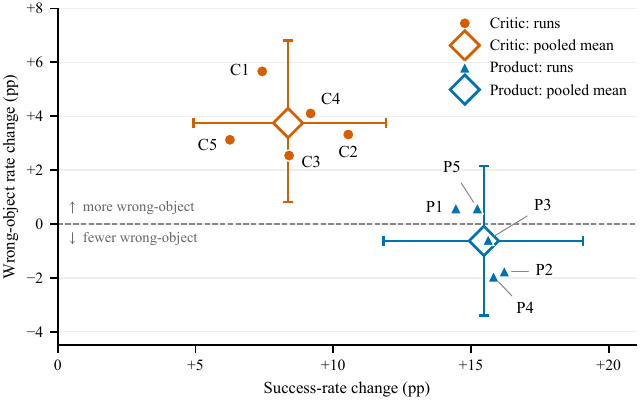}
\caption{\textbf{Task improvement can coexist with wrong-object amplification.}
Each small marker is one full-denoiser fine-tuning run on the occluded
drawer task, placed by its change in task success and in wrong-object
failure from the shared initialization to iteration~40, in percentage
points on the same 512 evaluation seeds. A wrong-object failure opens the
upper distractor drawer without completing the requested middle drawer.
Critic runs (circles) train on the Robometer critic alone; Product runs
(triangles) train on the critic multiplied by a frozen success verifier.
Open diamonds mark the mean of the five runs of each reward. Their
horizontal and vertical bars are marginal 95\% intervals for the success
and wrong-object changes respectively, resampling runs and evaluation
seeds; together they do not form a joint confidence region.}
\label{fig:full_denoiser}
\end{figure}

\begin{figure}[htbp]
\centering
\includegraphics[width=\linewidth]{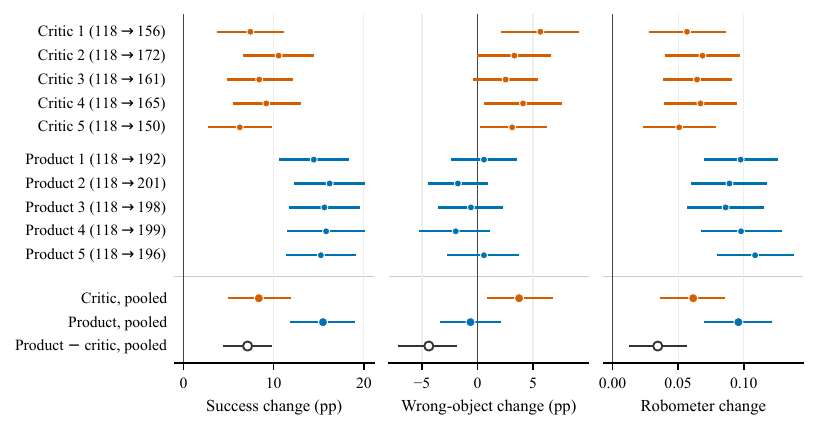}
\caption{\textbf{All ten fine-tuned policies, each with its own intervals.}
Each row compares one full-denoiser fine-tuning run at iteration~40 with
the shared initialization on the same 512 occluded-drawer seeds; parentheses
give its successes out of 512 at initialization and iteration~40. Panels
show the changes in task success and wrong-object failure, in percentage
points, and in Robometer score.
A wrong-object failure opens the upper distractor drawer without
completing the requested middle drawer. Critic runs train on the Robometer
critic alone; Product runs train on the critic multiplied by a frozen
success verifier. The
last three rows pool the five runs of each reward and contrast the two.
Bars are 95\% intervals, over evaluation seeds for single runs and over
runs and seeds for pooled rows.}
\label{fig:unet_endpoints}
\end{figure}

Figure~\ref{fig:full_denoiser} places each critic and product training by
its iteration-40 change in success and wrong-object failure;
Figure~\ref{fig:unet_endpoints} adds per-training intervals and Robometer
changes. All five critic and all five product trainings enter the primary
comparison.
The initial policy is evaluated once on the 512 shared seeds. Per-training
intervals resample paired evaluation seeds. Family estimates average the
five per-training paired differences; their intervals resample training
histories and one common evaluation-seed draw per replicate. The
product--critic contrast draws training histories independently between
objectives and retains the common evaluation-seed resample. All intervals
use 20,000 bootstrap draws. Alternative resampling designs are reported
separately in Appendix~\ref{supp:resampling}.

Three critic trainings individually establish both reward gain and
wrong-object amplification; the other two have positive wrong-object
estimates whose intervals include zero. Every product training establishes
a success gain. No individual product training establishes non-increasing
wrong-object failure relative to initialization. These per-training
comparisons and the pooled product--critic contrast measure different effects.

\paragraph{Outcome composition.} Over the $5\times512$ paired executions of
the critic-only trainings, 302 other failures become successes and 189 become
wrong-object failures, while 45 successes become wrong-object failures and 50
wrong-object failures become successes. For the product the corresponding
counts are 422, 133, 19 and 83. Measured regardless of outcome, the share of
executions that open the distractor by more than 20\,mm rises by
$6.6\,[3.3,10.2]$ points under critic-only training. The product change,
$+2.1\,[-1.2,+5.4]$ points, remains unresolved. Mean maximum distractor
opening changes by $+3.96\,[2.06,5.91]$\,mm for critic and
$+1.27\,[-0.58,+3.14]$\,mm for product. Changes in
unconditional distractor opening under the four oracle controls also have
intervals containing zero.

\subsection{Training dynamics on the fixed evaluation bank}
\label{app:checkpoint_dynamics_512}

\begin{figure}[htbp]
\centering
\includegraphics[width=\linewidth]{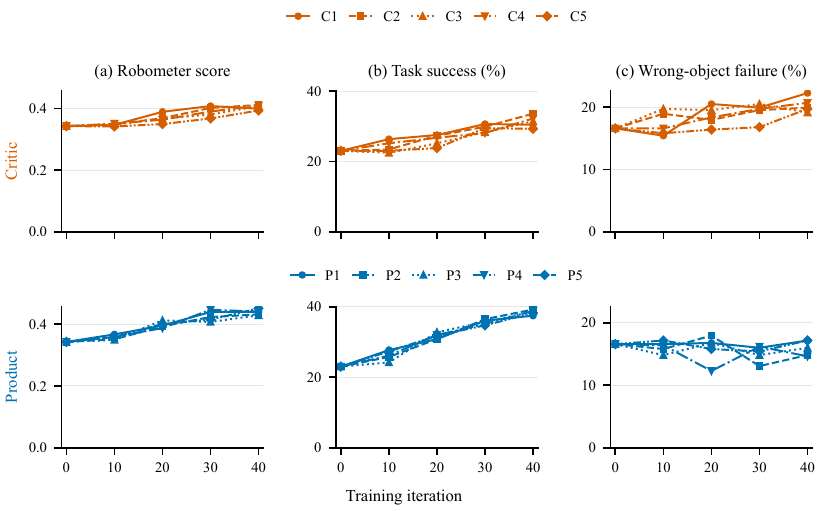}
\caption{\textbf{Individual full-denoiser training histories.}
Robometer score, task success and wrong-object failure on the same fixed
512-seed evaluation bank. C1--C5 are the five critic-only histories;
P1--P5 use critic $\times$ frozen success verifier (Product). Each line
averages one history's 512 executions at each checkpoint; all histories
share the initial policy. Lines connect evaluated checkpoints without
smoothing. The primary endpoint is iteration 40.}
\label{fig:checkpoint_histories_512}
\end{figure}

Figure~\ref{fig:checkpoint_histories_512} reports all five histories per
objective at the five fixed checkpoints. Main-paper
Figure~\ref{fig:checkpoint_dynamics_512} averages those histories and
shows 95\% pointwise percentile intervals from 20,000 crossed-bootstrap
draws. Each draw resamples five histories independently within each
objective and one common set of 512 seed columns. The same history and
seed draws apply across checkpoints, preserving their dependence.
The common initialization has only evaluation-seed uncertainty.
These intervals describe each checkpoint level, rather than simultaneous
coverage or the endpoint contrasts in Table~\ref{tab:whole_network_all}.

\newpage

\subsection{Policy, scorer and evaluation provenance}
\label{app:provenance}

\paragraph{Initial policy.} The shared initial weights are not a purely
supervised policy. They descend from a diffusion policy trained by supervised
learning on middle-drawer demonstrations, followed by three earlier updates
in this project: (i)~a five-epoch warm start on the supervised data mixed with
copies of one optimized episode from an earlier critic study (5.3\% of
samples; that episode pushes the distractor drawer closed without opening the
target, which is not the wrong-object pattern measured here); (ii)~online RL
against RoboReward-8B; and (iii)~600 steps of offline reward-weighted
denoising regression toward Robometer on the evaluation scene, which in its
own evaluation raised wrong-object failures from 1 to 7 of 32. The
whole-network study therefore starts from a policy already exposed to the
critic it optimizes. Every reported change is measured against this common
initialization, and the constrained-selector experiments use the same
decoder.

\paragraph{Scorer.} Robometer-4B at a fixed revision scores eight evenly
spaced frames of a third-person camera with the task text ``open the middle
drawer''; the reward is the success probability at the last frame, and the
base vision-language model is loaded offline at a fixed revision. The
verifier is the frozen model of Section~\ref{sec:verification}. Rescoring 32
stored episodes on a second installation reproduced the stored Robometer
scores, all eight per-frame success probabilities and the verifier outputs
exactly (maximum absolute difference 0).

All primary critic and product evaluations use the same OSMesa renderer,
evaluator and scorer. Weights are fixed at iteration 40; no intermediate
checkpoint is selected by its evaluation outcome. The training sampler is
stochastic DDIM with five steps and standard deviation .05. The artifact
records model identities, evaluation seeds, configurations and score files.

\clearpage
\subsection{Collected-trajectory semantic audit}
\label{app:semantic_audit}

The reward audit of Section~\ref{sec:reward_audit} rescores executions that
were already produced and scored, so it adds no rollouts. The panel contains
every distinct warm-start execution on the 512-seed bank: the shared
initialization once and iterations 10, 20, 30 and 40 of the five critic and
five Product trainings, 20,992 executions from 41 policies with 6,340
successes, 3,636 wrong-object failures and 11,016 other failures. Robometer's
mean score is $.701$ on wrong-object failures against $.651$ on successes and
$.137$ on other failures. Intervals resample evaluation seeds as clusters
(1,000 draws). Among the 16 highest-scored executions, 13 are wrong-object
failures under Robometer and all are successes under the verifier and the
product (Table~\ref{tab:topk_composition}). The inversion is already present
on the initial policy's own 512 executions (AUROC $.365\,[.288,.447]$) and
persists on the final critic ($.401$), Product ($.366$) and verifier-only
($.391$) policies; the supervised initialization's Robometer scores barely
separate the two classes ($.539\,[.449,.626]$), the case discussed in
Section~\ref{sec:tilt}.

\begin{table}[htbp]
\centering\small
\setlength{\tabcolsep}{3pt}
\begin{tabular}{@{}llrrrrrrrrr@{}}
\toprule
& & \multicolumn{3}{c}{Robometer} & \multicolumn{3}{c}{Verifier} & \multicolumn{3}{c}{Product} \\
\cmidrule(lr){3-5}\cmidrule(lr){6-8}\cmidrule(l){9-11}
Panel & $k$ & S & W & O & S & W & O & S & W & O \\
\midrule
All executions & 16 & 3 & 13 & 0 & 16 & 0 & 0 & 16 & 0 & 0 \\
 & 64 & 23 & 41 & 0 & 63 & 1 & 0 & 64 & 0 & 0 \\
 & 256 & 100 & 156 & 0 & 254 & 2 & 0 & 246 & 10 & 0 \\
\midrule
Final policies & 16 & 8 & 8 & 0 & 16 & 0 & 0 & 16 & 0 & 0 \\
 & 64 & 27 & 37 & 0 & 63 & 1 & 0 & 62 & 2 & 0 \\
 & 256 & 113 & 143 & 0 & 255 & 1 & 0 & 243 & 13 & 0 \\
\bottomrule
\end{tabular}
\caption{\textbf{Physical outcomes among the highest-scored executions.} Counts of successes (S), wrong-object failures (W) and other failures (O) among the top $k$ executions ranked by each score. The upper panel uses all 20,992 warm-start executions on the 512-seed bank; the lower panel the 5,120 executions of the ten iteration-40 critic and Product policies.}
\label{tab:topk_composition}
\end{table}

\clearpage
\section{Learnability and Supervised-initialization Controls}
\label{app:essential_controls}

\subsection{Oracle learnability and verifier-only feedback}
\label{app:oracle_controls}

The dense oracle optimizes measured target-completion increments; the
terminal oracle receives the completion once, at the end of the episode;
verifier-only training optimizes the frozen verifier's probability. All are
trained from the warm-start policy with the full-denoiser protocol and evaluated
at iteration 40 on the warm-start 512-seed bank with the renderer, evaluator and
scorer of Table~\ref{tab:whole_network_all} (Table~\ref{tab:weight_endpoints}).
Every oracle improves success by 12 to 19 points and none increases
wrong-object failures, so the policy family learns the task under the
implemented update, including from a single terminal reward. Pooled over the
three dense oracles, success rises by $13.3\,[9.6,17.0]$ points and
wrong-object failures change by $-1.6\,[-4.4,1.4]$; the critic produces
$5.3\,[2.6,8.1]$ points more wrong-object failures than the dense oracles
(exact permutation test over trainings, $p=.018$) and $4.7\,[1.7,7.8]$ more than
the terminal oracle. Six verifier-only trainings, the original one and five
additional runs, raise success by $15.8\,[12.2,19.4]$ points and do not
increase wrong-object failures; Product and verifier-only training are
indistinguishable on both outcomes. Sampler calibration and worker-reset
details are in Appendix~\ref{app:weight_calibration}.

\begin{table}[htbp]
\centering\small
\setlength{\tabcolsep}{4pt}
\begin{tabular}{@{}lr@{\,}lr@{\,}lr@{\,}l@{}}
\toprule
Policy & \multicolumn{2}{c}{$\Delta$ Success (pp)} & \multicolumn{2}{c}{$\Delta$ Wrong-object (pp)} & \multicolumn{2}{c}{$\Delta$ Completion} \\
\midrule
Dense oracle A & +14.5 & {\scriptsize[10.5, 18.4]} & $-$0.2 & {\scriptsize[$-$3.3, 2.9]} & +.127 & {\scriptsize[.092, .161]} \\
Dense oracle B & +13.1 & {\scriptsize[9.4, 16.8]} & $-$2.1 & {\scriptsize[$-$5.1, 0.8]} & +.134 & {\scriptsize[.101, .168]} \\
Dense oracle C & +12.3 & {\scriptsize[8.4, 16.2]} & $-$2.3 & {\scriptsize[$-$5.3, 0.6]} & +.123 & {\scriptsize[.089, .157]} \\
Terminal oracle & +19.1 & {\scriptsize[15.0, 23.2]} & $-$1.0 & {\scriptsize[$-$4.5, 2.3]} & +.176 & {\scriptsize[.140, .213]} \\
Verifier only (V1) & +15.2 & {\scriptsize[11.1, 19.3]} & $-$1.6 & {\scriptsize[$-$4.7, 1.6]} & +.146 & {\scriptsize[.109, .183]} \\
\midrule
Verifier only, six trainings & +15.8 & {\scriptsize[12.2, 19.4]} & $-$1.2 & {\scriptsize[$-$3.8, 1.4]} & +.146 & {\scriptsize[.112, .180]} \\
Product $-$ verifier only & $-$0.3 & {\scriptsize[$-$2.5, 1.8]} & +0.5 & {\scriptsize[$-$1.3, 2.5]} & $-$.006 & {\scriptsize[$-$.029, .017]} \\
\bottomrule
\end{tabular}
\caption{\textbf{Learnability and verifier-only controls on the warm-start 512-seed bank.} Each control is the iteration-40 policy of one training, compared with the shared initialization on the same 512 seeds, renderer and scorer as Table~\ref{tab:whole_network_all}. Single trainings use paired seed intervals; the pooled rows use crossed intervals over trainings and seeds.}
\label{tab:weight_endpoints}
\end{table}

\newpage
\subsection{Amplification from a capable supervised initialization}
\label{app:capable_supervised}

\paragraph{Policy and task.}
We use a supervised diffusion policy further trained for ten epochs on 167
physically validated drawer demonstrations, with its own visual encoder and
normalizer. No learned reward or reinforcement learning enters this policy's
training lineage. Before RL, a fixed capability screen evaluates this
checkpoint on twelve drawer layouts with eight seeds each. Layouts with
2--6 successes are eligible; we select the most successful, breaking ties by
manifest order. The selected layout has 3/8 successes. It differs from the
warm-start study's layout and from the ceiling-limited control in
Appendix~\ref{app:supervised_only}.

\paragraph{Training and evaluation.}
Five independent trainings per objective optimize either Robometer or
terminal target completion, with the same supervised initialization,
full-denoiser PPO update, 40 iterations and 126 feedback episodes per
iteration. Completion is the final target opening divided by 54\,mm and
clipped to $[0,1]$; a wrong-object failure requires an unsuccessful task and
more than 20\,mm additional distractor opening. The endpoint is fixed at
iteration 40, and all ten final policies and the common initialization are
evaluated on the same 512 held-out seeds. An independent repetition of three
trainings per objective, run with fresh training seeds on a second host, is
evaluated on the same seeds (Table~\ref{tab:clean_initialization_all}).

All five Robometer trainings individually amplify wrong-object failures, by
9.2 to 13.5 points, while the five completion trainings change them by $-0.4$
to $+4.1$ points. Pooled, Robometer minus completion gives $+9.2\,[5.6,13.0]$
points of wrong-object failure and $-3.7\,[-7.5,0.1]$ points of success.
Across all eight trainings per objective, the wrong-object difference is
$+8.3\,[5.0,11.5]$ points; exact permutation tests over training labels give
$p=.008$ for five against five and $p=.0002$ for eight against eight.
Pooled intervals use 20,000 crossed bootstrap draws over trainings and one
shared set of evaluation seeds; cross-objective contrasts resample the two
objectives' trainings independently and share the seed resample.

\begin{table}[htbp]
\centering\small
\setlength{\tabcolsep}{4pt}
\begin{tabular}{@{}lr@{\,}lr@{\,}lr@{\,}l@{}}
\toprule
Training & \multicolumn{2}{c}{$\Delta$ Success (pp)} & \multicolumn{2}{c}{$\Delta$ Wrong-object (pp)} & \multicolumn{2}{c}{$\Delta$ Robometer} \\
\midrule
Robometer 1 & +9.4 & {\scriptsize[5.5, 13.3]} & +13.5 & {\scriptsize[9.4, 17.6]} & +.123 & {\scriptsize[.092, .153]} \\
Robometer 2 & +9.8 & {\scriptsize[6.1, 13.5]} & +10.2 & {\scriptsize[6.4, 14.1]} & +.167 & {\scriptsize[.137, .198]} \\
Robometer 3 & +9.6 & {\scriptsize[5.9, 13.5]} & +11.9 & {\scriptsize[8.2, 15.8]} & +.129 & {\scriptsize[.098, .160]} \\
Robometer 4 & +11.7 & {\scriptsize[7.6, 15.8]} & +9.6 & {\scriptsize[5.7, 13.7]} & +.138 & {\scriptsize[.105, .171]} \\
Robometer 5 & +10.7 & {\scriptsize[6.6, 14.6]} & +9.2 & {\scriptsize[5.3, 13.1]} & +.130 & {\scriptsize[.099, .162]} \\
\midrule
Completion 1 & +17.6 & {\scriptsize[12.9, 22.3]} & +3.9 & {\scriptsize[0.4, 7.4]} & +.094 & {\scriptsize[.061, .126]} \\
Completion 2 & +11.7 & {\scriptsize[7.4, 16.0]} & +0.0 & {\scriptsize[$-$3.3, 3.3]} & +.046 & {\scriptsize[.014, .079]} \\
Completion 3 & +12.9 & {\scriptsize[8.6, 17.2]} & $-$0.4 & {\scriptsize[$-$3.9, 3.1]} & +.042 & {\scriptsize[.009, .075]} \\
Completion 4 & +13.5 & {\scriptsize[9.0, 18.0]} & +0.6 & {\scriptsize[$-$2.7, 3.9]} & +.064 & {\scriptsize[.031, .097]} \\
Completion 5 & +13.9 & {\scriptsize[9.4, 18.4]} & +4.1 & {\scriptsize[0.4, 7.8]} & +.065 & {\scriptsize[.031, .099]} \\
\midrule
Robometer 1 (repetition) & +8.8 & {\scriptsize[4.9, 12.7]} & +11.7 & {\scriptsize[8.0, 15.4]} & +.120 & {\scriptsize[.089, .150]} \\
Robometer 2 (repetition) & +11.3 & {\scriptsize[7.6, 15.2]} & +7.4 & {\scriptsize[3.7, 11.1]} & +.127 & {\scriptsize[.096, .157]} \\
Robometer 3 (repetition) & +8.4 & {\scriptsize[4.5, 12.5]} & +7.8 & {\scriptsize[4.3, 11.5]} & +.056 & {\scriptsize[.026, .086]} \\
Completion 1 (repetition) & +15.0 & {\scriptsize[10.5, 19.5]} & +3.9 & {\scriptsize[0.2, 7.6]} & +.069 & {\scriptsize[.036, .101]} \\
Completion 2 (repetition) & +14.5 & {\scriptsize[10.2, 18.8]} & +3.1 & {\scriptsize[$-$0.4, 6.4]} & +.084 & {\scriptsize[.053, .114]} \\
Completion 3 (repetition) & +16.8 & {\scriptsize[12.5, 21.1]} & +0.0 & {\scriptsize[$-$3.3, 3.3]} & +.075 & {\scriptsize[.044, .107]} \\
\midrule
Robometer, all eight & +10.0 & {\scriptsize[6.8, 13.2]} & +10.2 & {\scriptsize[6.8, 13.6]} & +.124 & {\scriptsize[.091, .155]} \\
Completion, all eight & +14.5 & {\scriptsize[10.4, 18.6]} & +1.9 & {\scriptsize[$-$1.4, 5.1]} & +.067 & {\scriptsize[.037, .097]} \\
Robometer $-$ completion, eight each & $-$4.5 & {\scriptsize[$-$7.9, $-$1.2]} & +8.3 & {\scriptsize[5.0, 11.5]} & +.056 & {\scriptsize[.025, .087]} \\
\bottomrule
\end{tabular}
\caption{\textbf{All sixteen trainings from the supervised initialization on its 512-seed bank.} Five trainings per reward form the primary comparison (Table~\ref{tab:whole_network_all}); an independent repetition of three trainings per reward was evaluated on the same bank. Single trainings use paired seed intervals; pooled rows use crossed intervals, and the contrast resamples the two rewards independently.}
\label{tab:clean_initialization_all}
\end{table}

\clearpage
\section{Constrained-policy Recurrence and Matched Verification}
\label{app:constrained_evidence}

\begin{figure}[htbp]
\centering
\includegraphics[width=\linewidth]{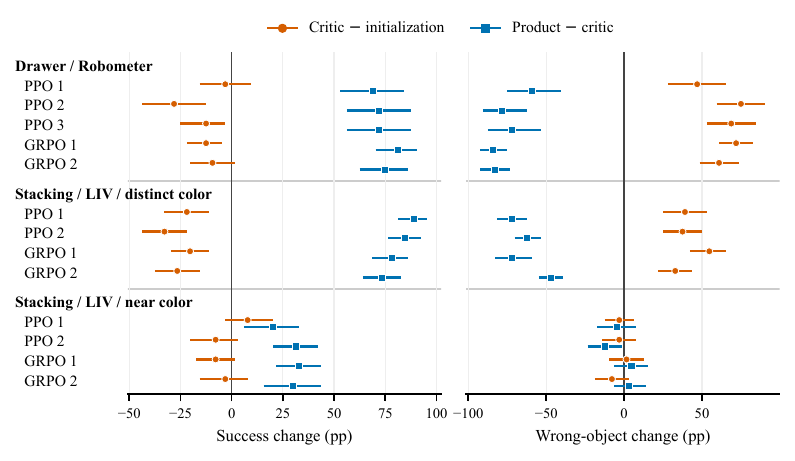}
\caption{\textbf{Semantic amplification and the verification intervention.}
Critic minus initialization (orange circles) and product minus critic
(blue squares), with 95\% paired execution intervals in percentage points.
Rows are grouped by task, critic and color condition. These 13 strata contain all nine repeated-error
trainings and both trained stacking colors; the near-color strata are retained
even when induction is unresolved. All 29 strata, including Robometer stacking
and drawer LIV boundaries, appear in Figure~\ref{fig:selector_forest_all}.
Training numbers identify histories within each task--critic--optimizer group; the two stacking colors share histories. Original IDs and complete counts are in Supplementary Tables~\ref{tab:reader_run_ids} and~\ref{tab:repair_endpoints}.}
\label{fig:selector_forest}
\end{figure}

Figure~\ref{fig:selector_forest} contains the repeated drawer Robometer
and LIV stacking comparisons, including both trained stacking colors.
The tables below report all nine repeated-error histories across these
settings. The complete 29-stratum grid, all supervision budgets and
boundary settings appear in Appendix~\ref{supp:grid}.
Training numbers are local to each task--critic--optimizer group. Original
run IDs are retained in Table~\ref{tab:reader_run_ids}.

\newpage
\subsection{Oracle, Exploitation and Repair}
\label{app:rm_ppo_repeat}
\label{app:drawer_grpo_endpoints}
\label{app:drawer_grpo_replication}
\label{sec:balanced_liv_grpo}
\label{sec:balanced_liv_ppo}
\label{sec:independent_liv_grpo_near}
\label{sec:independent_liv_ppo_near}

Tables~\ref{tab:claim_chain_absolute} and~\ref{tab:claim_chain_intervals}
report original critic scores, physical task successes and paired changes
for all nine histories.
All nine oracle policies improve true success with intervals above zero.
For LIV stacking, their native-score changes are negative in all four
trainings, also with intervals excluding zero. The oracle learns the
physical task despite this critic disagreement. Conversely, original-LIV
optimization raises native score while losing success in all four.

\begin{table}[htbp]
\centering\small
\begin{tabular*}{\linewidth}{@{\extracolsep{\fill}}lrrrrr@{}}
\toprule
Training & $n$ & Initialization & Oracle & Critic & Product \\
\midrule
\multicolumn{6}{l}{\textbf{(a) Original critic score}} \\
\multicolumn{6}{l}{\textit{Drawer / Robometer}} \\
\quad PPO 1 & 32 & 0.4050 & 0.5517 & 0.7141 & 0.6594 \\
\quad PPO 2 & 32 & 0.3903 & 0.4817 & 0.8102 & 0.6249 \\
\quad PPO 3 & 32 & 0.3393 & 0.4273 & 0.7709 & 0.6456 \\
\quad GRPO 1 & 64 & 0.3008 & 0.6152 & 0.7937 & 0.6314 \\
\quad GRPO 2 & 64 & 0.3609 & 0.5946 & 0.7803 & 0.6120 \\
\multicolumn{6}{l}{\textit{Stacking / LIV / distinct color}} \\
\quad PPO 1 & 64 & $-$0.1700 & $-$0.1784 & $-$0.1631 & $-$0.1784 \\
\quad PPO 2 & 64 & $-$0.1710 & $-$0.1762 & $-$0.1625 & $-$0.1764 \\
\quad GRPO 1 & 64 & $-$0.1693 & $-$0.1757 & $-$0.1627 & $-$0.1750 \\
\quad GRPO 2 & 64 & $-$0.1707 & $-$0.1765 & $-$0.1637 & $-$0.1754 \\
\midrule
\multicolumn{6}{l}{\textbf{(b) Task successes (count)}} \\
\multicolumn{6}{l}{\textit{Drawer / Robometer}} \\
\quad PPO 1 & 32 & 4 & 24 & 3 & 25 \\
\quad PPO 2 & 32 & 11 & 19 & 2 & 25 \\
\quad PPO 3 & 32 & 6 & 18 & 2 & 25 \\
\quad GRPO 1 & 64 & 12 & 54 & 4 & 56 \\
\quad GRPO 2 & 64 & 10 & 58 & 4 & 52 \\
\multicolumn{6}{l}{\textit{Stacking / LIV / distinct color}} \\
\quad PPO 1 & 64 & 20 & 63 & 6 & 63 \\
\quad PPO 2 & 64 & 24 & 57 & 3 & 57 \\
\quad GRPO 1 & 64 & 15 & 55 & 2 & 52 \\
\quad GRPO 2 & 64 & 22 & 51 & 5 & 52 \\
\midrule
\multicolumn{6}{l}{\textbf{(c) Wrong-object failures (count)}} \\
\multicolumn{6}{l}{\textit{Drawer / Robometer}} \\
\quad PPO 1 & 32 & 9 & 2 & 24 & 5 \\
\quad PPO 2 & 32 & 4 & 3 & 28 & 3 \\
\quad PPO 3 & 32 & 6 & 4 & 28 & 5 \\
\quad GRPO 1 & 64 & 8 & 1 & 54 & 0 \\
\quad GRPO 2 & 64 & 14 & 0 & 53 & 0 \\
\multicolumn{6}{l}{\textit{Stacking / LIV / distinct color}} \\
\quad PPO 1 & 64 & 21 & 0 & 46 & 0 \\
\quad PPO 2 & 64 & 18 & 2 & 42 & 2 \\
\quad GRPO 1 & 64 & 18 & 6 & 53 & 7 \\
\quad GRPO 2 & 64 & 15 & 9 & 36 & 6 \\
\bottomrule
\end{tabular*}
\caption{\textbf{Oracle, exploitation and repair across nine training histories.} Panel (a) reports mean original critic scores: Robometer success probability for drawer and native LIV cosine for stacking. Panels (b) and (c) report physical task successes and wrong-object failures out of $n$ evaluation executions per policy. Initialization is the uniform noise-selection policy; Oracle, Critic and Product optimize physical completion, the visual critic, and critic $\times$ frozen verifier, respectively. Oracle policies are evaluated with the same visual critic as the other policies. Stacking uses the 2,536-label verifier. Original run IDs are mapped in Table~\ref{tab:reader_run_ids}.}
\label{tab:claim_chain_absolute}
\end{table}

Drawer PPO 2 and 3 and GRPO 1 also meet the strict reward-up,
success-down, repaired-success-up criterion. Drawer PPO 1 and GRPO 2 have
success-loss intervals crossing zero, but reliably increase wrong-object
failures while increasing reward. Drawer PPO 1 additionally reduces mean clipped
target completion from .2576 to .1082, a paired change of
$-.1494\,[-.2917,-.0137]$. The oracle and product reach
.9146 and .8300.

\begin{table}[htbp]
\centering\small
\setlength{\tabcolsep}{4pt}
\begin{tabular}{@{}lrr@{\,}lr@{\,}lr@{\,}lr@{\,}l@{}}
\toprule
 & & \multicolumn{2}{c}{Reward change} & \multicolumn{6}{c}{Task-success change (count of $n$)} \\
\cmidrule(lr){3-4}\cmidrule(l){5-10}
Training & $n$ & \multicolumn{2}{c}{Critic $-$ Init.} & \multicolumn{2}{c}{Oracle $-$ Init.} & \multicolumn{2}{c}{Critic $-$ Init.} & \multicolumn{2}{c}{Product $-$ Critic} \\
\midrule
\multicolumn{10}{@{}l}{\textit{Drawer / Robometer}} \\
\quad PPO 1 & 32 & $+$.3091 & {\scriptsize[$+$.1915, $+$.4252]} & $+$20 & {\scriptsize[$+$14, $+$26]} & $-$1 & {\scriptsize[$-$5, $+$3]} & $+$22 & {\scriptsize[$+$17, $+$27]} \\
\quad PPO 2 & 32 & $+$.4199 & {\scriptsize[$+$.3139, $+$.5218]} & $+$8 & {\scriptsize[$+$3, $+$14]} & $-$9 & {\scriptsize[$-$14, $-$4]} & $+$23 & {\scriptsize[$+$18, $+$28]} \\
\quad PPO 3 & 32 & $+$.4316 & {\scriptsize[$+$.3184, $+$.5410]} & $+$12 & {\scriptsize[$+$6, $+$18]} & $-$4 & {\scriptsize[$-$8, $-$1]} & $+$23 & {\scriptsize[$+$18, $+$28]} \\
\quad GRPO 1 & 64 & $+$.4929 & {\scriptsize[$+$.4252, $+$.5601]} & $+$42 & {\scriptsize[$+$34, $+$50]} & $-$8 & {\scriptsize[$-$14, $-$3]} & $+$52 & {\scriptsize[$+$45, $+$58]} \\
\quad GRPO 2 & 64 & $+$.4195 & {\scriptsize[$+$.3398, $+$.4970]} & $+$48 & {\scriptsize[$+$41, $+$54]} & $-$6 & {\scriptsize[$-$13, $+$1]} & $+$48 & {\scriptsize[$+$40, $+$55]} \\
\multicolumn{10}{@{}l}{\textit{Stacking / LIV / distinct color}} \\
\quad PPO 1 & 64 & $+$.0070 & {\scriptsize[$+$.0047, $+$.0093]} & $+$43 & {\scriptsize[$+$36, $+$50]} & $-$14 & {\scriptsize[$-$21, $-$7]} & $+$57 & {\scriptsize[$+$52, $+$61]} \\
\quad PPO 2 & 64 & $+$.0085 & {\scriptsize[$+$.0065, $+$.0105]} & $+$33 & {\scriptsize[$+$25, $+$41]} & $-$21 & {\scriptsize[$-$28, $-$14]} & $+$54 & {\scriptsize[$+$49, $+$59]} \\
\quad GRPO 1 & 64 & $+$.0067 & {\scriptsize[$+$.0047, $+$.0086]} & $+$40 & {\scriptsize[$+$32, $+$48]} & $-$13 & {\scriptsize[$-$19, $-$7]} & $+$50 & {\scriptsize[$+$44, $+$55]} \\
\quad GRPO 2 & 64 & $+$.0070 & {\scriptsize[$+$.0051, $+$.0091]} & $+$29 & {\scriptsize[$+$22, $+$36]} & $-$17 & {\scriptsize[$-$24, $-$10]} & $+$47 & {\scriptsize[$+$41, $+$53]} \\
\bottomrule
\end{tabular}
\caption{\textbf{Oracle learnability, reward exploitation and repair: paired changes.} Each row is one training history; entries are paired differences with 95\% intervals in brackets. Init.\ is the uniform noise-selection policy. The reward change is the change in mean original critic score under critic-only training: Robometer success probability for drawer and native LIV cosine for stacking. Success changes are counts out of $n$ evaluation executions. Seven of nine trainings have a success-loss interval strictly below zero; drawer PPO 1 and GRPO 2 establish amplified wrong-object behavior but not a reliable binary-success loss. Native reward and repair success increase in all nine.}
\label{tab:claim_chain_intervals}
\end{table}

Continuation from the three exploited drawer policies, with product,
original-critic, verifier-only and oracle rewards and a drawer opening held
out from continuation, is reported in
Appendix~\ref{app:checkpoint_continuation}.

\newpage
\subsection{Verifier-only Ablation}
\label{app:verifier_only_ablation}

The ablation removes the original critic and optimizes $R_V=p_\theta$
with the identical verifier, labels, policy capacity and update protocol.
Supplemental same-head controls retain the original seeds and repeat
uniform executions to check reference reuse. These paired ablations reuse
the original training histories. Table~\ref{tab:all_final_budgets}
retains every available matched control and marks missing controls explicitly.

\paragraph{Source of the hacking repairs.}
For drawer Robometer GRPO, verifier-only reaches 55 and 53/64 successes
with zero wrong-object errors, compared with product outcomes 56/0 and
52/0. Product-minus-only success differences are $+1[0,3]/64$ and
$-1[-3,0]/64$. For distinct-color LIV stacking, expanded-only, the
verifier-only control with the 2,536-label head, reaches
63/57 successes under PPO and 50/55 under GRPO; the product reaches
63/57 and 52/52. Neither optimizer establishes an additional physical
benefit from retaining LIV. In the second PPO training, product and only
have identical outcomes on every paired seed. These controls attribute
much of the repair to outcome verification.

\paragraph{Fixed support and replication.}
The repeated trainings vary training and evaluation seeds on fixed
24-component development supports; they do not resample support banks.
The drawer support was selected during development without an outcome-blind
bank-selection rule. Both stacking colors participate in training.
Intervals condition on the fitted policies and selected support.
Appendix~\ref{app:identity_head_capacity} specifies the label costs and
shared heads; Appendix~\ref{sec:categorical_ppo} gives the exact selector
representation, objectives and optimizer settings.

\paragraph{Temporal verification in ball placement.}
The controlled ball-placement extension compares full-history and
final-quarter verifiers in a matched PPO ablation at equal label and frame
budgets. In two independent histories, the product with the full-history
verifier reaches 63 and 64/64 successes on new seeds, against 12/64 for the
uniform selector and 2 and 5/64 for the original reward; the same product
with a final-quarter verifier reaches 17 and 2/64. Its protocol and
new-seed endpoints are in
Appendix~\ref{app:temporal_verifier} and
Table~\ref{tab:ballot-online}.

\clearpage
\section{Sim-to-Real Case Study: Protocol and Additional Results}
\label{app:hardware}
\label{app:pending_validation}
\label{app:pending_p3}

\begin{figure}[htbp]
\centering
\includegraphics[width=0.6\linewidth]{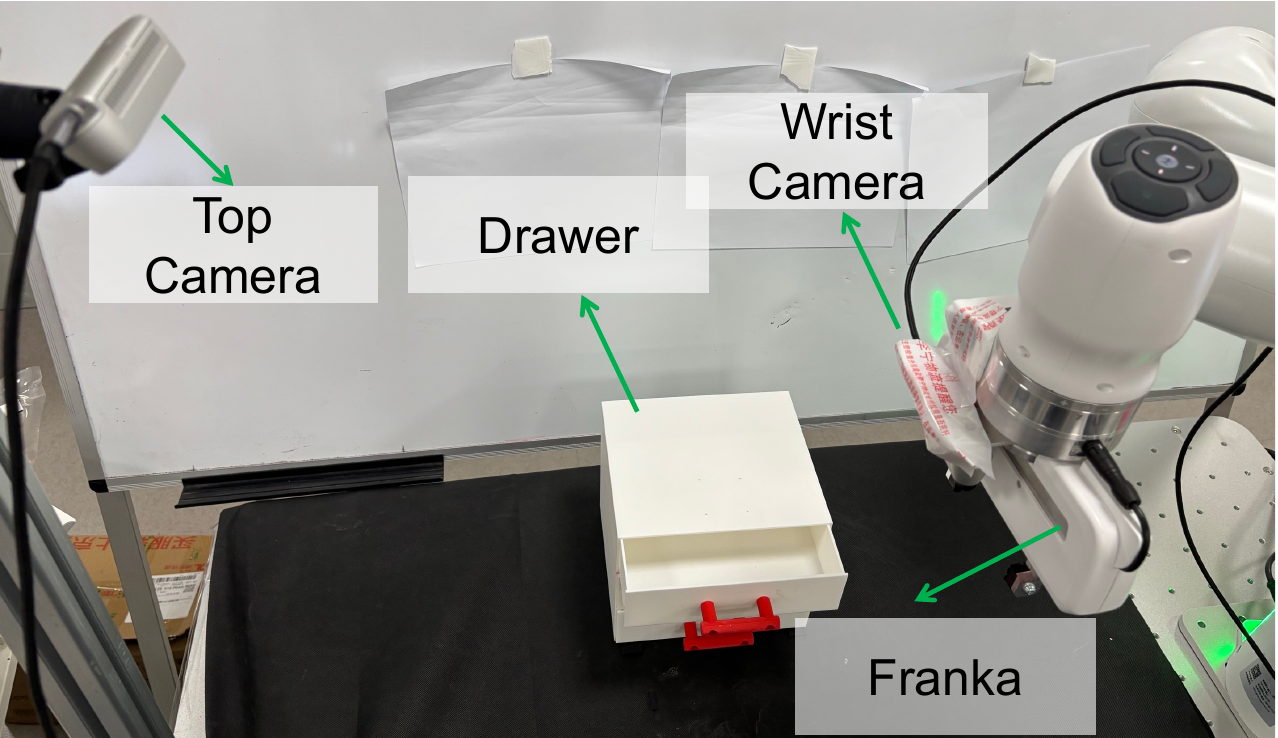}
\caption{Real-robot deployment}
\label{fig:real}
\end{figure}

The main hardware outcomes are reported in
Section~\ref{sec:real_robot}. Here, we provide the physical setup,
deployed policy identities, and outcome definitions for the
real-robot drawer evaluation. Fig. \ref{fig:real} shows the environment setup for deployment on a Franka robot.

\paragraph{Physical setup and reset procedure.}
Experiments use a Franka Research 3 robot in a physical drawer setup
matched to the simulation at 1:1 scale, including cabinet placement,
drawer geometry, initial drawer openings, and camera placement.
The requested task is to open the middle drawer. Before each trial,
the target middle drawer is reset to its closed position, and the
upper distractor drawer is returned to the same marked partial
opening. All policies are evaluated under the same physical setup
and deployment configuration.

\paragraph{Policy identities and checkpoint mapping.}
The hardware comparison includes three conditions:
\emph{Initialization}, \emph{Critic-trained} and
\emph{Completion-trained}. Initialization is the shared warm-start
initialization of the full-denoiser study (Appendix~\ref{app:provenance}),
which has prior reward exposure. Critic-trained is Critic 5 at iteration 40,
chosen after the simulation evaluations without a prespecified quantitative
selection rule; the checkpoint was fixed before any physical trials.
No hardware outcomes were used in the choice.
Among the five critic trainings, Critic 5 has the second-largest
wrong-object increase on the 128-seed checkpoint-diagnostic panel of
Section~\ref{sec:reward_audit} (per-training values in
Appendix~\ref{supp:execution}) and the
smallest success gain on the 512-seed primary panel
(Figure~\ref{fig:unet_endpoints}).
Completion-trained is the terminal-completion oracle of
Appendix~\ref{app:oracle_controls} at iteration 40, trained from the same
initialization. All policies are deployed with the five-step DDIM sampler
used during fine-tuning. Because only one training history per fine-tuned
condition was deployed, the hardware study is a sim-to-real case study
rather than an estimate of transfer across training histories.

\newpage
\paragraph{Deployment and evaluation.}
Each policy executes 40 trials, giving 120 trials across the three
conditions. Policy parameters remain fixed throughout hardware
evaluation, with no training or fine-tuning on the physical robot.
The evaluation therefore examines the physical execution of
policies trained in simulation. Task outcomes are assessed from
physical drawer motion independently of the learned visual reward.

\paragraph{Outcome definitions.}
Each trial is assigned to one of three mutually exclusive categories.
A \emph{success} requires opening the target middle drawer by at
least 6\,cm. A \emph{wrong-object failure} occurs when the upper
distractor drawer is pulled more than 2\,cm beyond its initial
marked opening while the requested task remains unsuccessful.
All remaining unsuccessful trials are classified as
\emph{other failures}. Grasping a drawer handle alone does not
satisfy the success criterion.

Table~\ref{tab:real_robot_counts} gives the outcome rates and their changes
from Initialization for the hardware comparison in Section~\ref{sec:real_robot}.

\begin{table}[htbp]
\centering\small
\setlength{\tabcolsep}{3.5pt}
\begin{tabular}{@{}lr@{\,}lr@{\,}lr@{\,}lr@{\,}lr@{\,}l@{}}
\toprule
 & \multicolumn{6}{c}{Outcome rate (\%)} & \multicolumn{4}{c}{Change from Initialization (pp)} \\
\cmidrule(lr){2-7}\cmidrule(l){8-11}
Policy & \multicolumn{2}{c}{Success} & \multicolumn{2}{c}{Wrong-object} & \multicolumn{2}{c}{Other} & \multicolumn{2}{c}{Success} & \multicolumn{2}{c}{Wrong-object} \\
\midrule
Initialization & 30.0 & {\scriptsize[18.1, 45.4]} & 22.5 & {\scriptsize[12.3, 37.5]} & 47.5 & {\scriptsize[32.9, 62.5]} & \multicolumn{2}{c}{--} & \multicolumn{2}{c}{--} \\
Critic-trained & 25.0 & {\scriptsize[14.2, 40.2]} & 37.5 & {\scriptsize[24.2, 53.0]} & 37.5 & {\scriptsize[24.2, 53.0]} & $-$5.0 & {\scriptsize[$-$23.8, $+$14.3]} & $+$15.0 & {\scriptsize[$-$5.0, $+$33.5]} \\
Completion-trained & 40.0 & {\scriptsize[26.3, 55.4]} & 17.5 & {\scriptsize[8.7, 31.9]} & 42.5 & {\scriptsize[28.5, 57.8]} & $+$10.0 & {\scriptsize[$-$10.6, $+$29.5]} & $-$5.0 & {\scriptsize[$-$22.4, $+$12.7]} \\
\bottomrule
\end{tabular}
\caption{\textbf{Real-robot outcomes.} Each policy executes 40 physical drawer trials. Rates are percentages of trials with Wilson 95\% intervals; changes relative to Initialization are in percentage points with Newcombe 95\% intervals. No change is statistically resolved (two-sided Fisher exact $p=.80$ and $.22$ for Critic-trained success and wrong-object, $.48$ and $.78$ for Completion-trained). Success opens the requested middle drawer by at least 6\,cm; a wrong-object failure pulls the upper distractor more than 2\,cm beyond its initial opening without success. Initialization is the shared warm-start initialization of the full-denoiser study; Critic-trained is Critic 5 at iteration 40; Completion-trained is the terminal-completion oracle of Appendix~\ref{app:oracle_controls} at iteration 40.}
\label{tab:real_robot_counts}
\end{table}

\clearpage
\section{Scope and Retrospective Analyses}
\label{app:robustness}

\subsection{Prevalence across the constrained grid}
\label{app:prevalence}

A constrained-grid stratum shows wrong-object amplification if and only if
the paired 95\% interval for the critic-minus-uniform change in
wrong-object failure has a lower bound above zero. Under this definition,
11 of the 29 evaluated strata (37.9\%) show amplification
(Table~\ref{tab:prevalence}). The 29 strata come from 20 training
histories, because the two stacking colors of one training share its
history; they are not 29 independent trainings. In all 11 amplified strata,
the critic-minus-uniform change in original reward also has an interval
lower bound above zero. One drawer LIV stratum has a wrong-object interval
whose lower bound is exactly zero and is not counted. In the two amplified
near-color Robometer stacking strata, amplification occurs together with a
positive success change. The nine repeated-error histories of
Section~\ref{sec:ppo_results} come from the two settings with amplification
in every stratum and do not estimate prevalence.
Table~\ref{tab:prevalence_strata} lists every stratum with its intervals;
Appendix~\ref{supp:grid} adds per-policy outcome counts and all
supervision budgets, and separate RoboReward experiments outside the grid
are in Appendix~\ref{supp:additional}.

\begin{table}[htbp]
\centering\small
\begin{tabular}{llr}
\toprule
Task & Critic, decoy color & Amplified strata \\
\midrule
Drawer & Robometer & 5/5 \\
Stacking & LIV, distinct color & 4/4 \\
Stacking & LIV, near color & 0/4 \\
Stacking & Robometer, near color & 2/5 \\
Stacking & Robometer, distinct color & 0/5 \\
Drawer & LIV & 0/6 \\
\midrule
All & & 11/29 \\
\bottomrule
\end{tabular}
\caption{\textbf{Wrong-object amplification across the constrained grid.}
A stratum counts as amplified when the paired 95\% interval for the
critic-minus-uniform change in wrong-object failure lies above zero. The
29 strata come from 20 training histories; the two stacking colors of a
training share its history.}
\label{tab:prevalence}
\end{table}

\begin{table}[htbp]
\centering\small
\setlength{\tabcolsep}{4pt}
\begin{tabular}{@{}llrr@{\,}lr@{\,}lr@{\,}lc@{}}
\toprule
 & & & \multicolumn{4}{c}{Change in count of $n$} & & & \\
\cmidrule(lr){4-7}
Training & Decoy & $n$ & \multicolumn{2}{c}{Wrong-object} & \multicolumn{2}{c}{Success} & \multicolumn{2}{c}{Critic score} & Amplified \\
\midrule
\multicolumn{10}{@{}l}{\textit{Drawer / Robometer}} \\
\quad PPO 1 & -- & 32 & $+$15 & {\scriptsize[$+$9, $+$21]} & $-$1 & {\scriptsize[$-$5, $+$3]} & $+$.3091 & {\scriptsize[$+$.1915, $+$.4252]} & \checkmark \\
\quad PPO 2 & -- & 32 & $+$24 & {\scriptsize[$+$19, $+$29]} & $-$9 & {\scriptsize[$-$14, $-$4]} & $+$.4199 & {\scriptsize[$+$.3139, $+$.5218]} & \checkmark \\
\quad PPO 3 & -- & 32 & $+$22 & {\scriptsize[$+$17, $+$27]} & $-$4 & {\scriptsize[$-$8, $-$1]} & $+$.4316 & {\scriptsize[$+$.3184, $+$.5410]} & \checkmark \\
\quad GRPO 1 & -- & 64 & $+$46 & {\scriptsize[$+$39, $+$53]} & $-$8 & {\scriptsize[$-$14, $-$3]} & $+$.4929 & {\scriptsize[$+$.4252, $+$.5601]} & \checkmark \\
\quad GRPO 2 & -- & 64 & $+$39 & {\scriptsize[$+$31, $+$47]} & $-$6 & {\scriptsize[$-$13, $+$1]} & $+$.4195 & {\scriptsize[$+$.3398, $+$.4970]} & \checkmark \\
\multicolumn{10}{@{}l}{\textit{Stacking / LIV}} \\
\quad PPO 1 & near & 64 & $-$2 & {\scriptsize[$-$8, $+$4]} & $+$5 & {\scriptsize[$-$2, $+$13]} & $+$.0007 & {\scriptsize[$-$.0006, $+$.0021]} &  \\
\quad PPO 1 & distinct & 64 & $+$25 & {\scriptsize[$+$16, $+$34]} & $-$14 & {\scriptsize[$-$21, $-$7]} & $+$.0070 & {\scriptsize[$+$.0047, $+$.0093]} & \checkmark \\
\quad PPO 2 & near & 64 & $-$2 & {\scriptsize[$-$9, $+$5]} & $-$5 & {\scriptsize[$-$13, $+$2]} & $+$.0047 & {\scriptsize[$+$.0031, $+$.0064]} &  \\
\quad PPO 2 & distinct & 64 & $+$24 & {\scriptsize[$+$16, $+$32]} & $-$21 & {\scriptsize[$-$28, $-$14]} & $+$.0085 & {\scriptsize[$+$.0065, $+$.0105]} & \checkmark \\
\quad GRPO 1 & near & 64 & $+$1 & {\scriptsize[$-$6, $+$8]} & $-$5 & {\scriptsize[$-$11, $+$1]} & $+$.0012 & {\scriptsize[$-$.0002, $+$.0026]} &  \\
\quad GRPO 1 & distinct & 64 & $+$35 & {\scriptsize[$+$27, $+$42]} & $-$13 & {\scriptsize[$-$19, $-$7]} & $+$.0067 & {\scriptsize[$+$.0047, $+$.0086]} & \checkmark \\
\quad GRPO 2 & near & 64 & $-$5 & {\scriptsize[$-$12, $+$2]} & $-$2 & {\scriptsize[$-$10, $+$5]} & $+$.0025 & {\scriptsize[$+$.0009, $+$.0042]} &  \\
\quad GRPO 2 & distinct & 64 & $+$21 & {\scriptsize[$+$14, $+$28]} & $-$17 & {\scriptsize[$-$24, $-$10]} & $+$.0070 & {\scriptsize[$+$.0051, $+$.0091]} & \checkmark \\
\multicolumn{10}{@{}l}{\textit{Stacking / Robometer}} \\
\quad PPO 1 & near & 64 & $+$13 & {\scriptsize[$+$5, $+$21]} & $+$6 & {\scriptsize[$-$2, $+$14]} & $+$.1770 & {\scriptsize[$+$.1047, $+$.2491]} & \checkmark \\
\quad PPO 1 & distinct & 64 & $-$16 & {\scriptsize[$-$22, $-$10]} & $+$40 & {\scriptsize[$+$33, $+$47]} & $+$.4278 & {\scriptsize[$+$.3518, $+$.5030]} &  \\
\quad PPO 2 & near & 64 & $+$2 & {\scriptsize[$-$6, $+$10]} & $+$10 & {\scriptsize[$+$3, $+$17]} & $+$.1361 & {\scriptsize[$+$.0725, $+$.2002]} &  \\
\quad PPO 2 & distinct & 64 & $-$19 & {\scriptsize[$-$26, $-$12]} & $+$34 & {\scriptsize[$+$27, $+$41]} & $+$.3450 & {\scriptsize[$+$.2654, $+$.4233]} &  \\
\quad PPO 3 & near & 64 & $+$13 & {\scriptsize[$+$5, $+$21]} & $+$6 & {\scriptsize[$-$1, $+$13]} & $+$.1448 & {\scriptsize[$+$.0828, $+$.2109]} & \checkmark \\
\quad PPO 3 & distinct & 64 & $-$12 & {\scriptsize[$-$19, $-$6]} & $+$35 & {\scriptsize[$+$28, $+$43]} & $+$.4290 & {\scriptsize[$+$.3485, $+$.5098]} &  \\
\quad GRPO 1 & near & 64 & $+$7 & {\scriptsize[$-$1, $+$15]} & $+$12 & {\scriptsize[$+$5, $+$19]} & $+$.1674 & {\scriptsize[$+$.1040, $+$.2322]} &  \\
\quad GRPO 1 & distinct & 64 & $-$19 & {\scriptsize[$-$26, $-$12]} & $+$41 & {\scriptsize[$+$33, $+$48]} & $+$.4345 & {\scriptsize[$+$.3538, $+$.5102]} &  \\
\quad GRPO 2 & near & 64 & $+$2 & {\scriptsize[$-$5, $+$9]} & $+$19 & {\scriptsize[$+$13, $+$25]} & $+$.2655 & {\scriptsize[$+$.2036, $+$.3295]} &  \\
\quad GRPO 2 & distinct & 64 & $-$4 & {\scriptsize[$-$11, $+$3]} & $+$31 & {\scriptsize[$+$25, $+$37]} & $+$.3914 & {\scriptsize[$+$.3178, $+$.4640]} &  \\
\multicolumn{10}{@{}l}{\textit{Drawer / LIV}} \\
\quad PPO 1 & -- & 64 & $-$1 & {\scriptsize[$-$5, $+$3]} & $+$3 & {\scriptsize[$-$1, $+$7]} & $+$.0023 & {\scriptsize[$+$.0002, $+$.0049]} &  \\
\quad PPO 2 & -- & 64 & $+$2 & {\scriptsize[$+$0, $+$5]} & $-$1 & {\scriptsize[$-$3, $+$0]} & $-$.0001 & {\scriptsize[$-$.0008, $+$.0006]} &  \\
\quad PPO 3 & -- & 64 & $-$10 & {\scriptsize[$-$16, $-$4]} & $-$7 & {\scriptsize[$-$13, $-$1]} & $+$.0222 & {\scriptsize[$+$.0194, $+$.0250]} &  \\
\quad PPO 4 & -- & 64 & $-$5 & {\scriptsize[$-$10, $-$1]} & $+$1 & {\scriptsize[$-$9, $+$11]} & $+$.0220 & {\scriptsize[$+$.0193, $+$.0245]} &  \\
\quad GRPO 1 & -- & 64 & $-$11 & {\scriptsize[$-$16, $-$7]} & $-$4 & {\scriptsize[$-$12, $+$3]} & $+$.0219 & {\scriptsize[$+$.0192, $+$.0245]} &  \\
\quad GRPO 2 & -- & 64 & $-$7 & {\scriptsize[$-$12, $-$3]} & $+$2 & {\scriptsize[$-$6, $+$11]} & $+$.0219 & {\scriptsize[$+$.0194, $+$.0244]} &  \\
\bottomrule
\end{tabular}
\caption{\textbf{All 29 strata of the constrained grid.} Each row is one training history evaluated on one task and critic and, for stacking, one decoy color; the two stacking colors of a training share its history. Entries are critic-minus-uniform changes after optimization with paired 95\% intervals: wrong-object failures and successes as counts out of $n$ evaluation executions, and the original critic score in its native units (Robometer success probability; LIV cosine). A check marks the 11 strata whose wrong-object interval lies above zero.}
\label{tab:prevalence_strata}
\end{table}

\subsection{Retrospective association with initial-policy misranking}
\label{app:retrospective}

For each of the 29 strata, two AUROCs are computed from initial-policy
trajectories alone, in the notation of Equation~\ref{eq:auc_mixture}: the
aggregate success--failure AUROC $A_{S,F}$ and the semantic
success--wrong-object AUROC $A_{S,W}$. Each is compared with the later
critic-minus-uniform change in wrong-object failure, $\Delta W$. The
Spearman correlation of $A_{S,W}$ with $\Delta W$ is $-.82\,[-.91,-.68]$,
against $-.34\,[-.63,+.05]$ for $A_{S,F}$; their difference is
$-.48\,[-.80,-.20]$. Intervals resample the 20 training histories.
Computing each AUROC and $\Delta W$ on disjoint halves of the initial-policy
panel gives $-.74$ for $A_{S,W}$ and $-.31$ for $A_{S,F}$. The association
is carried by differences between task--critic families and between decoy
colors: it is absent within both drawer families, and four near-color LIV
strata have $A_{S,W}<.5$ without amplification. It is therefore a
retrospective description of these settings, not a validated prospective
predictor. Metric definitions, the disjoint-halves procedure and the
complete correlations are in Appendix~\ref{sec:static_vulnerability}
and Table~\ref{tab:static_vulnerability}.

\clearpage
\section{Robustness of the Full-denoiser Results}
\label{app:full_denoiser_robustness}

\subsection{Native diffusion sampler}
\label{app:native_sampler}

Fine-tuning and the primary evaluation decode each action chunk with the
five-step stochastic DDIM chain of Equation~\ref{eq:denoising_policy}. The
policy checkpoint also defines a native sampler: its 100-step DDPM scheduler
with the clipped clean-sample estimate, with a standard-deviation floor of .05
applied at the final step only. We re-evaluate the shared warm-start
initialization and the five iteration-40 critic policies with the native sampler
on the same 512 seeds, OSMesa renderer and scorer. Episodes are decoded in
lockstep batches; each episode draws its initial latent and every step's noise
from its own generator seeded by the evaluation seed, in the order of the
sequential native evaluator, which the batched implementation reproduces
exactly on a decoded chain and outcome for outcome on 16 reference seeds.
The initialization itself performs differently under the two samplers, so each
sampler's policies are compared with that sampler's evaluation of the
initialization (Table~\ref{tab:native_critic}).

\begin{table}[htbp]
\centering\small
\setlength{\tabcolsep}{4pt}
\begin{tabular}{@{}lr@{\,}lr@{\,}lr@{\,}l@{}}
\toprule
Training & \multicolumn{2}{c}{$\Delta$ Success (pp)} & \multicolumn{2}{c}{$\Delta$ Wrong-object (pp)} & \multicolumn{2}{c}{$\Delta$ Robometer} \\
\midrule
Critic 1 & +8.0 & {\scriptsize[3.7, 12.3]} & +6.6 & {\scriptsize[2.5, 10.7]} & +.092 & {\scriptsize[.060, .124]} \\
Critic 2 & +13.1 & {\scriptsize[8.4, 18.0]} & +4.9 & {\scriptsize[0.8, 9.0]} & +.097 & {\scriptsize[.062, .132]} \\
Critic 3 & +10.4 & {\scriptsize[5.9, 14.8]} & +7.2 & {\scriptsize[3.1, 11.3]} & +.075 & {\scriptsize[.042, .108]} \\
Critic 4 & +13.3 & {\scriptsize[8.8, 17.8]} & +1.2 & {\scriptsize[$-$2.7, 5.1]} & +.085 & {\scriptsize[.052, .118]} \\
Critic 5 & +9.6 & {\scriptsize[5.5, 13.7]} & +5.1 & {\scriptsize[1.2, 9.0]} & +.069 & {\scriptsize[.038, .101]} \\
\midrule
Critic, pooled & +10.9 & {\scriptsize[6.6, 15.2]} & +5.0 & {\scriptsize[1.0, 8.9]} & +.084 & {\scriptsize[.054, .114]} \\
\bottomrule
\end{tabular}
\caption{\textbf{Critic-only policies under the policy's native sampler.} The shared initialization and the five iteration-40 critic policies of Table~\ref{tab:whole_network_all}, decoded with the checkpoint's 100-step DDPM sampler (standard-deviation floor .05 at the final step) instead of the five-step fine-tuning sampler, on the same 512 seeds, renderer and scorer. The initialization succeeds in 31.4\% and fails on the wrong object in 19.3\% of these executions.}
\label{tab:native_critic}
\end{table}

\subsection{Distance from the initial policy}
\label{app:kl_behavior}

The full-denoiser learner adapts its actor learning rate to a fixed KL target
per update, so all objectives move at a similar rate. We measure how far each
final policy moved: each decision's five-step chain is a product of Gaussian
transitions with the same standard deviation under both policies, so its KL
divergence is
$\sum_k \lVert\mu_\phi(x_k)-\mu_0(x_k)\rVert^2/(2\sigma^2)$ with $x_k$ drawn
from the evaluated policy. The evaluation episodes store $x_k$,
$\mu_\phi(x_k)$ and the observation condition of every decision, so only the
initial network's means are recomputed; on the initial policy's own episodes
the recomputed and stored means agree to $3\times10^{-6}$. Product and
verifier-only policies end slightly further from the initial policy than the
critic policies, and the supervised Robometer and completion policies end at
nearly the same distance (Table~\ref{tab:kl_behavior}), so the outcome
differences are not explained by how far each policy moved.

\begin{table}[htbp]
\centering\small
\setlength{\tabcolsep}{4pt}
\begin{tabular}{@{}lrrrr@{}}
\toprule
Family and checkpoint & Trainings & KL per decision & $\Delta$ Success (pp) & $\Delta$ Wrong-object (pp) \\
\midrule
Warm start, critic, iteration 10 & 5 & 0.46 [0.42, 0.50] & +1.1 & +0.7 \\
Warm start, critic, iteration 20 & 5 & 1.01 [0.98, 1.05] & +3.1 & +2.0 \\
Warm start, critic, iteration 30 & 5 & 1.62 [1.57, 1.67] & +6.2 & +2.7 \\
Warm start, critic, iteration 40 & 5 & 2.15 [2.03, 2.25] & +8.4 & +3.8 \\
\midrule
Warm start, product, iteration 10 & 5 & 0.55 [0.51, 0.60] & +3.2 & $-$0.4 \\
Warm start, product, iteration 20 & 5 & 1.20 [1.17, 1.23] & +8.6 & $-$0.7 \\
Warm start, product, iteration 30 & 5 & 1.80 [1.74, 1.92] & +12.6 & $-$1.5 \\
Warm start, product, iteration 40 & 5 & 2.38 [2.25, 2.48] & +15.5 & $-$0.6 \\
Warm start, verifier only, iteration 40 & 5 & 2.45 [2.22, 2.54] & +15.9 & $-$1.1 \\
\midrule
Supervised, Robometer, iteration 40 & 8 & 2.46 [2.23, 2.78] & +10.0 & +10.2 \\
Supervised, completion, iteration 40 & 8 & 2.33 [2.11, 2.70] & +14.5 & +1.9 \\
\bottomrule
\end{tabular}
\caption{\textbf{Distance from the initial policy and behavior.} The KL divergence of each policy's five-step denoising chain from the initial policy's, per decision, estimated on that policy's own 512 evaluation episodes (mean over trainings, range in brackets), beside the changes in success and wrong-object failure on the same seeds.}
\label{tab:kl_behavior}
\end{table}

\subsection{Distractor-opening threshold}
\label{app:threshold_sweep}

A wrong-object failure requires the requested task to fail and the distractor
to open by more than 20\,mm. Table~\ref{tab:threshold_sweep} repeats the
primary comparisons for thresholds from 5 to 50\,mm.

\begin{table}[htbp]
\centering\small
\setlength{\tabcolsep}{3pt}
\begin{tabular}{@{}lr@{\,}lr@{\,}lr@{\,}lr@{\,}l@{}}
\toprule
Threshold & \multicolumn{2}{c}{Warm-start critic} & \multicolumn{2}{c}{Critic $-$ product} & \multicolumn{2}{c}{Supervised Robometer} & \multicolumn{2}{c}{Robometer $-$ completion} \\
\midrule
5\,mm & +3.9 & {\scriptsize[0.7, 7.0]} & +5.2 & {\scriptsize[2.6, 8.0]} & +12.1 & {\scriptsize[8.4, 15.9]} & +10.9 & {\scriptsize[7.0, 14.9]} \\
10\,mm & +3.7 & {\scriptsize[0.6, 6.8]} & +5.0 & {\scriptsize[2.4, 7.7]} & +10.9 & {\scriptsize[7.1, 14.7]} & +9.6 & {\scriptsize[5.9, 13.6]} \\
15\,mm & +4.1 & {\scriptsize[1.0, 7.2]} & +4.8 & {\scriptsize[2.2, 7.6]} & +10.4 & {\scriptsize[6.7, 14.1]} & +9.6 & {\scriptsize[6.0, 13.4]} \\
20\,mm & +3.8 & {\scriptsize[0.8, 6.8]} & +4.4 & {\scriptsize[1.8, 7.1]} & +10.9 & {\scriptsize[7.3, 14.4]} & +9.2 & {\scriptsize[5.5, 13.0]} \\
25\,mm & +3.9 & {\scriptsize[1.2, 6.8]} & +4.2 & {\scriptsize[1.8, 6.8]} & +10.9 & {\scriptsize[7.5, 14.5]} & +9.0 & {\scriptsize[5.5, 12.5]} \\
30\,mm & +4.3 & {\scriptsize[1.5, 7.1]} & +4.5 & {\scriptsize[1.9, 7.1]} & +10.7 & {\scriptsize[7.3, 14.3]} & +8.8 & {\scriptsize[5.3, 12.5]} \\
40\,mm & +3.1 & {\scriptsize[0.4, 5.9]} & +3.0 & {\scriptsize[0.5, 5.5]} & +10.3 & {\scriptsize[6.7, 14.0]} & +8.2 & {\scriptsize[4.5, 11.9]} \\
50\,mm & +2.7 & {\scriptsize[$-$0.1, 5.6]} & +3.1 & {\scriptsize[0.6, 5.7]} & +10.7 & {\scriptsize[7.3, 14.3]} & +7.9 & {\scriptsize[4.4, 11.5]} \\
\bottomrule
\end{tabular}
\caption{\textbf{Wrong-object changes across distractor-opening thresholds.} Each entry is the pooled change (pp) in failures that leave the target closed while opening the distractor by more than the threshold, with crossed bootstrap intervals. The paper uses 20\,mm.}
\label{tab:threshold_sweep}
\end{table}

\clearpage
\section{Tilt Model Details}
\label{app:tilt}

\paragraph{Full denoiser.}
For each study and optimized reward we reweight the initial policy's 512
evaluation episodes by $q_i\propto\exp(\lambda r_i)$ and choose $\lambda$ by
bisection so that $\sum_i q_i r_i$ exceeds the initial mean by the reward gain
the family reached at iteration 40 (Robometer for critic runs, completion for
completion runs, the product and the verifier probability for Product and
verifier-only runs). Predicted changes are the reweighted minus initial outcome
rates. Intervals resample the 512 initial episodes and re-solve $\lambda$ in
each of 2,000 draws (Table~\ref{tab:tilt_full}). Reweighting can only move mass
among behaviors the initial policy already produced, which is why it is a
first-order description of full-denoiser fine-tuning.

\begin{table}[htbp]
\centering\small
\setlength{\tabcolsep}{3pt}
\begin{tabular}{@{}lrr@{\,}lrr@{\,}lr@{}}
\toprule
Family & Reward gain & \multicolumn{2}{c}{Predicted $\Delta$ S} & Measured & \multicolumn{2}{c}{Predicted $\Delta$ W} & Measured \\
\midrule
Supervised, Robometer & +.124 & +9.7 & {\scriptsize[0.1, 0.1]} & +10.0 & +7.2 & {\scriptsize[0.1, 0.1]} & +10.2 \\
Supervised, completion & +.163 & +14.3 & {\scriptsize[0.1, 0.2]} & +14.5 & $-$2.5 & {\scriptsize[$-$0.0, $-$0.0]} & +1.9 \\
Warm start, critic & +.061 & +4.1 & {\scriptsize[0.0, 0.0]} & +8.4 & +3.9 & {\scriptsize[0.0, 0.0]} & +3.8 \\
Warm start, product & +.203 & +27.7 & {\scriptsize[0.3, 0.3]} & +15.5 & $-$5.0 & {\scriptsize[$-$0.1, $-$0.0]} & $-$0.6 \\
Warm start, verifier only & +.339 & +33.3 & {\scriptsize[0.3, 0.4]} & +15.9 & $-$5.8 & {\scriptsize[$-$0.1, $-$0.0]} & $-$1.1 \\
\bottomrule
\end{tabular}
\caption{\textbf{Tilt predictions for the full-denoiser studies.} The initial policy's 512 evaluation episodes are reweighted by $\exp(\lambda r)$ with $\lambda$ chosen so that the reweighted mean of the optimized reward equals the gain each family reached at iteration 40. Intervals resample the initial episodes. Changes are in percentage points.}
\label{tab:tilt_full}
\end{table}

\paragraph{Constrained selectors.}
The drawer selector is unconditional and starts every episode from the same
scene, so its KL-regularized optimum from the uniform selector is
$q_k\propto\exp(\lambda\bar r_k)$ over the 24 sequences. For each stratum we
estimate $\bar r_k$, the success rate and the wrong-object rate of every
sequence from all executions of that stratum's run outside the critic arm:
the uniform draws of the first round, the oracle, Product and verifier-only
arms in training, and their fixed evaluation; every stratum covers all 24
sequences. The critic arm's measured reward gain on the fixed evaluation sets
$\lambda$. The stacking selector conditions on the initial image, so pooling
its sequences over contexts makes the same computation an approximation; in
three distinct-color Robometer stacking strata the conditional selector's
reward gain exceeds every sequence's mean reward and no unconditional tilt
reaches it, leaving 26 strata. The first-order predictor is the covariance over
sequences between an outcome rate and the standardized $\bar r_k$, which needs
no tilt strength. The raw evaluation records reproduce the frozen stratum
changes exactly.

\section{Robometer's Published Online-RL Recipe}
\label{app:robometer_recipe}

Robometer's automatic online RL pays $-1$ per step plus the model's reward
prediction in $[0,1]$, pays $0$ instead of $-1$ on the step where the success
probability exceeds $.6$, and terminates there \citep{liang2026robometer}. The
drawer selector makes one choice per episode, so the recipe enters it only
through the episode return. Every executed drawer episode of the five Robometer
selector trainings stores Robometer's success probability and progress
prediction at eight frames (control steps 0, 43, \ldots, 299), from one forward
pass over the clip; we hold progress at the last frame's value between frames
and evaluate physical outcomes on the full episode. Table~\ref{tab:robometer_recipe}
reports how often each outcome triggers termination, how each return orders the
outcomes, and the tilted selector at the KL from uniform that the trained
terminal-reward selectors reached ($2.07$ nats on average). Thresholds of $.5$
and $.7$ give the same ordering (AUROC $.152$ and $.120$ for success against
wrong-object failures).

\begin{table}[htbp]
\centering\small
\setlength{\tabcolsep}{4pt}
\begin{tabular}{@{}lrrr@{}}
\toprule
& Success & Wrong-object & Other failure \\
\midrule
Executions & 2,009 & 1,130 & 1,325 \\
Detected as success (probability $>$ .6) & 73.4\% & 89.4\% & 1.6\% \\
Mean detection step (of 300) & 295 & 236 & 297 \\
\midrule
& \multicolumn{3}{c}{AUROC: S vs.\ W / other / all failures} \\
Terminal success probability & \multicolumn{3}{c}{0.195 / 0.951 / 0.603} \\
Dense progress with success termination & \multicolumn{3}{c}{0.125 / 0.935 / 0.562} \\
\midrule
& \multicolumn{3}{c}{Tilt at the trained KL (2.07 nats): $\Delta$S / $\Delta$W (pp)} \\
Terminal success probability & \multicolumn{3}{c}{-9.1 / +65.5} \\
Dense progress with success termination & \multicolumn{3}{c}{-8.7 / +62.0} \\
\bottomrule
\end{tabular}
\caption{\textbf{Robometer's published online-RL recipe on the drawer selector.} All 4,464 executed drawer episodes of the five Robometer selector trainings, rescored with the recipe: every step pays $-1$ plus the predicted progress, the step whose success probability exceeds .6 pays $0$ and ends the episode. The last block tilts the uniform selector by each reward to the KL that the trained terminal-reward selectors reached.}
\label{tab:robometer_recipe}
\end{table}

\clearpage
\section{Evaluator Robustness and Statistical Sensitivity}
\label{supp:execution}

These analyses check the statistical design of the primary endpoint and the
fine-tuning sampler's calibration.

\subsection{Bootstrap design for the primary endpoint}
\label{supp:resampling}

The main text uses crossed training-by-evaluation-seed intervals and draws
trainings independently across objectives for product--critic contrasts.
The following alternatives retain every training and the same 512 seeds.

\begin{center}
\small
\begin{tabular}{lrrr}
\toprule
Pooled effect & Success (pp) & Wrong-object (pp) & $\Delta r_b$ \\
\midrule
Critic, nested & $+8.4\,[+6.3,+10.5]$ & $+3.8\,[+2.0,+5.6]$ & $+.061\,[+.047,+.075]$ \\
Critic, crossed & $+8.4\,[+4.9,+11.9]$ & $+3.8\,[+0.8,+6.8]$ & $+.061\,[+.036,+.086]$ \\
Product, nested & $+15.5\,[+13.6,+17.3]$ & $-0.6\,[-2.3,+1.0]$ & $+.096\,[+.081,+.111]$ \\
Product, crossed & $+15.5\,[+11.8,+19.1]$ & $-0.6\,[-3.4,+2.1]$ & $+.096\,[+.070,+.122]$ \\
P $-$ C, nested & $+7.1\,[+5.3,+8.9]$ & $-4.4\,[-6.3,-2.5]$ & $+.034\,[+.017,+.053]$ \\
P $-$ C, crossed, index-paired & $+7.1\,[+4.6,+9.7]$ & $-4.4\,[-7.0,-1.8]$ & $+.034\,[+.011,+.058]$ \\
P $-$ C, crossed, independent & $+7.1\,[+4.4,+9.8]$ & $-4.4\,[-7.1,-1.8]$ & $+.034\,[+.013,+.056]$ \\
\bottomrule
\end{tabular}
\end{center}

Nested resampling draws evaluation seeds independently within each sampled
training and understates their shared variation. Index-paired crossed
resampling treats training indices as matched; they do not identify matched
training randomness. These two rows are sensitivity analyses; the main
contrast uses independent training resampling. Pooled completion rises by
$.087\,[.057,.118]$ under critic-only training and
$.140\,[.106,.175]$ under the product.

\subsection{Sampler calibration and worker-reset diagnostic}
\label{app:weight_calibration}

Before any training we evaluated the unmodified network under the short
stochastic sampler on 64 paired seeds per cell.

\begin{center}\small
\begin{tabular}{lrrrr}
\toprule
Sampler & Success/64 & Mean completion & Mean final opening & Distractor gain $>5$\,mm \\
\midrule
$K{=}5$, $\sigma{=}.05$ & 12 & $.244$ & 20.1\,mm & 21 \\
$K{=}10$, $\sigma{=}.05$ & 7 & $.255$ & 19.7\,mm & 23 \\
$K{=}5$, $\sigma{=}.10$ & 1 & $.039$ & 3.1\,mm & 14 \\
$K{=}10$, $\sigma{=}.10$ & 1 & $.067$ & 4.1\,mm & 14 \\
\bottomrule
\end{tabular}
\end{center}

The two $\sigma{=}.10$ cells are not interpretable as written. The batched
worker created its environments and ran its first batch before the simulator
performed a hard reset, so those episodes started from the construction
state rather than the initial state. Scheduling placed 64/64 of the
$K{=}5,\sigma{=}.10$ episodes and 48/64 of the $K{=}10,\sigma{=}.10$
episodes in such first batches, while both $\sigma{=}.05$ cells contain none.
The 16 unaffected $\sigma{=}.10$ episodes give 0 successes and completion
$.110$, weaker than $\sigma{=}.05$ but too few to establish the point. We
therefore report $\sigma{=}.10$ as untested rather than as a negative
result. The fix is a discarded warmup batch per worker. Oracle history A had
already used first-batch data in one actor update and was restarted from
iteration~0. In histories B and C, first-batch data entered only the first
iteration, which trains the value network alone; their actors never used such
data and they were not restarted, but that first value-network update did see
it. Every endpoint evaluation and every learned-reward training ran with the
corrected workers. We did not
sweep $\sigma<.05$; that is an unswept hyperparameter and we record it as
such.

\paragraph{Actor learning-rate calibration.}
An initial actor learning-rate test at $5\times10^{-5}$ diverged, with
mean sampled KL 23 per denoising step after three optimizer steps.
The implemented protocol sums coordinate log-densities and uses a
KL-adaptive actor rate bounded by $[10^{-8},10^{-4}]$.

\section{Complete Constrained-policy Grid and Trajectory Diagnostics}
\label{supp:grid}
\label{app:common_protocol_matrix}

This section retains all 20 main training histories, 29 task/color strata,
43 supervision-budget rows and five matched-only controls. It includes
settings with and without wrong-object amplification. The repeated-error
subset in the main text is not an estimate of prevalence across this grid
or across robot tasks generally.

\subsection{Optimization and Independent Evaluation}
\label{sec:categorical_ppo}

The common policy family and PPO/GRPO updates are specified in
the constrained-policy protocol of Section~\ref{sec:online_noise_policy}. The frozen
decoder uses 100-step denoising diffusion probabilistic model (DDPM)
sampling~\citep{ho2020ddpm}. All 24 complete stochastic paths are retained,
including transition noise, and the decoder conditions
on new observations throughout execution. Fresh Gaussian jitter of
standard deviation $.05$ is applied to each noise draw. The trainable
policy chooses the noise component: 24 unconditional drawer logits, or a
$24\times129$ image-conditioned cube matrix. The diffusion decoder is
frozen. Each rollout generates actions from fresh visual observations;
scene IDs and object-state labels are excluded from the selector input.

\paragraph{Noise-bank construction and reuse.}
The drawer support contains all 24 seeds $52103000$--$52103023$, the first
of three development banks with consecutive 24-seed blocks starting at
$52103000$, $53103000$ and $54103000$. It is reused in all eleven drawer
panels. The nine stacking panels share the 24 seeds
$120100000$--$120100023$. No member is removed within either selected bank.
These are fixed development supports: the drawer banks had been evaluated
before the first recorded downstream configuration, and no outcome-blind
bank-selection rule was recorded. Repetitions vary training and evaluation
seeds on these supports; they do not sample new supports.

Each objective receives eight rounds of 24 PPO or 32 GRPO feedback
episodes, with eight Adam epochs per round and persistent optimizer
state. Initial uniform feedback is shared across identical policies;
later feedback is collected separately. Thus, unique trajectory counts
depend on the number of objectives rather than just the per-objective
budget. GRPO groups contain four candidates with identical initial
physical state and images but distinct rollout noise. Drawer R114's
eight groups are draws at the same physical state, not eight different
task contexts. Cube groups balance the trained color conditions.

Both optimizers use clipping $.2$ and gradient-norm limit 1. The learning
rate is $.2$ for drawer logits and $.02$ for conditional cube policies, and
the PPO entropy coefficient is $.02$; grouped GRPO replaces that term with
exact categorical KL to the fixed initial uniform policy at coefficient
$.04$. Cube feedback is balanced across the two trained colors at the same
total budget, and the shared initial round avoids duplicate executions of
identical initial policies.

Every main endpoint uses the frozen eighth checkpoint, without selecting
checkpoints by final success. Drawer Robometer PPO uses 32 new paired
executions; other common panels use 64 per condition. Original drawer
Robometer analyses use 10,000 paired execution resamples, cube uses
20,000 paired resamples within each of four cells, and drawer LIV uses
20,000 paired bank-cluster resamples preserving four members of each of
16 banks. All intervals are percentile 95\% intervals conditional on one
training and the fixed support. Neither multiple colors, label budgets,
nor matched-only extensions increase the number of independent trainings.
PPO and GRPO have different feedback budgets and are not ranked for
sample efficiency.

Recorded sampling probabilities, final parameter arrays and continuous
Adam state are verified against the saved updates. Execution manifests
retain seeds, model identities, renderer assignments and original scores;
physical outcomes are independently reconstructed from state/contact
traces. Paired references share a renderer within each panel. Changes
between EGL and OSMesa across replications are recorded, not assumed
pixel-identical or isolated from training-seed changes. The summaries
below retain complete endpoints rather than partial-progress snapshots.

\subsection{Selector Representation and Exact Update Objective}
\label{app:selector_details}

For the image-conditioned stacking selector, a fixed covariance transform of
128 visual coordinates followed by a bias gives $g(o_0)\in\mathbb R^{129}$,
with $p_\phi(I\mid o_0)=\operatorname{softmax}(Wg(o_0))_I$. The transform is
fitted on 864 unlabeled development inputs, with relative eigenvalue floor
.001 and no RMS-preserving gain. Drawer policies instead use 24
unconditional logits.

The categorical PPO loss is
\begin{equation}
 \mathcal L_{\rm PPO}=-\frac1B\sum_i\min\!\left(
 \rho_i\widehat A_i,\operatorname{clip}(\rho_i,1-\epsilon,1+\epsilon)
 \widehat A_i\right)-\beta\E[\mathcal H(p_\phi)].
\end{equation}
Advantages center terminal reward within a batch. Stacking additionally
standardizes advantages; drawer PPO retains original-scale and explicitly
normalized variants separately. Grouped GRPO standardizes within groups of
four and replaces entropy with an exact categorical KL to the fixed initial
uniform policy, coefficient .04. Conditional on the chosen component, the
noise jitter, frozen decoder and environment do not depend on trainable
selector parameters, so $\rho_i=p_\phi(I_i\mid o_{0,i})/
p_{\rm old}(I_i\mid o_{0,i})$ is the exact trajectory likelihood ratio for
this policy family.

\subsection{Observed Dynamics and Residual Errors}
\label{app:ppo_endpoints}

The first sorted R82 evaluation seed supplies the physical trajectory
example; it was selected independently of outcome. Images in the main
text are the original critic frames. High-score wrong-object failures
increase from 7 to 24 of 32 executions. All 24 rewarded wrong-object
failures in this panel keep the target below 6.141\,mm throughout the
episode while additionally opening the distractor by 50.121--63.124\,mm.
After mitigation, four failures still receive high original reward and
five satisfy the broader wrong-object definition. The four rewarded
residuals keep the target below 29.722\,mm; three sample component 14
and one component 6. Their verifier probabilities are .0671--.0932 and
product rewards .05685--.08041. Detection does not eliminate every
stochastic failure, and this panel does not isolate exploration from
execution variability.

\begin{figure}[htbp]
\centering
\includegraphics[width=.94\linewidth]{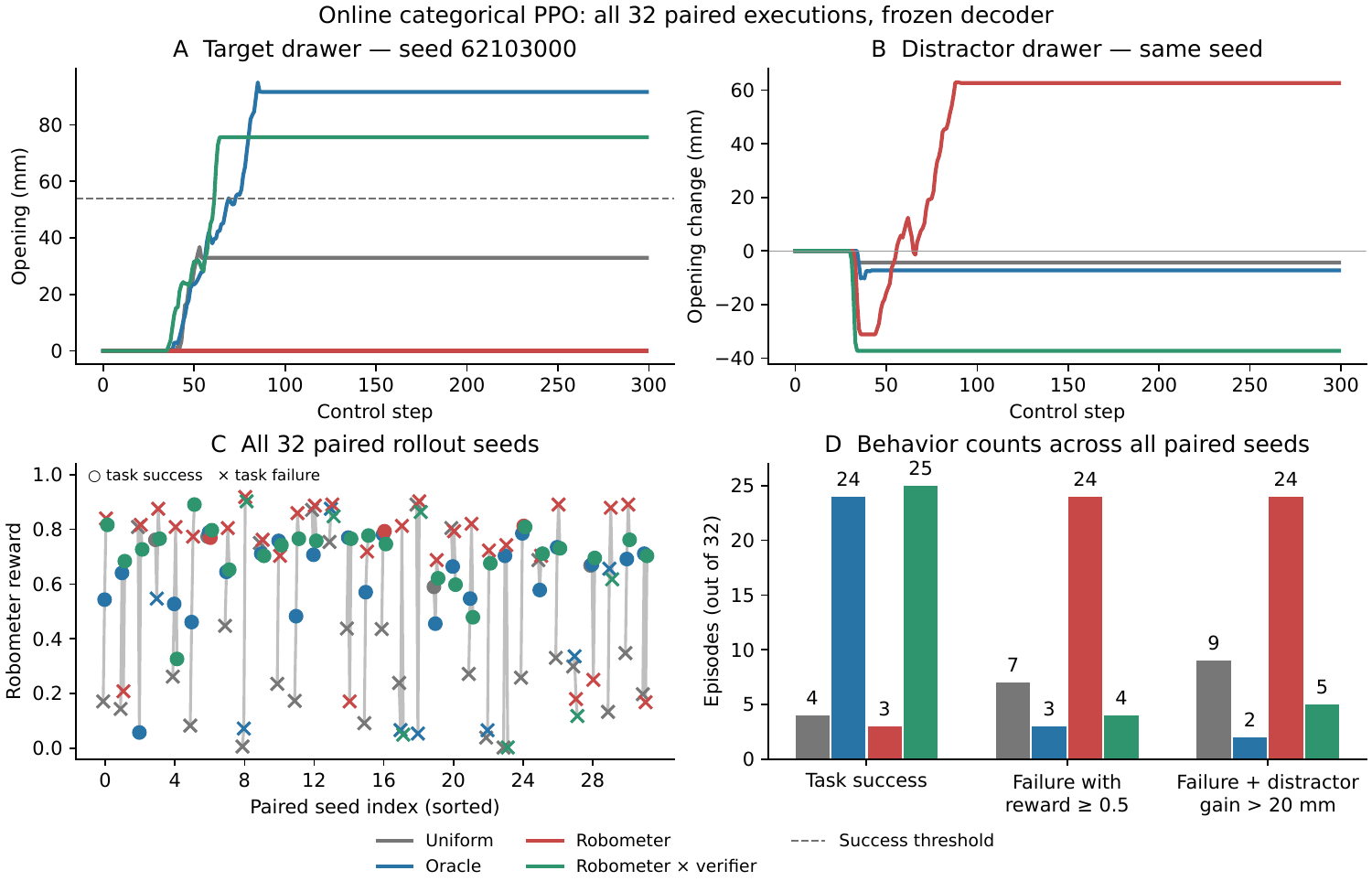}
\caption{R82 fixed final PPO evaluation: the prespecified first paired seed,
all 32 paired scores and complete physical behavior counts. Wrong-object
counts include failures regardless of original reward.}
\label{fig:ppo_paired}
\end{figure}

\begin{figure}[htbp]
\centering
\includegraphics[width=.94\linewidth]{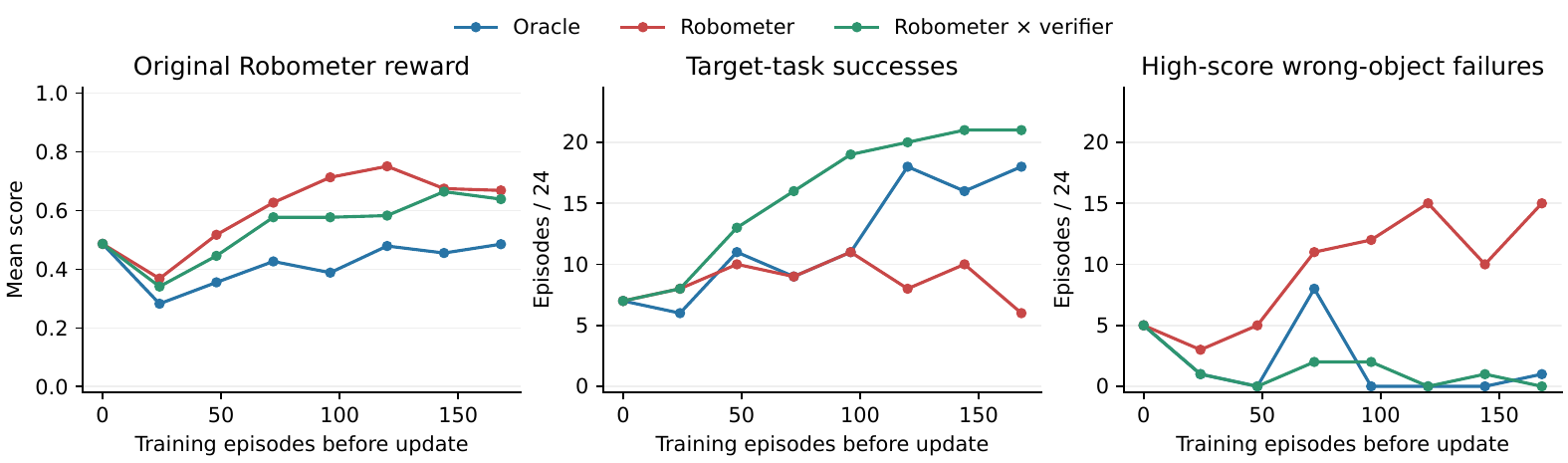}
\caption{All eight R82 PPO feedback batches, 24 executions per objective
before each update, sharing the first uniform batch. Reward always denotes
original Robometer, including on the product-trained trajectory. Training
curves are not substituted for the new-seed final evaluation.}
\label{fig:ppo_training}
\end{figure}

\subsection{All Budgets, Conditions and Matched-only Controls}
\label{sec:balanced_rm_grpo_complete}
\label{sec:rm_grpo_independent_near}
\label{sec:rm_grpo_matched_repeat}

Table~\ref{tab:all_final_budgets} contains every one of the 43 budget rows
from 20 complete main trainings and 29 task/color strata. The missing
drawer PPO only controls and missing 456-only controls in the repeated
LIV cube trainings remain missing. For R164/R166/R167/R165/R184,
supplemental R168/R170/R171/R172/R189 supply identical-456-head only
policies with matched feedback budgets. They reuse the reference seeds
and repeat uniform executions; they are not independent confirmations.
R189 was selected after observing the complete R184 result. Its 128
uniform reexecutions match that reference, but its distinct-color product
advantage does not retrospectively make the first Robometer GRPO run successful.

\begin{table}[t]
\centering\small
\begin{tabular}{lrlrrrr}
\toprule
Run / condition & $n$ & $\Delta r_{C-U}$ [95\% CI] & $U$ & $O$ & $C$ & $P$ \\
\midrule
\multicolumn{7}{l}{\textbf{Drawer / Robometer / PPO}}\\
R82 & 32 & $+.309\,[+.192,+.425]$ & 4/9 & 24/2 & 3/24 & 25/5 \\
R87 & 32 & $+.420\,[+.314,+.522]$ & 11/4 & 19/3 & 2/28 & 25/3 \\
R88 & 32 & $+.432\,[+.318,+.541]$ & 6/6 & 18/4 & 2/28 & 25/5 \\
\midrule
\multicolumn{7}{l}{\textbf{Drawer / Robometer / GRPO}}\\
R114 & 64 & $+.493\,[+.425,+.560]$ & 12/8 & 54/1 & 4/54 & 56/0 \\
R151 & 64 & $+.419\,[+.340,+.497]$ & 10/14 & 58/0 & 4/53 & 52/0 \\
\midrule
\multicolumn{7}{l}{\textbf{Stacking / LIV / PPO}}\\
R167 / near & 64 & $+.0007\,[-.0006,+.0021]$ & 20/21 & 38/17 & 25/19 & 38/16 \\
R167 / distinct & 64 & $+.0070\,[+.0047,+.0093]$ & 20/21 & 63/0 & 6/46 & 63/0 \\
R174 / near & 64 & $+.0047\,[+.0031,+.0064]$ & 26/19 & 47/13 & 21/17 & 41/9 \\
R174 / distinct & 64 & $+.0085\,[+.0065,+.0105]$ & 24/18 & 57/2 & 3/42 & 57/2 \\
\midrule
\multicolumn{7}{l}{\textbf{Stacking / LIV / GRPO}}\\
R165 / near & 64 & $+.0012\,[-.0002,+.0026]$ & 16/20 & 41/15 & 11/21 & 32/24 \\
R165 / distinct & 64 & $+.0067\,[+.0047,+.0086]$ & 15/18 & 55/6 & 2/53 & 52/7 \\
R173 / near & 64 & $+.0025\,[+.0009,+.0042]$ & 23/16 & 38/16 & 21/11 & 40/13 \\
R173 / distinct & 64 & $+.0070\,[+.0051,+.0091]$ & 22/15 & 51/9 & 5/36 & 52/6 \\
\bottomrule
\end{tabular}
\caption{Physical repair in drawer Robometer and stacking LIV. Each policy entry is success/wrong-object count out of $n$; $\Delta r$ is original native reward change. $U/O/C/P$ denote uniform/oracle/critic/product. Drawer uses 384 labels; stacking uses 2,536 (2,080 beyond the smaller head). Both trained colors and all nine main trainings are retained. The verifier-only ablation appears in Section~\ref{app:verifier_only_ablation}; all budgets and frozen physical intervals appear in Section~\ref{app:common_protocol_matrix}.}
\label{tab:repair_endpoints}
\end{table}

\begin{figure}[htbp]
\centering
\includegraphics[width=\linewidth]{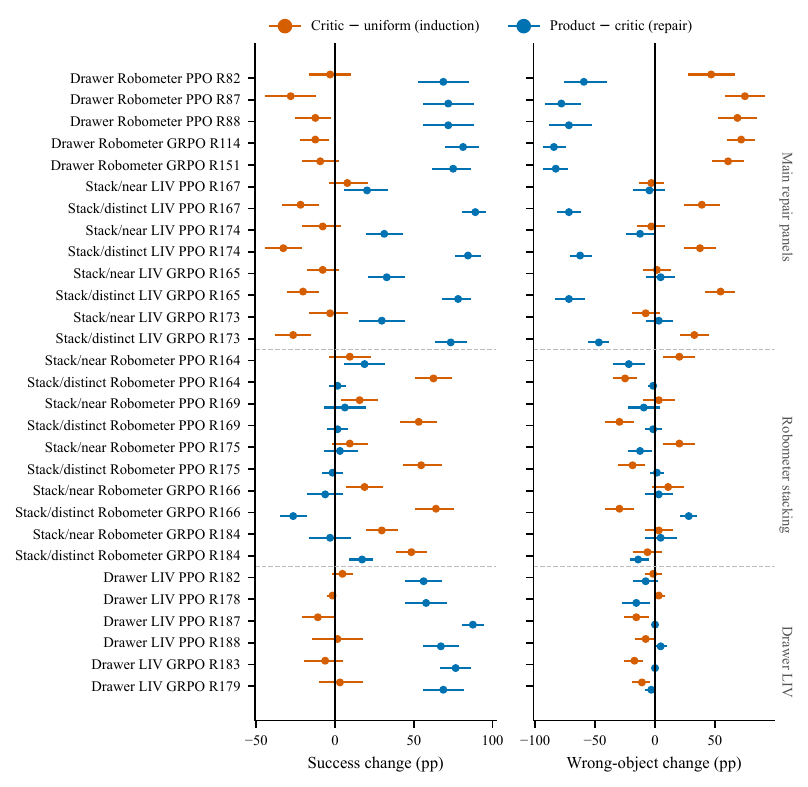}
\caption{All 29 task/color strata of the constrained-policy grid, grouped as
main repair panels, Robometer stacking and drawer LIV. Orange: critic minus
uniform; blue: product minus critic; frozen 95\% paired intervals. Robometer
stacking and drawer LIV largely lack induction, and product does not uniformly
help there (for example the distinct-color R166 row).}
\label{fig:selector_forest_all}
\end{figure}
\begin{figure}[htbp]
\centering
\includegraphics[width=\linewidth]{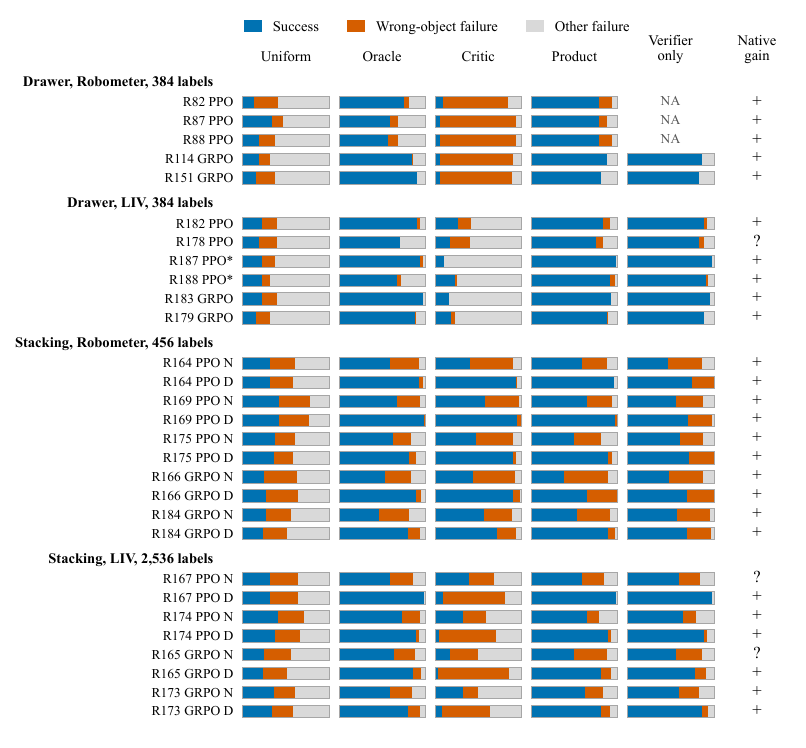}
\caption{Absolute outcomes of every arm in all 29 constrained-policy strata. Each
bar partitions all fresh executions (32 for R82/87/88, 64 otherwise) into
success, wrong-object failure and other failure; widths are proportions, not
uncertainty. N/D: near/distinct trained color; *: normalized PPO; NA: no
matched verifier-only control. The last column marks whether the
critic-minus-uniform native-reward interval is strictly positive (+) or
unresolved (?); it is not a hacking verdict.}
\label{fig:selector_outcomes}
\end{figure}
\begin{figure}[htbp]
\centering
\includegraphics[width=\linewidth]{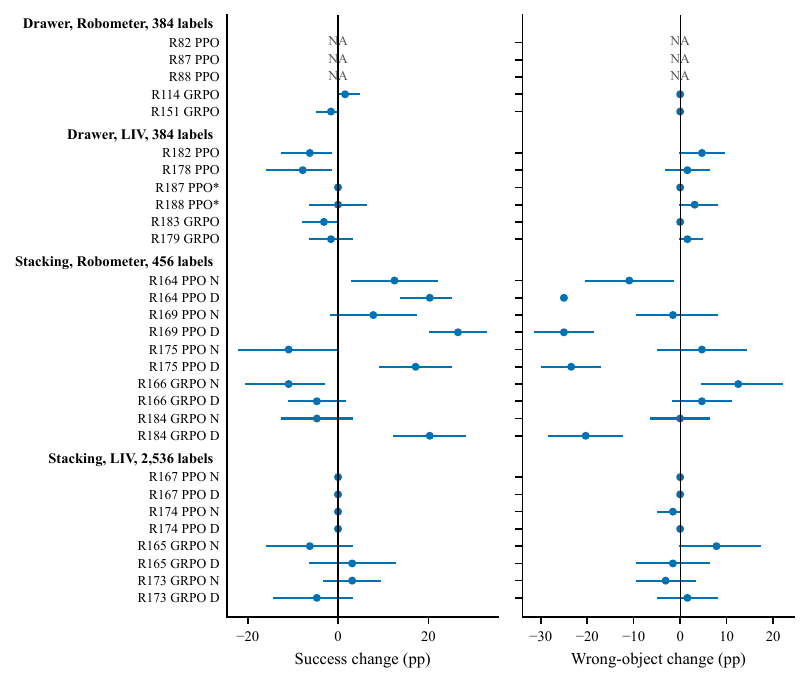}
\caption{Product minus verifier-only in the 26 strata with a matched
verifier-only control, with frozen 95\% paired intervals. Retaining the original
reward raises success in several distinct-color Robometer-stacking strata,
lowers it in R166 near and in drawer LIV R182 and R178, and is indistinguishable
elsewhere; NA marks the three drawer PPO trainings without that control.}
\label{fig:selector_product_verifier}
\end{figure}
\begingroup\small
\setlength{\tabcolsep}{5pt}
\begin{longtable}{lrrrrrrr}
\caption{Every final common-protocol endpoint and supervision budget. Cells are success / total wrong-object counts. N/D denote near/distinct cube colors, both trained. $V$ is the identical-head verifier-only control; -- denotes a missing matched control. Expanded cube heads add 2,080 labels.}\label{tab:all_final_budgets}\\
\toprule
Run / condition & $n$ & Labels & $U$ & $O$ & $C$ & $P$ & $V$ \\
\midrule
\endfirsthead
\multicolumn{8}{l}{Table~\ref{tab:all_final_budgets} continued}\\
\toprule
Run / condition & $n$ & Labels & $U$ & $O$ & $C$ & $P$ & $V$ \\
\midrule
\endhead
\bottomrule
\endfoot
\multicolumn{8}{l}{\textbf{Drawer / Robometer / PPO}}\\*
R82 & 32 & 384 & 4/9 & 24/2 & 3/24 & 25/5 & -- \\
R87 & 32 & 384 & 11/4 & 19/3 & 2/28 & 25/3 & -- \\
R88 & 32 & 384 & 6/6 & 18/4 & 2/28 & 25/5 & -- \\
\multicolumn{8}{l}{\textbf{Drawer / Robometer / GRPO}}\\*
R114 & 64 & 384 & 12/8 & 54/1 & 4/54 & 56/0 & 55/0 \\
R151 & 64 & 384 & 10/14 & 58/0 & 4/53 & 52/0 & 53/0 \\
\multicolumn{8}{l}{\textbf{Drawer / LIV / PPO}}\\*
R182 & 64 & 384 & 14/11 & 58/2 & 17/10 & 53/5 & 57/2 \\
R178 & 64 & 384 & 12/13 & 45/0 & 11/15 & 48/5 & 53/4 \\
R187 / norm. & 64 & 384 & 14/10 & 60/2 & 7/0 & 63/0 & 63/0 \\
R188 / norm. & 64 & 384 & 14/6 & 43/3 & 15/1 & 58/4 & 58/2 \\
\multicolumn{8}{l}{\textbf{Drawer / LIV / GRPO}}\\*
R183 & 64 & 384 & 14/11 & 62/0 & 10/0 & 59/0 & 61/0 \\
R179 & 64 & 384 & 10/10 & 56/1 & 12/3 & 56/1 & 57/0 \\
\multicolumn{8}{l}{\textbf{Stacking / Robometer / PPO}}\\*
R164 / N & 64 & 456 & 20/19 & 38/21 & 26/32 & 38/18 & 30/25 \\
R164 / N & 64 & 2536 & 20/19 & 38/21 & 26/32 & 31/22 & 34/24 \\
R164 / D & 64 & 456 & 20/17 & 59/3 & 60/1 & 61/0 & 48/16 \\
R164 / D & 64 & 2536 & 20/17 & 59/3 & 60/1 & 62/0 & 60/2 \\
R169 / N & 64 & 456 & 27/23 & 43/17 & 37/25 & 41/19 & 36/20 \\
R169 / D & 64 & 456 & 27/22 & 63/1 & 61/3 & 62/2 & 45/18 \\
R175 / N & 64 & 456 & 24/15 & 40/13 & 30/28 & 32/20 & 39/17 \\
R175 / D & 64 & 456 & 23/14 & 52/5 & 58/2 & 57/3 & 46/18 \\
\multicolumn{8}{l}{\textbf{Stacking / Robometer / GRPO}}\\*
R166 / N & 64 & 456 & 16/24 & 34/19 & 28/31 & 24/33 & 31/25 \\
R166 / N & 64 & 2536 & 16/24 & 34/19 & 28/31 & 37/23 & 33/25 \\
R166 / D & 64 & 456 & 17/24 & 57/4 & 58/5 & 41/23 & 44/20 \\
R166 / D & 64 & 2536 & 17/24 & 57/4 & 58/5 & 52/12 & 51/13 \\
R184 / N & 64 & 456 & 17/19 & 30/22 & 36/21 & 34/24 & 37/24 \\
R184 / N & 64 & 2536 & 17/19 & 30/22 & 36/21 & 30/22 & 27/25 \\
R184 / D & 64 & 456 & 15/18 & 51/9 & 46/14 & 57/5 & 44/18 \\
R184 / D & 64 & 2536 & 15/18 & 51/9 & 46/14 & 49/10 & 46/13 \\
\multicolumn{8}{l}{\textbf{Stacking / LIV / PPO}}\\*
R167 / N & 64 & 456 & 20/21 & 38/17 & 25/19 & 28/30 & 29/29 \\
R167 / N & 64 & 2536 & 20/21 & 38/17 & 25/19 & 38/16 & 38/16 \\
R167 / D & 64 & 456 & 20/21 & 63/0 & 6/46 & 43/18 & 41/20 \\
R167 / D & 64 & 2536 & 20/21 & 63/0 & 6/46 & 63/0 & 63/0 \\
R174 / N & 64 & 456 & 26/19 & 47/13 & 21/17 & 35/22 & -- \\
R174 / N & 64 & 2536 & 26/19 & 47/13 & 21/17 & 41/9 & 41/10 \\
R174 / D & 64 & 456 & 24/18 & 57/2 & 3/42 & 45/18 & -- \\
R174 / D & 64 & 2536 & 24/18 & 57/2 & 3/42 & 57/2 & 57/2 \\
\multicolumn{8}{l}{\textbf{Stacking / LIV / GRPO}}\\*
R165 / N & 64 & 456 & 16/20 & 41/15 & 11/21 & 31/27 & 28/29 \\
R165 / N & 64 & 2536 & 16/20 & 41/15 & 11/21 & 32/24 & 36/19 \\
R165 / D & 64 & 456 & 15/18 & 55/6 & 2/53 & 43/16 & 49/14 \\
R165 / D & 64 & 2536 & 15/18 & 55/6 & 2/53 & 52/7 & 50/8 \\
R173 / N & 64 & 456 & 23/16 & 38/16 & 21/11 & 25/30 & -- \\
R173 / N & 64 & 2536 & 23/16 & 38/16 & 21/11 & 40/13 & 38/15 \\
R173 / D & 64 & 456 & 22/15 & 51/9 & 5/36 & 30/28 & -- \\
R173 / D & 64 & 2536 & 22/15 & 51/9 & 5/36 & 52/6 & 55/5 \\
\end{longtable}
\endgroup

\begin{figure}[htbp]
\centering
\includegraphics[width=\linewidth]{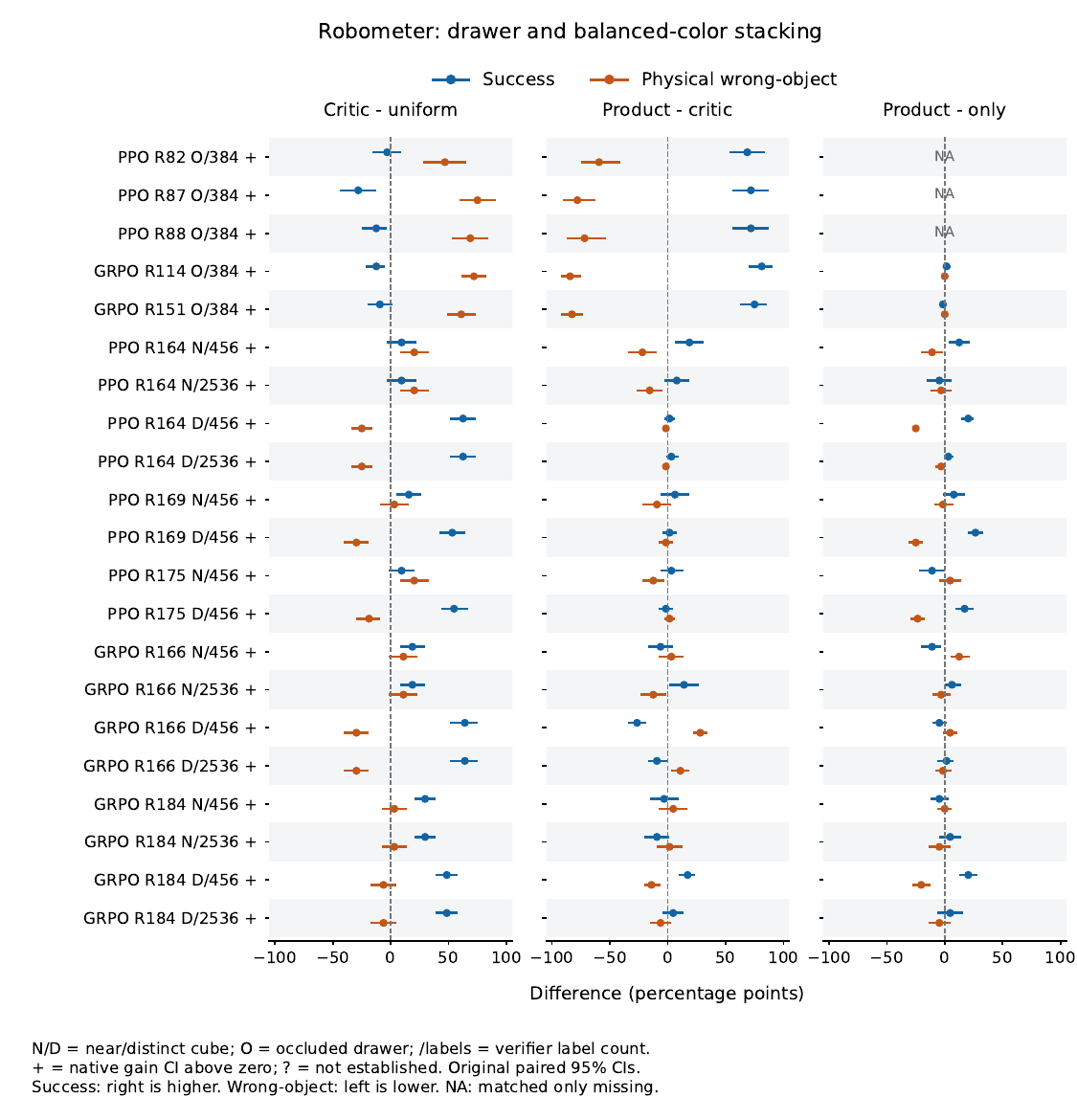}
\caption{All Robometer physical contrasts with their original paired
95\% intervals. Success and total wrong-object changes are separate;
$+$ indicates a native-reward-gain interval above zero. NA denotes a
missing identical-budget only control. Both cube colors are trained.
The rows retain near-color failures and both signs of distinct-color
GRPO effects; row count is not independent-training count.}
\label{fig:common_rm_matrix}
\end{figure}

\begin{figure}[p]
\centering
\includegraphics[width=\linewidth]{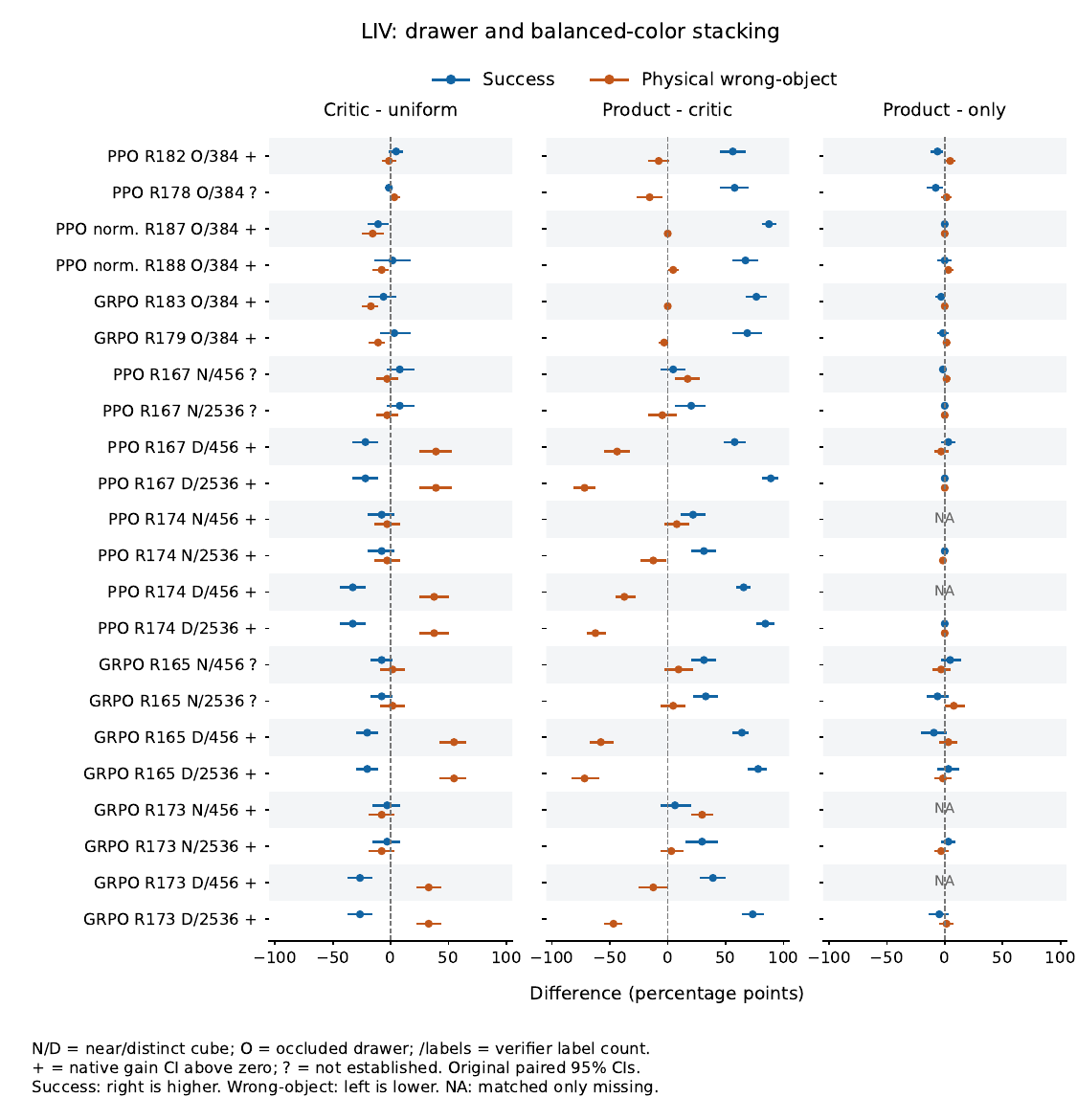}
\caption{All LIV physical contrasts and original paired 95\% intervals,
including both cube label budgets and all six drawer panels. Distinct-color
cube repair repeats, but an added original-LIV contribution over the
identical verifier is not established. Near-color and drawer negative
inductions remain visible. Drawer wrong-opening excludes upper closing;
continuous movement is reported in Section~\ref{app:drawer_liv_scale}.}
\label{fig:common_liv_matrix}
\end{figure}

\subsection{Initial-policy Semantic Misranking}
\label{sec:static_vulnerability}

Using only initial-policy trajectories, we computed the success--all-failure
area under the receiver operating characteristic curve (AUROC; S1),
the success--wrong-object AUROC (S2), a standardized wrong-minus-success
score margin (S3) and top-quartile wrong-object enrichment (S4). We compared
them with the subsequent $\Delta W$ across the 29 distinct induction strata from
20 training histories; every metric is defined in every stratum. The analysis is
retrospective: ``initial-policy'' identifies the policy distribution, not the
collection date. Semantic misranking tracks amplification. The Spearman
correlation of S2 with $\Delta W$ is $-.82\,[-.91,-.68]$ (training-history
bootstrap), against $-.34\,[-.63,+.05]$ for the generic S1; S3 and S4 agree
(Table~\ref{tab:static_vulnerability}). Computing each metric and $\Delta W$ on
disjoint halves of the initial-policy panel gives $-.74\,[-.82,-.61]$ for S2
and $-.31\,[-.58,+.08]$ for S1: the association persists without direct reuse
of baseline observations, although this does not remove family/color
confounding. The association is carried by differences between task/critic
families and between colors: it is absent within both drawer families, and
four near-color LIV strata have S2 below .5 without amplification
(Section~\ref{app:static_vulnerability}). We read it as a retrospective
description of these settings, not a validated predictor.

\paragraph{Metric definitions and analysis.}
\label{app:static_vulnerability}

Each of the 29 strata contributes its initial-policy ($U$) endpoint
executions, scored by its original critic in native units; post-optimization
trajectories never enter a static metric. With $S$, $W$ and $F$ the
successes, wrong-object failures and all failures of a stratum, the metrics
are: (S1) AUROC of $S$ versus $F$; (S2) AUROC of $S$ versus $W$; (S3)
$(\bar r_W-\bar r_S)/s_r$, with $s_r$ the population standard deviation of
all $U$ scores; and (S4) the wrong-object fraction among episodes at or above
the 75th score percentile minus its overall fraction. AUROCs count ties as one
half. An AUROC or margin is unavailable when a required class has fewer than
three episodes, and S3 is unavailable when the scores have no variance. The
response is $\Delta W$, the critic-minus-uniform change in wrong-object rate.
Intervals resample the 20 training histories with all of their strata
(20,000 draws). These definitions and the bootstrap were committed before any
stratum metric was computed. The disjoint-halves analysis and the difference
row of Table~\ref{tab:static_vulnerability} were added afterwards, each
declared before its own computation.

\begin{table}[t]
\centering\small
\setlength{\tabcolsep}{3pt}
\resizebox{\linewidth}{!}{%
\begin{tabular}{lrrrr}
\toprule
Static metric (initial policy) & Full panel $\rho$ & Strata & Disjoint halves $\rho$ (combined) & Strata (A$\to$B / B$\to$A) \\
\midrule
S1 AUROC, success vs all failures & $-.34\,[-.63,+.05]$ & 29 & $-.31\,[-.58,+.08]$ & 28 / 28 \\
S2 AUROC, success vs wrong-object & $-.82\,[-.91,-.68]$ & 29 & $-.74\,[-.82,-.61]$ & 25 / 27 \\
S3 standardized margin & $+.72\,[+.51,+.87]$ & 29 & $+.63\,[+.41,+.76]$ & 29 / 29 \\
S4 top-quartile enrichment & $+.87\,[+.72,+.95]$ & 29 & $+.83\,[+.70,+.90]$ & 29 / 29 \\
\midrule
$\rho$(S2) $-$ $\rho$(S1), common strata & $-.48\,[-.80,-.20]$ & 29 & $-.34\,[-.64,-.14]$ & -- \\
\bottomrule
\end{tabular}}
\caption{Spearman correlation between each initial-policy static metric and the wrong-object change after critic optimization, across induction strata, with 95\% bootstrap intervals over 20 training clusters. The full panel uses the same initial-policy episodes for the metric and the change; the disjoint-halves column computes them on label-blind halves of each panel in both directions and averages the two. The difference row is not pre-declared in the original analysis plan.}
\label{tab:static_vulnerability}
\end{table}

\begin{figure}[htbp]
\centering
\includegraphics[width=\linewidth]{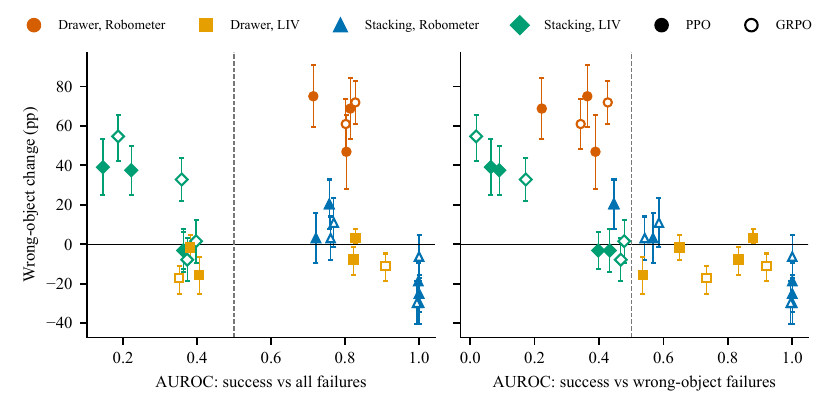}
\caption{Wrong-object change after critic optimization against the
initial-policy generic AUROC (left) and semantic AUROC (right), one point per
stratum, marked by task and critic.}
\label{fig:static_vulnerability}
\end{figure}

\paragraph{Shared baseline.} Both a static metric and $\Delta W$ depend on
the same $U$ episodes. Each panel was therefore split into halves by
identifiers alone: whole evaluation banks alternate for the drawer strata, and
seeds alternate within each design cell for stacking. Metrics come from one
half's $U$ episodes and $\Delta W$ from the other half's paired executions,
in both directions, averaged. Halving leaves S2 unavailable in 4 and 2
strata. The negative association persists when the static metric and the
amplification estimate use disjoint baseline halves. This reduces direct reuse
of baseline observations but does not eliminate family/color confounding,
establish prospective prediction, or quantify all sources of dependence.

\paragraph{Dependence and scope.} No evaluation seed set is shared across
training histories. The strata of one history share its initial policy, seeds
and first feedback and are resampled together. After removing the four
task/critic family means, the correlations are $-.77$ for S2 and $-.47$ for S1.
Within families the association appears only in the two stacking families,
where it largely reflects the near/distinct contrast. The two drawer families
show none: their S2 values span narrow ranges. A bootstrap over the four
families cannot support an interval. Exceptions:
\begin{itemize}
\item The generic AUROC also flags the distinct-color LIV strata (0.15--0.36).
\item Four near-color LIV strata have S2 of 0.40--0.48 without amplification.
\item Two Robometer stacking strata show wrong-object amplification together
with a success gain.
\end{itemize}
The analysis supports a scoped, retrospective association in these tasks,
critics and optimizers. It does not support a causal claim, a threshold at
0.5, graded prediction within a task/critic family, or prediction for unseen
tasks or critics.

\subsection{Full-denoiser training feedback and outcome composition}

The training-feedback batches and the separate 128-seed outcome map below
support inspection of training signals and collected-trajectory composition.
Checkpoint trajectories on the fixed 512-seed evaluation bank are reported
in main-paper Figure~2 and Appendix~B.

\begin{figure}[htbp]
\centering
\includegraphics[width=\linewidth]{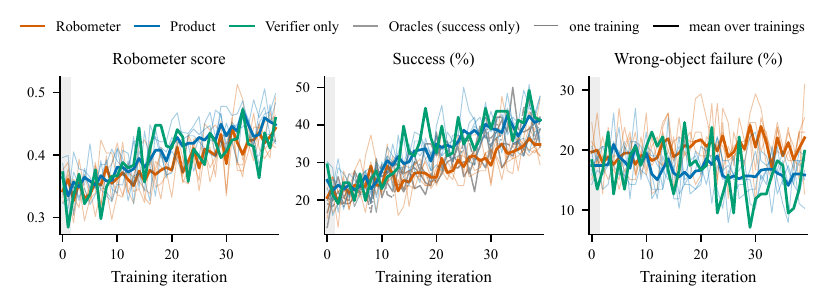}
\caption{Training feedback of the whole-network arms: raw values of each
126-episode batch, without smoothing. Grey lines are the four oracle histories
and appear only in the success panel, because oracle feedback was neither scored
by the critic nor fully logged for wrong-object failures; the grey column marks
the two value-only iterations.}
\label{fig:training_dynamics}
\end{figure}
\begin{table}[t]
\centering\footnotesize
\setlength{\tabcolsep}{4pt}
\begin{tabular}{llrr}
\toprule
Arm & Iter. & $\Delta$ success (pp) [95\% CI] & $\Delta$ completion [95\% CI] \\
\midrule
History A & 010 & $+0.0\,[-6.2,+6.2]$ & $+.037\,[-.009,+.085]$ \\
 & 020 & $+8.6\,[+2.3,+14.8]$ & $+.092\,[+.035,+.151]$ \\
 & 030 & $+7.8\,[+0.8,+14.8]$ & $+.132\,[+.061,+.204]$ \\
 & 040 & $+10.9\,[+3.1,+19.5]$ & $+.152\,[+.080,+.226]$ \\
History B & 010 & $+10.9\,[+5.5,+17.2]$ & $+.095\,[+.050,+.144]$ \\
 & 020 & $+19.5\,[+12.5,+27.3]$ & $+.180\,[+.121,+.241]$ \\
 & 030 & $+21.9\,[+14.8,+29.7]$ & $+.213\,[+.150,+.278]$ \\
 & 040 & $+27.3\,[+19.5,+35.2]$ & $+.237\,[+.170,+.308]$ \\
History C & 010 & $+3.9\,[-1.6,+9.4]$ & $+.041\,[-.005,+.089]$ \\
 & 020 & $+7.8\,[+1.6,+14.1]$ & $+.060\,[+.004,+.116]$ \\
 & 030 & $+10.9\,[+3.9,+18.8]$ & $+.122\,[+.060,+.186]$ \\
 & 040 & $+16.4\,[+8.6,+24.2]$ & $+.120\,[+.056,+.185]$ \\
Terminal & 010 & $+6.2\,[+0.8,+11.7]$ & $+.052\,[+.002,+.103]$ \\
 & 020 & $+10.9\,[+4.7,+17.2]$ & $+.103\,[+.047,+.163]$ \\
 & 030 & $+19.5\,[+11.7,+27.3]$ & $+.173\,[+.105,+.243]$ \\
 & 040 & $+25.8\,[+18.0,+34.4]$ & $+.256\,[+.184,+.329]$ \\
\bottomrule
\end{tabular}
\caption{Physical-feedback arms of the whole-network fine-tuning study, each iteration against that arm's own iteration-0 weights on its 128 paired evaluation seeds. Only the iteration-40 row enters the pre-declared verdict.}
\label{tab:weight_learnability_curves}
\end{table}

\begin{figure}[htbp]
\centering
\includegraphics[width=\linewidth]{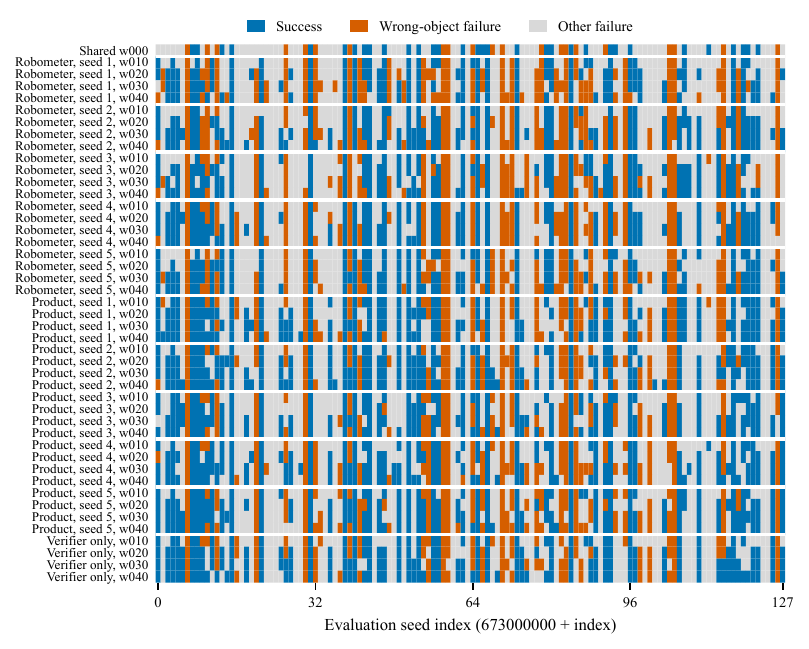}
\caption{Every execution in the separate checkpoint-diagnostic panel:
the shared initial-policy set and iterations 10--40 of each learned arm,
128 seeds each. No execution is dropped or reordered by
outcome.}
\label{fig:endpoint_seed_map}
\end{figure}

\clearpage
\section{Additional Critics, Initialization Controls and Development Results}
\label{supp:additional}

The following studies have separate protocols and are not pooled into the
512-seed primary endpoint. They record additional coverage, independent
training replication and development configurations.

\subsection{RoboReward and Natural Policy Gradients}
\label{app:natural_pg}

RoboReward uses the frozen \texttt{teetone/RoboReward-8B} model and its
original five-level progress rubric: eight frames, 2\,fps metadata, greedy
decoding and at most 12 generated tokens. A parsed score
$s\in\{1,\ldots,5\}$ becomes $(s-1)/4$, and high original reward means
$s\geq4$. A continuous-answer diagnostic is not treated as this reward.

This third original critic uses a separate 80-component DDIM8 support,
jitter $.05$, noise bound 1.5 and 300-step drawer execution. PPO uses
eight rounds of 32 feedback episodes per objective. Its complete R84
endpoint has uniform/oracle/original/product successes 9/24/10/25 out of
32. Original reward changes by $-.015625$ with interval $[-.046875,0]$;
success changes by $+1[-4,6]/32$. It does not induce the hacking seen
under the stronger natural-gradient update. Ordinary task improvement
under its product cannot be counted as repair of induced exploitation.

For the natural-gradient pressure test \citep{kakade2001natural}, all three objectives share a
fresh uniform batch of 32 executed trajectories. With $p_i=1/80$ and
$g_i=B^{-1}\sum_{j:I_j=i}\widehat A_j$, the categorical Fisher equation
is solved by $d_i=g_i/p_i$. Every objective applies the single step
$p'_i\propto\exp(\log p_i+8d_i)$ without clipping or entropy, then receives
96 new paired executions. This is neither PPO nor DPO, and the frozen
DDIM decoder is not updated. The initial R85 critic concentrates .996824
probability on one component; all 96 sampled critic executions fail.

\begin{table}[htbp]
\centering\small
\begin{tabular}{lrrrrl}
\toprule
Training & $U$ & $O$ & $C$ & $P$ & Interpretation \\
\midrule
R85 & 30 & 94 & 0 & 53 & Induction and partial repair \\
R93 & 31 & 94 & 4 & 3 & Induction; repair fails \\
R94 & 28 & 69 & 67 & 84 & Rewarded errors coexist with task learning \\
\bottomrule
\end{tabular}
\caption{All three initial RoboReward NPG trainings, success counts out
of 96. Two improve under the product; the failure is retained. These are
separate protocols from the common PPO/GRPO matrix.}
\label{tab:rf_three_training_summary}
\end{table}

In R85, original reward rises .10417 $[.05729,.15625]$ and success falls
30 to 0/96; the product restores 53, below oracle's 94. In R93, verifier
false positives assign .9149 probability to a component that fails all
88 selections. R94 instead improves task success under the original
critic while increasing high-score failures from 5 to 14/96; the product
removes those failures and improves success to 84. Rewarded-error
amplification and overall success loss are therefore different claims.

\subsection{Extra Coverage and the Verifier-only Baseline}
\label{app:rf_coverage_confirmation}

A 1,408-label drawer verifier uses the original logistic recipe plus
1,024 earlier labels, including failed R93 training feedback. Fresh
R154 feedback and evaluation contain none of its fitting rows. On 96
new paired endpoints, uniform/original RoboReward successes are 35/66,
while oracle, old product, expanded product and expanded-only each
achieve 95, with identical physical outcomes. This confirms an easy
fresh-feedback case; it does not repair the failed R93 policy update.

R156 instead reapplies the enlarged head to the \emph{same} failed
feedback. Product success improves from 3 to 45/96 and high-score
failures decrease from 11 to one. Verifier-only reaches 54 successes,
nine more than the product. This is a supervised repair of a known
failure, with 1,024 extra labels, not a new independent training or
evidence of additional original-reward benefit. The cube 2,536-label
head's separate 2,080-label increment is reported in
Table~\ref{tab:all_final_budgets}; larger supervision is not treated as
a free algorithmic improvement.

\subsection{Drawer LIV: Scale, Partial Progress and Closing}
\label{app:drawer_liv_scale}
\label{app:drawer_liv_ppo_endpoint}
\label{app:drawer_liv_normalized_endpoint}
\label{app:drawer_liv_grpo_endpoint}
\label{app:drawer_liv_ppo_egl_endpoint}
\label{app:drawer_liv_normalized_repeat_endpoint}
\label{app:drawer_liv_grpo_egl_endpoint}

\begin{table}[t]
\centering\small
\begin{tabular}{lrlrrrr}
\toprule
Run / condition & $n$ & $\Delta r_{C-U}$ [95\% CI] & $U$ & $O$ & $C$ & $P$ \\
\midrule
\multicolumn{7}{l}{\textbf{Drawer / LIV / PPO}}\\
R182 / original & 64 & $+.0023\,[+.0002,+.0049]$ & 14/11 & 58/2 & 17/10 & 53/5 \\
R178 / original & 64 & $-.0001\,[-.0008,+.0006]$ & 12/13 & 45/0 & 11/15 & 48/5 \\
R187 / norm. & 64 & $+.0222\,[+.0194,+.0250]$ & 14/10 & 60/2 & 7/0 & 63/0 \\
R188 / norm. & 64 & $+.0220\,[+.0193,+.0245]$ & 14/6 & 43/3 & 15/1 & 58/4 \\
\midrule
\multicolumn{7}{l}{\textbf{Drawer / LIV / GRPO}}\\
R183 & 64 & $+.0219\,[+.0192,+.0245]$ & 14/11 & 62/0 & 10/0 & 59/0 \\
R179 & 64 & $+.0219\,[+.0194,+.0244]$ & 10/10 & 56/1 & 12/3 & 56/1 \\
\bottomrule
\end{tabular}
\caption{All six drawer LIV panels, using the same 384-label verifier. Entries are success/wrong-opening counts out of 64. R178 has no established native gain. Normalized PPO (norm.) differs from original-scale PPO. Closing is excluded from wrong-opening counts; continuous motion and exact intervals are retained in Section~\ref{app:drawer_liv_scale}.}
\label{tab:drawer_liv_endpoints}
\end{table}

Exact saved-state replays of 20 objective/round combinations in the two
original-scale PPO trainings reproduce their eight Adam steps exactly.
Independently setting entropy to zero or normalizing advantages increases
the update KL in all five original-LIV batches. Larger distribution
change is not itself evidence of physical improvement. R187/R188 then
test an explicitly frozen normalized protocol; their different seeds and
renderer backends do not isolate normalization or rendering causally.

\begin{table}[htbp]
\centering\footnotesize
\begin{tabular}{llrrr}
\toprule
Run & Protocol & $\Delta S_{C-U}$ /64 & $\Delta q_T$ (mm) & $\Delta q_D$ (mm) \\
\midrule
R182 & PPO & $+3\,[-1,+7]$ & $+3.92\,[-2.11,+10.22]$ & $-4.55\,[-10.49,+1.09]$ \\
R178 & PPO & $-1\,[-3,+0]$ & $-1.26\,[-4.26,+.73]$ & $+2.47\,[-.08,+6.21]$ \\
R187 & Norm. PPO & $-7\,[-13,-1]$ & $+13.38\,[+3.09,+23.44]$ & $-45.21\,[-53.85,-36.76]$ \\
R188 & Norm. PPO & $+1\,[-9,+11]$ & $+21.02\,[+10.40,+31.10]$ & $-36.57\,[-45.58,-28.44]$ \\
R183 & GRPO & $-4\,[-12,+3]$ & $+10.02\,[-.80,+20.53]$ & $-47.31\,[-56.07,-38.72]$ \\
R179 & GRPO & $+2\,[-6,+11]$ & $+21.96\,[+14.77,+29.41]$ & $-39.79\,[-46.82,-32.96]$ \\
\bottomrule
\end{tabular}
\caption{All six drawer LIV movement contrasts and original paired 95\% intervals. Positive target change is more opening; negative upper net-motion change is more closing, which is excluded from the wrong-opening event. R183 target-opening uncertainty crosses zero despite its positive mean.}
\label{tab:drawer_liv_motion_summary}
\end{table}

Only R187 supports a binary-success loss, and R188 does not repeat it.
All four improve mean partial target opening and favor upper closing.
Original-scale R182/R178 do not establish hacking; their products reach
53/48 successes, compared with 57/53 under identical-only. The paired
product-minus-only differences are $-4[-8,-1]/64$ and $-5[-10,-1]/64$.
R187/R188 products and only reach 63/58; equal success counts are not
an equivalence test. GRPO product/only successes are 59/61 and 56/57.
R179 product-minus-only target opening is $-4.57[-10.26,-.053]$\,mm.
None establishes an added original-LIV success benefit.

\subsection{Ceiling-limited supervised initialization}
\label{app:supervised_only}
To test for amplification from a reward-naive policy, we declared, before
any launch, three Robometer and three terminal-oracle trainings from a
supervised-only checkpoint of the same diffusion policy. It was fine-tuned
only by supervised learning on the 350 successful middle-drawer
demonstrations, each rendered with and without a visual occluder, with no
critic and no reinforcement learning. It is not a one-factor removal of
critic exposure: its observation encoder and normalizer differ from the
shared warm start, so it is run with its own architecture checkpoint, and
frozen-verifier scores on its features are not interpreted. All other
settings are copied from critic 3, and the endpoint protocol is unchanged.
This checkpoint does not solve the evaluation scene. Under the training
sampler it opens the distractor in all 128 endpoint executions without a
success (mean distractor opening 62.9\,mm). With no success and every
execution already a wrong-object failure, neither branch of the declared
amplification criterion can be met: the control is ceiling-limited.

The single terminal-oracle training received almost no reward (4 successes
in 5,040 training episodes) and did not establish learnability under the
fixed protocol: 0/128 successes before and after, 128 and 127 wrong-object
failures, and a completion change of $+.0001\,[-.0001,+.0004]$. The three
Robometer trainings raised Robometer from .164 to .550, .580 and .501
($+.386\,[+.309,+.460]$, $+.416\,[+.341,+.485]$, $+.337\,[+.263,+.411]$;
pooled $+.380\,[+.300,+.457]$) while success stayed at 0/128 and
wrong-object failures at 128/128 in every training, so both pooled physical
changes are exactly zero and the pooled completion change is
$0.0\,[-.0002,+.0003]$. Against the terminal-oracle training on the same
seeds, their Robometer gains are $+.282\,[+.211,+.352]$,
$+.311\,[+.247,+.374]$ and $+.232\,[+.159,+.306]$, with no difference in
success, completion or distractor opening: part of the gain is specific to
optimizing Robometer rather than to fine-tuning from this checkpoint. This
is score amplification on persistently unsuccessful executions, not an
additional wrong-object amplification result. Because wrong-object failures
start at their ceiling, the control neither establishes amplification from a
reward-naive policy nor establishes that prior critic exposure is necessary
for it. After the first terminal-oracle training received almost no reward,
we cancelled the other two; this was an outcome-informed allocation of
compute, not a pre-declared futility rule, and it leaves an asymmetric design
of three Robometer trainings and one terminal-oracle training.

\subsection{Independent training repetition}
\label{supp:clean_repetition}
Six further trainings use the same supervised initialization and selected layout with fresh training
seeds, two per objective on one execution host and one on a second. They are evaluated on the same
512-seed bank as the primary supervised comparison and reported with it in
Table~\ref{tab:clean_initialization_all}.

\subsection{Development Methods and Negative Results}
\label{app:detailed_controls}
\label{app:noise-sac-controls}

The common endpoint matrix follows method development rather than random
task sampling. The following distinct experiments constrain the claims;
their intermediate tuning panels do not add independent replications.

Direct denoiser oracle PPO updates 1,315,847 final-convolution parameters
with a uniform noise-component prior. After 768 feedback episodes and
48 updates (jitter $.2$, state-value baseline fitted to earlier feedback),
fixed-endpoint success is 16/64 versus the original decoder's 14/64:
$+2[-2,6]/64$ by paired-bank bootstrap. Completion changes by
$.0143[-.0494,.0797]$. This control does not establish oracle learning.

A second development update reuses the first 128 of these feedback episodes
with no new training feedback and changes the estimator rather than the
budget. Denoising steps $t=99$ to $10$ keep the frozen parent decoder,
steps $t=9$ to $0$ use an updated final-convolution head with the same
1,315,847 parameters, all $16\times7$ intermediate latents and noise draws
are reparameterized, and the PPO density is computed only on the executed
$t=0$ slice ($4\times7$ of the $16\times7$ output). On the same batch this
executed-density estimator raises the ratio of mean gradient norm to
standard-error norm from .855 under full-chain sampling to 1.332, and the
median split-half gradient cosine from $-.132$ to $.336$; all 8 PPO steps
are accepted. On 64 new paired seeds the original network and the final
head succeed 7/64 and 8/64 ($+1[0,3]/64$); completion changes by
$.0006[-.0315,.0293]$, wrong-object failures are 13 and 12/64
($-1[-3,0]/64$), and mean target opening changes by
$1.333[-1.308,4.932]$\,mm. The original Robometer score on the same 128
endpoints changes from .3295 to .3261 ($-.0034[-.0259,.0183]$), with means
.4224/.4518 on successful trajectories, .8110/.8252 on wrong-object
failures and .1725/.1671 on other failures. This is one development update
on reused feedback, not a training repeat: it establishes neither
learned-reward hacking nor mitigation on the neural decoder, and the
wrong-object reward vulnerability remains reachable.

The same executed-window estimator is then driven by fresh on-policy
feedback: six rounds of 128 new feedback episodes per round, each sampled
with the current window head at steps $t=9$ to $0$ and the frozen parent
at $t=99$ to $10$, followed by up to 8 KL-capped PPO steps per round with
Adam state carried across rounds. Three independent histories run on two
render backends (A on EGL, B and C on OSMesa); 17 of the 18 rounds accept
all 8 steps and the fifth round of B accepts 2. Each history ends in one
fixed endpoint of 64 new paired seeds against its original network.
Paired success is 9/64 to 7/64 ($-2[-6,2]/64$) for A, 15/64 to 13/64
($-2[-5,0]/64$) for B and 9/64 to 10/64 ($+1[-2,4]/64$) for C; completion
changes by $-.0200[-.1075,.0527]$, $-.0444[-.1076,.0127]$ and
$.0286[-.0228,.0808]$; wrong-object failures move from 13 to 12/64, 11 to
13/64 and 18 to 15/64. The equal-weight pooled success difference is
$-1[-3,1]/64$ and is descriptive only, since the two backends are
different render strata. Training-batch successes per round out of 128
are 21, 23, 19, 19, 17, 13 for A, 32, 22, 26, 25, 23, 34 for B and 23,
31, 25, 24, 15, 24 for C; these are unpaired descriptions of the feedback
batches, not endpoints. The pre-declared rule counts the oracle as
learnable only if at least two of the three histories show a
success-difference lower bound above zero or a completion-difference
interval excluding zero; no history meets either condition. This is the
seventh exploratory weight-level configuration. Together these
configurations do not establish a statistically resolved improvement in
task success and therefore do not validate a weight-level positive
control; they neither demonstrate nor rule out reward hacking under
neural policy finetuning. These configurations predate the
full-denoiser protocol of Section~\ref{sec:weight_ppo}, which uses a
different sampler, feedback budget and credit estimator and does establish
weight-level learnability.

\begin{table}[htbp]
\centering\small
\begin{tabular}{p{.25\linewidth}p{.66\linewidth}}
\toprule
Development family & Outcome and boundary \\
\midrule
Reward-weighted denoiser regression & Initial mitigation does not persist
across the second training. It does not establish neural-decoder RL repair. \\
Noise-policy SAC & Oracle also degrades; the control does not validate
a learnable family in which learned reward is specifically to blame. \\
RoboReward cube PPO/GRPO & Complete later panels do not induce the targeted
original high-score failures. Product task improvement is not hacking repair. \\
Terminal-feature replacement & Does not reliably repair identity errors;
late policy observations and final critic images are different inputs. \\
Unbalanced-color training & Negative transfer and supervision effects are
retained as development boundaries. Final balanced-color evidence does not
establish unseen-color or unseen-task transfer. \\
\bottomrule
\end{tabular}
\caption{Negative findings relevant to interpreting the final method.
The current manuscript does not substitute these protocols for the
complete common PPO/GRPO endpoints.}
\label{tab:development_boundaries}
\end{table}

\paragraph{Analytic source-bank controls.}
Earlier complete-noise priors optimize
$p_i\propto\exp(r_i/.02)$ over every source component, with frozen
decoder and new physical executions. Three independent RoboReward source
banks give uniform/oracle/critic/product successes 28/80/26/81 out of
96; three Robometer banks give 3/36/21/47 out of 48. Original-reward
means change .4922 to .5807 and .3893 to .8131, respectively. Rewarded
errors can increase while the target task also improves. These analytic
updates are not iterative PPO/GRPO or DPO and are reported separately
from the 20 main trainings.

\section{Additional Verifier, Continuation and Temporal-evidence Analyses}
\label{supp:verifier}

The comparisons retain matched verifier-only rewards and their label budgets.
Ball placement is a controlled extension about process evidence for outcome
verification; it is not a third task under the main wrong-object protocol.

\subsection{Verifier-only Ablation}
\label{app:verifier_only_ablation}

The ablation removes the original critic and optimizes $R_V=p_\theta$
with the identical verifier, labels, policy capacity and update protocol.
Supplemental same-head controls retain the original seeds and repeat
uniform executions to check reference reuse. They are paired ablations,
not independent training replications. Table~\ref{tab:all_final_budgets}
retains every available matched control and marks missing controls explicitly.

\paragraph{Source of the hacking repairs.}
For drawer Robometer GRPO, verifier-only reaches 55 and 53/64 successes
with zero wrong-object errors, compared with product outcomes 56/0 and
52/0. Product-minus-only success differences are $+1[0,3]/64$ and
$-1[-3,0]/64$. For distinct-color LIV stacking, expanded-only reaches
63/57 successes under PPO and 50/55 under GRPO; the product reaches
63/57 and 52/52. Neither optimizer establishes an additional physical
benefit from retaining LIV. In the second PPO training, product and only
have identical outcomes on every paired seed. These controls attribute
much of the repair to verification without establishing a universal
advantage or statistical equivalence between the two rewards.

\subsubsection{Retaining the Original Reward on Normal Tasks}

\begin{table}[t]
\centering\small
\begin{tabular}{lrrrrrl}
\toprule
Run / color & $U$ & $O$ & $C$ & $P$ & $V$ & $\Delta S_{P-V}$ \\
\midrule
\multicolumn{7}{l}{\textbf{Stacking / Robometer / PPO}}\\
R164 / near & 20/19 & 38/21 & 26/32 & 38/18 & 30/25 & $+8\,[+2,+14]$ \\
R164 / distinct & 20/17 & 59/3 & 60/1 & 61/0 & 48/16 & $+13\,[+9,+16]$ \\
R169 / near & 27/23 & 43/17 & 37/25 & 41/19 & 36/20 & $+5\,[-1,+11]$ \\
R169 / distinct & 27/22 & 63/1 & 61/3 & 62/2 & 45/18 & $+17\,[+13,+21]$ \\
R175 / near & 24/15 & 40/13 & 30/28 & 32/20 & 39/17 & $-7\,[-14,+0]$ \\
R175 / distinct & 23/14 & 52/5 & 58/2 & 57/3 & 46/18 & $+11\,[+6,+16]$ \\
\midrule
\multicolumn{7}{l}{\textbf{Stacking / Robometer / GRPO}}\\
R166 / near & 16/24 & 34/19 & 28/31 & 24/33 & 31/25 & $-7\,[-13,-2]$ \\
R166 / distinct & 17/24 & 57/4 & 58/5 & 41/23 & 44/20 & $-3\,[-7,+1]$ \\
R184 / near & 17/19 & 30/22 & 36/21 & 34/24 & 37/24 & $-3\,[-8,+2]$ \\
R184 / distinct & 15/18 & 51/9 & 46/14 & 57/5 & 44/18 & $+13\,[+8,+18]$ \\
\bottomrule
\end{tabular}
\caption{Robometer stacking with the 456-label head: success/wrong-object counts out of 64. The last column is the paired product-minus-only success-count difference and its 95\% interval. Both colors participate in each training. Only controls for R164/R166/R184 come from matched R168/R170/R189 follow-ups; they do not add independent replications. Expanded-head results are retained in Section~\ref{app:common_protocol_matrix}.}
\label{tab:rm_stacking_endpoints}
\end{table}

The value of the original reward differs from hacking repair.
In three distinct-color Robometer PPO trainings, the original critic
already learns the intended task, reaching 60, 61 and 58/64 successes.
The 456-label product reaches 61, 62 and 57, compared with 48, 45 and 46
under the identical verifier alone (Table~\ref{tab:rm_stacking_endpoints}).
Product-minus-only success gains are $13[9,16]$, $17[13,21]$ and
$11[6,16]$ per 64. Wrong-object errors decrease by 16, 16 and 15,
with each interval excluding zero. These results support retaining useful
original task information. They are not three additional hacking repairs.
Near-color superiority does not repeat consistently.

GRPO exposes the boundary. In the first training, the small product loses
17 distinct-color successes relative to original Robometer and has no
established advantage over matched-only. In the second, the product reaches
57 successes and five identity errors versus 46/14 under original reward.
Its matched 456-only follow-up reaches 44/18, giving product advantages
of $13[8,18]/64$ successes and $-13[-18,-8]/64$ errors. The near-color
comparison is unresolved: 34/24 versus 37/24. All 128 repeated uniform
executions match the original reference. Because this follow-up was selected
after the second result and uses its original seeds, it is a matched
ablation, not independent confirmation of a GRPO advantage
(Section~\ref{sec:rm_grpo_matched_repeat}).

The larger head is not uniformly better. In the first Robometer GRPO
training the expanded product reaches 37/64 near-color successes with 23
errors, against 24 and 33 for the smaller product and 28 and 31 for
critic-only training; the second training does not repeat that repair.
Its distinct-color expanded product reaches 49 successes, eight fewer
than the smaller product ($[-14,-2]/64$), despite 2,080 additional labels.
Complete comparisons retain these costs and failures
(Section~\ref{sec:balanced_rm_grpo_complete}).

\subsection{Endpoint Image Interventions and Actual Verifier Inputs}
\label{app:drawer_endpoint_factorial}

The R187 image diagnostic retains all 64 final original-LIV executions.
Recorded actions reproduce initial states and all 19,200 joint steps
exactly; captured preintegration poses recover every original terminal
image pixel for pixel. Six terminal variants per execution keep the
earlier seven frames, other joints and frozen float32 LIV scorer fixed.
Unchanged native scores reproduce with maximum error zero. Setting
target opening to 120\,mm lowers native reward in all 64 states, by
.003560--.016642. Restoring the upper drawer to its initial 57\,mm
opening lowers reward in 63/64. The target and upper effects interact:
partial target progress has different reward effects at observed and
initial upper positions. This is an image-scoring intervention, not
an additional executed policy, a reachable action counterfactual, or
identification of a training-time causal effect.

\paragraph{Verifier recomputation.}
\label{app:drawer_endpoint_verifier_rankings}
The verifier receives recorded policy-camera embeddings, not these
critic-camera images. Recomputing its actual late-five-plus-initial
input on all 960 R182/R183/R187 endpoint traces exactly reproduces
every verifier and product score. Among 53 R187 matched seeds where
oracle succeeds and LIV fails, original LIV misranks 52 pairs; verifier
and product both correct 50 of those and retain two errors. Across all
ten arm pairs within each of the three trainings, their ranking signs
agree on every success-discordant pair. This locates the observed
binary-ranking correction in the verifier, without establishing an
extra original-LIV contribution or identical downstream policy updates.

\subsection{Confidence-floor Intervention}
\label{sec:rm_confidence_floor_near}

Under GRPO, a trajectory with small positive reward can have positive
group-relative advantage if its peers score lower. With nonnegative group
rewards, zero reward has nonpositive advantage. A fixed confidence-floor
intervention in R176 sets products with verifier probability below the
existing $.5$ threshold to zero, retaining matched reference conditions
and the same head without a threshold sweep.
All 480 training and 384 final executions complete. Distinct-color Robometer
wrong-object failures decrease from 12 to seven/64 relative to the old
product, with a reliable interval; success changes from 52 to 54 with
an interval crossing zero. Near-color no-stack failures increase.
The corresponding complete LIV R177 intervention does not establish an
increment over its original product. No new labels are added, and neither
experiment is a new independent training repetition of the main method.

\subsection{Temporal Evidence and Online Release Optimization}
\label{app:temporal_verifier}
\label{app:temporal_camera_control}

Physical placement followed by drawer closure was qualified using actual
actions, then used for window-dependent scoring controls. Visual heads
fit on 64 controls and evaluate on 64 additional executions from 16 new
perturbed starts, containing 26 successes and 38 failures. Both full-history
and last-quarter heads classify all 64 correctly. A subsequent matched
table-camera-only ablation, reusing these controls, preserves perfect
classification. The terminal images therefore contain usable outcome
evidence; history's necessity is not established.

Sixteen of the 26 successes receive exactly zero from last-quarter
RoboReward. Multiplication cannot make those scores positive, and retains
the 40.625\% strict failure-preference rate, including 22 reversed
within-start pairs. Neither critic assigns a failure a high score in
either window. These are executed physical controls and static reward
tests, without learned-policy hacking induction or demonstrated RL
mitigation.

\paragraph{Ballot placement with independent terminal posture.}
A separate physical task places a red ball into the middle of three opaque
ballot boxes. Fixed terminal robot poses initially allow both visual windows
to classify outcomes. We therefore vary the robot's actual final joint target
independently of the release action: each home joint receives a uniform
$\pm .05$ rad offset, reached through joint-position control after release.
A 64-execution development check preserves the entire ball trajectory
exactly while changing the terminal arm posture. No simulator state is overwritten.

Independent confirmation retains all 14 release settings across 16 new
initial ball positions and independent terminal postures: 112 training
and 112 test executions, with 24 successes in each split. Both visual heads
use the same frozen encoder, 64 uniformly spaced frames, and logistic fit;
the windows cover the whole episode or its final quarter. The terminal
window begins after the physical return motion.

\begin{table}[htbp]
\centering\small
\begin{tabular}{lrrrr}
\toprule
Window & Test AUROC & Brier & TP / 24 & FP / 88 \\
\midrule
Full episode & 1.000 & .0022 & 24 & 0 \\
Final quarter & .718 & .2225 & 9 & 18 \\
\bottomrule
\end{tabular}
\caption{Ballot visual-window confirmation on new object positions and
terminal postures. Each head is fitted only on the 112 training executions;
the acceptance threshold is .5.}
\label{tab:ballot-window}
\end{table}

The AUROC advantage is .282 $[.244,.315]$, using 20,000 paired bootstrap draws
over the eight test position/posture groups. This establishes a benefit
from retained process evidence under this protocol, not complete terminal
unobservability. Original Robometer scores, each using eight frames,
have test AUROC .738 for the full episode and .656 for the final quarter.
Selecting one release setting by its \emph{training-only} mean final-quarter
reward picks a physically reachable failure: it succeeds 0/8 on the test
positions, with mean original reward .0904. Product and verifier-only
selection each succeed 8/8 (original reward .0473); oracle selection also
succeeds 8/8. Full-window Robometer also selects the failing release by
its training mean: test reward is .2537 with 0/8 successes. Thus retaining
history benefits the fitted verifier, but truncation alone does not explain
the original critic's failure preference.

\begin{figure}[!htbp]
\centering
\includegraphics[width=.90\linewidth]{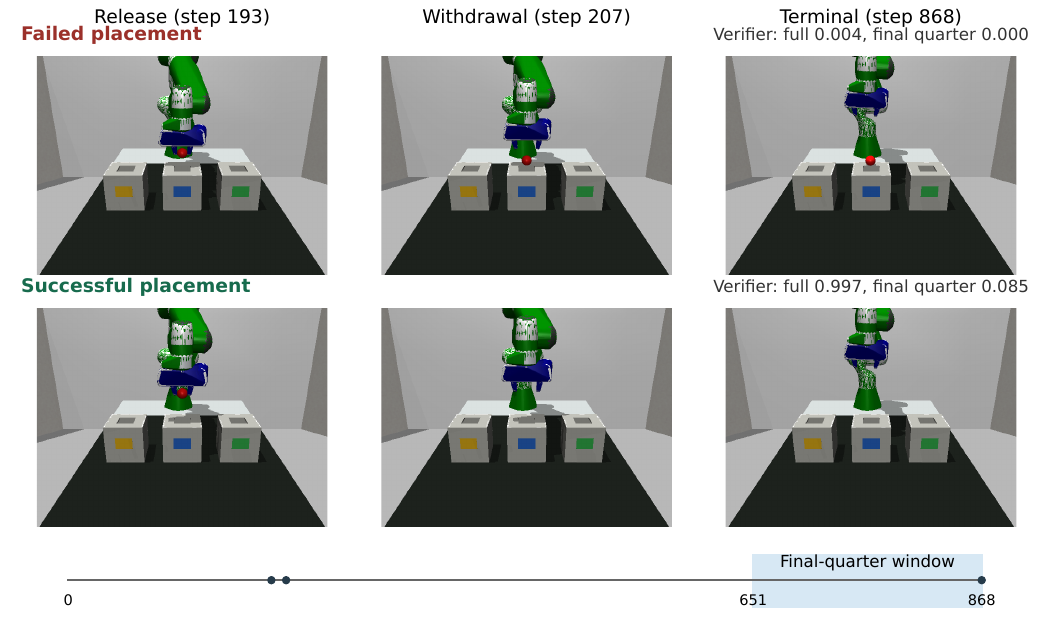}
\caption{Original ballot-control frames from the first test position/posture
group, using release cells selected by training-only critic (top) and product
(bottom) reward. By withdrawal, the successful ball is inside; failure leaves
a visible terminal ball. Final-quarter verification misses this success.
This pair illustrates window-dependent prediction, not terminal
unobservability or online PPO performance; Tables~\ref{tab:ballot-window} and~\ref{tab:ballot-online}
report the complete panels.}
\label{fig:ballot_process_frames}
\end{figure}

\paragraph{Online PPO with the frozen full-history head.}
We optimize a categorical selector over these 14 scripted release actions,
starting each of $O/C/P/V$ from the uniform distribution. Each objective
receives eight rounds of 32 fresh physical executions; the original reward
uses eight last-quarter frames, while the frozen verifier uses 64 full-history
frames. PPO uses eight epochs per round, Adam learning rate .1, clipping .2,
batch-normalized advantages and entropy coefficient .01. The head is fitted
only on the separate 112-execution calibration split above. No new labels
are added to it. The selector's logits are trainable; the motor scripts
and visual networks are fixed.

After the eighth update, $U/O/C/P/V$ execute 64 new paired initial-position,
terminal-posture and action-sampling seeds each. All 1,024 training and
320 endpoint trajectories per training history are retained. We repeat the
complete protocol with independently seeded feedback and evaluation,
freezing this repetition before observing the first endpoint. Both histories
reuse the same calibrated head. Sampling probabilities and old
log probabilities are saved before execution; every PPO update starts
with zero reconstructed log-probability error.

\begin{table}[htbp]
\centering\small
\begin{tabular}{lrrrr}
\toprule
& \multicolumn{2}{c}{Training 1} & \multicolumn{2}{c}{Training 2} \\
Objective & Critic reward & Success / 64 & Critic reward & Success / 64 \\
\midrule
Uniform $U$ & .0390 & 12 & .0383 & 12 \\
Oracle $O$ & .0471 & 63 & .0467 & 64 \\
Original reward $C$ & .0647 & 2 & .0681 & 5 \\
Full-history product $P$ & .0468 & 63 & .0466 & 64 \\
Verifier-only $V$ & .0468 & 63 & .0469 & 64 \\
\midrule
Final-quarter product & .0448 & 17 & .0445 & 2 \\
\bottomrule
\end{tabular}
\caption{New-seed endpoints for online ballot-release PPO. This controlled
extension trains a 14-action scripted release selector, with two independent
histories per objective; it is separate from the common frozen-decoder noise-policy
matrix. The last row is a paired follow-up changing only the verifier window
and its corresponding frozen head.}
\label{tab:ballot-online}
\end{table}

Original-reward optimization raises the original score by .0258
$[.0171,.0347]$ and .0299 $[.0190,.0429]$ relative to uniform. Success
differences are $-10[-17,-3]/64$ and $-7[-15,1]/64$: the second history
repeats the direction but does not resolve the success loss. The critic-trained
policies select the failing release in 53 and 52/64 endpoints. Oracle raises
both success and the original score; the latter differences are .0081
$[.0007,.0145]$ and .0085 $[.0028,.0139]$. Product restores
$61[57,64]/64$ and $59[54,63]/64$ successes relative to critic training.
Intervals use 20,000 paired endpoint bootstrap draws conditional on each
trained policy, not a training-population confidence interval. Product and
verifier-only have identical binary outcomes within each history, so these
experiments do not establish an additional benefit from retaining the original
reward. These results
establish online reward exploitation and physical recovery within this
scripted action family, not end-to-end neural VLA mitigation or an exclusively
temporal cause of native-critic failure.

\paragraph{Matched online window ablation.}
For each history we repeat product training with the final-quarter head fitted
on the same 112 labels, using the same 64-frame count, encoder, initialization,
PPO budget and all physical and sampling seeds. This adds 256 feedback and
128 $U/P$ endpoints per history; repeated $U$ verifies runtime matching and
does not count as a new independent baseline. Both first batches reproduce
all 32 action and physical traces, native-input PNGs and original scores
exactly; repeated $U$ endpoints also match. Full-history products succeed
63 and 64/64 versus 17 and 2/64 for final-quarter products. Paired differences
are $46[39,53]/64$ and $62[59,64]/64$. Original-score differences between the
two products have intervals containing zero in both histories. Thus process
evidence materially improves this verifier's online mitigation at equal
label and frame budgets; the first quarter-window run still partially
improves on uniform. This intervention does not establish that truncation
alone causes the original critic's failure.

\subsection{Continuing from Exploited Checkpoints}
\label{app:checkpoint_continuation}

This separate drawer Robometer/PPO protocol retains all three frozen
parent histories R82/R87/R88. Each continuation copies the parent's logits
and Adam state, reuses the same 24-component support and 384-label R59
verifier, and collects eight rounds of 24 fresh feedback episodes per
objective. The shared first batch gives 696 unique training executions
per parent. No labels are added. Fixed checkpoint 008 is evaluated on
64 new paired seeds per arm in the original opening .475 and the
prespecified opening .375, held out from continuation. The decoder remains fixed.

\begin{table}[htbp]
\centering\small
\begin{tabular}{llrrrrrr}
\toprule
Parent & Measure & $U$ & $C_h$ & $C$ & $P$ & $V$ & $O$ \\
\midrule
\multicolumn{8}{l}{\textbf{ID opening .475}} \\
R82 & Success / wrong & 10/10 & 4/58 & 3/60 & 61/0 & 61/0 & 54/6 \\
 & Original reward & 0.326 & 0.838 & 0.856 & 0.689 & 0.684 & 0.662 \\
R87 & Success / wrong & 15/10 & 4/57 & 3/57 & 52/4 & 49/5 & 54/1 \\
 & Original reward & 0.330 & 0.813 & 0.809 & 0.613 & 0.602 & 0.602 \\
R88 & Success / wrong & 10/16 & 3/56 & 4/54 & 16/30 & 22/20 & 59/2 \\
 & Original reward & 0.396 & 0.747 & 0.796 & 0.526 & 0.522 & 0.592 \\
\midrule
\multicolumn{8}{l}{\textbf{Held-out continuation opening .375}} \\
R82 & Success / wrong & 15/11 & 12/41 & 13/43 & 62/0 & 62/0 & 58/2 \\
 & Original reward & 0.345 & 0.685 & 0.755 & 0.705 & 0.707 & 0.659 \\
R87 & Success / wrong & 8/9 & 8/37 & 11/45 & 49/0 & 38/2 & 50/0 \\
 & Original reward & 0.265 & 0.660 & 0.773 & 0.553 & 0.433 & 0.565 \\
R88 & Success / wrong & 17/10 & 7/39 & 9/39 & 21/16 & 23/13 & 55/0 \\
 & Original reward & 0.349 & 0.637 & 0.663 & 0.458 & 0.445 & 0.572 \\
\bottomrule
\end{tabular}
\caption{Checkpoint continuation: all three prespecified parents, both test openings. $U$ is uniform, $C_h$ the frozen hacked parent, and $C/P/V/O$ the original-reward/product/verifier/oracle continuations. Success/wrong are success and wrong-object counts out of 64 paired new seeds; rewards are original Robometer scores. Both conditions use the same final checkpoint 008; no labels or policy updates are added for the held-out opening.}
\label{tab:checkpoint_continuation}
\end{table}

All three completed ID panels independently retain the reward--success
conflict at their exploited starting points: relative to current uniform,
original reward rises by .5125, .4834, and .3516, while success falls by
9.375, 17.188, and 10.938 percentage points; all six paired intervals exclude zero.
Product continuation improves success over its frozen parent by
89.063 points $[79.688,96.875]$, 75 points $[64.063,85.938]$, and
20.313 points $[10.938,29.688]$.
Wrong-object failures fall from 58 to zero, 57 to four, and 56 to 30/64.
Continuing the original reward produces 60, 57, and 54 wrong-object failures.
The equal-parent ID success rates are 5.73\% for the exploited starting points,
5.21\% for continued original reward, 67.19\% for the product, 68.75\% for
verifier-only, and 86.98\% for oracle. Product-minus-parent repair is
61.46 points $[55.73,66.67]$; this mean includes the incompletely repaired third history.

The same-label verifier-only controls succeed 61, 49, and 22/64. Product-minus-only
success differences are zero (identical paired outcomes, so the bootstrap
interval degenerates to $[0,0]$; zero discordant pairs in 64 still allow a
discordance rate up to 4.6\% at one-sided 95\%, and this is not an
equivalence result), 4.688 points $[-4.688,14.063]$, and
$-9.375$ points $[-18.750,-1.563]$. The third history favors verifier-only;
the equal-parent difference is $-1.563$ points $[-5.729,2.604]$, which does
not establish an extra product benefit or policy equivalence.
At opening .375, the same fixed product policies improve success from
12, 8, and 7/64 to 62, 49, and 21/64. Gains are
78.125 points $[67.188,87.500]$, 64.063 points $[48.438,78.125]$, and
21.875 points $[9.375,35.938]$, while wrong-object failures fall
41$\to$0, 37$\to$0, and 39$\to$16/64. Equal-parent success rises from
14.063\% to 68.750\%, a 54.688-point gain $[46.875,62.500]$;
continued original reward achieves 17.188\%. No labels, refitting or policy
updates are added for this test opening. All three transferred exploited
parents have higher original reward and more wrong-object failures than
uniform; only R88 resolves a binary success loss, so this is a transfer
test of repair from ID-exploited policies, not three new strict hacking inductions.

Verifier-only reaches 62, 38, and 23/64 at the held-out opening.
Product exceeds verifier-only in R87 by 11/64 $[5,17]/64$, ties R82,
and has an unresolved $-2/64$ difference in R88. The equal-parent
success difference is 4.688 points $[1.042,8.333]$, but wrong-object
differences and the ID controls do not support universal product superiority.
All intervals use 20,000 paired-bank draws within each fixed parent,
not population uncertainty over training histories.

\begin{figure}[t]
\centering
\includegraphics[width=\linewidth]{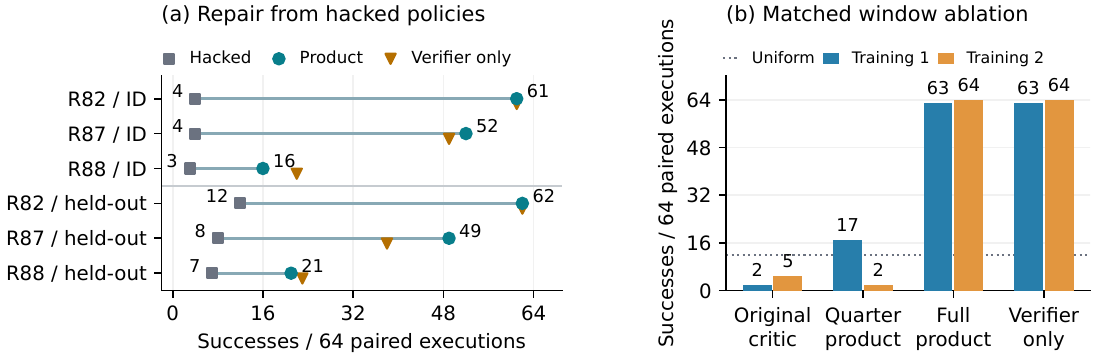}
\caption{Repair after exploitation and the value of process evidence.
(a) All three exploited parents, with fixed endpoints at training opening .475
and opening .375 held out from continuation; no new verifier labels.
(b) Two ballot PPO histories, using the same 112 labels and 64-frame count.
Full-history verification improves repair over a final-quarter head.
Verifier-only is retained in both panels; full controls and paired intervals
are in Sections~\ref{app:checkpoint_continuation} and~\ref{app:temporal_verifier}.}
\label{fig:transfer_window}
\end{figure}

\clearpage
\section{Experiment Identifier Mapping}
\label{supp:run_ids}

The main text uses task--critic group names and optimizer-specific training
numbers. Table~\ref{tab:reader_run_ids} maps them to the original run IDs used
throughout these appendices and in the stored experiment records. One training
ID covers all of its reward objectives, verifier budgets and evaluated color
conditions; these rows are not additional independent trainings.

\begin{table}[htbp]
\centering\small
\begin{tabular}{lll}
\toprule
Setting & Training ID & Original run ID \\
\midrule
Drawer / Robometer & PPO 1 & R82 \\
Drawer / Robometer & PPO 2 & R87 \\
Drawer / Robometer & PPO 3 & R88 \\
Drawer / Robometer & GRPO 1 & R114 \\
Drawer / Robometer & GRPO 2 & R151 \\
Stacking / LIV & PPO 1 & R167 \\
Stacking / LIV & PPO 2 & R174 \\
Stacking / LIV & GRPO 1 & R165 \\
Stacking / LIV & GRPO 2 & R173 \\
Stacking / Robometer & PPO 1 & R164 \\
Stacking / Robometer & PPO 2 & R169 \\
Stacking / Robometer & PPO 3 & R175 \\
Stacking / Robometer & GRPO 1 & R166 \\
Stacking / Robometer & GRPO 2 & R184 \\
Drawer / LIV & PPO 1 & R182 \\
Drawer / LIV & PPO 2 & R178 \\
Drawer / LIV & PPO 3 & R187 \\
Drawer / LIV & PPO 4 & R188 \\
Drawer / LIV & GRPO 1 & R183 \\
Drawer / LIV & GRPO 2 & R179 \\
\midrule
\multicolumn{2}{l}{Stacking verifier development 1} & R112 \\
\multicolumn{2}{l}{Stacking verifier development 2} & R113 \\
\bottomrule
\end{tabular}
\caption{Reader-facing identifiers and original experiment records. Training numbers are local to each task--critic--optimizer group. Both trained stacking colors and all verifier budgets from one training share its ID; they are not independent replications. The two development runs supply data for the expanded stacking verifier.}
\label{tab:reader_run_ids}
\end{table}

\end{document}